\documentclass[sigconf]{acmart}
\copyrightyear{2022}
\acmYear{2022}
\setcopyright{acmcopyright}
\acmConference[MM '22]{Proceedings of the 30th ACM International Conference on Multimedia}{October 10--14, 2022}{Lisboa, Portugal}
\acmBooktitle{Proceedings of the 30th ACM International Conference on Multimedia (MM '22), October 10--14, 2022, Lisboa, Portugal}
\acmPrice{15.00}
\acmDOI{10.1145/3503161.3548381}
\acmISBN{978-1-4503-9203-7/22/10}
\begin{document}

\title[OCFD-Net]{Learning Occlusion-aware Coarse-to-Fine Depth Map for Self-supervised Monocular Depth Estimation}



\author{Zhengming Zhou}
\email{zhouzhengming2020@ia.ac.cn}
\orcid{0000-0002-6792-0739}
\author{Qiulei Dong}
\authornote{Corresponding author.}
\email{qldong@nlpr.ia.ac.cn}
\orcid{0000-0003-4015-1615}
\affiliation{%
  \institution{National Laboratory of Pattern Recognition, Institute of Automation, Chinese Academy of Sciences}
  \city{Beijing}
  \country{China}
}
\affiliation{%
  \institution{School of Artificial Intelligence, University of Chinese Academy of Sciences}
  \city{Beijing}
  \country{China}
}


\begin{abstract}
  Self-supervised monocular depth estimation, aiming to learn scene depths from single images in a self-supervised manner, has received much attention recently.
  In spite of recent efforts in this field, how to learn accurate scene depths and alleviate the negative influence of occlusions for self-supervised depth estimation, still remains an open problem.
  Addressing this problem, we firstly empirically analyze the effects of both the continuous and discrete depth constraints which are widely used in the training process of many existing works.
  Then inspired by the above empirical analysis, we propose a novel network to learn an Occlusion-aware Coarse-to-Fine Depth map for self-supervised monocular depth estimation, called OCFD-Net.
  Given an arbitrary training set of stereo image pairs, the proposed OCFD-Net does not only employ a discrete depth constraint for learning a coarse-level depth map, but also employ a continuous depth constraint for learning a scene depth residual, resulting in a fine-level depth map.
  In addition, an occlusion-aware module is designed under the proposed OCFD-Net, which is able to improve the capability of the learnt fine-level depth map for handling occlusions.
  Experimental results on KITTI demonstrate that the proposed method outperforms the comparative state-of-the-art methods under seven commonly used metrics in most cases. In addition, experimental results on Make3D demonstrate the effectiveness of the proposed method in terms of the cross-dataset generalization ability under four commonly used metrics.
  The code is available at https://github.com/ZM-Zhou/OCFD-Net\_pytorch.
\end{abstract}



\maketitle

\section{Introduction}
Monocular depth estimation, which aims to estimate scene depths from single images, is a challenging topic in the computer vision community.
According to whether ground truth depths are given for model training, the existing methods for monocular depth estimation could be divided into two categories: supervised monocular depth estimation methods~\cite{Eigen2014Depth, Li2015Depth, Cao2017Estimating, Fu2018Deep, Gan2018Monocualr, Xing2021Gated} and self-supervised monocular depth estimation methods~\cite{Garg2016Unsupervised, Zhou2017Unsupervised, Godard2019Digging, Watson2019Self, Gonzalezbello2020Forget}.
Since it is difficult and time-consuming to obtain high-quality and dense depths for large-scale outdoor scenes as ground truth, self-supervised monocular depth estimation has attracted more and more attention in recent years.

The existing works for self-supervised monocular depth estimation generally use either monocular video sequences~\cite{Zhou2017Unsupervised, Godard2019Digging} or stereo image pairs~\cite{Garg2016Unsupervised, Gonzalezbello2020Forget} as training data.
At the training stage, the methods which are trained with video sequences do not only predict scene depths, but also estimate the camera poses, while the methods which are trained with stereo image pairs generally predict the pixel disparities between stereo pairs.
Regardless of the types of training data, most of these methods focus on learning scene depths by introducing a continuous depth constraint (CDC)~\cite{Garg2016Unsupervised, Zhou2017Unsupervised, Poggi2018Learning, Godard2019Digging, Watson2019Self, Guizilini20203d}, and recently, a few methods employ a discrete depth constraint (DDC) for pursuing scene depths~\cite{Gonzalezbello2020Forget, Gonzalez2021Plade}.  
It is noted that in spite of rapid development for self-supervised monocular depth estimation, the following two problems still remain:
(1) What are the advantage and disadvantage of both the CDC and DDC?
(2) How to utilize the CDC and DDC more effectively to learn scene depth maps, particularly for occluded regions?

Addressing the two problems, we firstly empirically give an analysis on the effects of the CDC and DDC by utilizing two typical architectures, and we find that each of the two constraints has its own advantage and disadvantage.
Then inspired by this analysis, a novel network for self-supervised monocular depth estimation is proposed, which learns an Occlusion-aware Coarse-to-Fine Depth map, called OCFD-Net.
The OCFD-Net is trained with stereo image pairs.
It uses a DDC for learning a coarse-level depth map and a CDC for learning a scene depth residual, and then it outputs a fine-level depth map by integrating the obtained coarse-level depth map with the scene depth residual.
In addition, we explore an occlusion-aware module under the proposed network, in order to strengthen the obtained fine-level depth map for resisting occlusions.

In sum, our main contributions include:

(1) We empirically analyze the effects of the CDC and DDC, finding that a relatively higher prediction accuracy could be achieved by imposing the DDC, while a relatively smoother depth map could be obtained by imposing the CDC.This analysis could not only contribute to a better understanding of the depth constraints, but also give new insights into the design strategies for self-supervised monocular depth estimation.

(2) We explore an occlusion-aware module, which is able to alleviate the negative influence of occluded regions on self-supervised monocular depth estimation.

(3) Based on the aforementioned analysis on the effects of both CDC and DDC as well as the explored occlusion-aware module, we propose the OCFD-Net.
It achieves better performances on the KITTI~\cite{Geiger2012We} and Make3D~\cite{Saxena2008Make3d} datasets than the comparative state-of-the-art methods in most cases as demonstrated in Section~\ref{sec:experiments}.

\section{Related Work}
\label{sec:related}
In this section, we review the self-supervised monocular depth estimation methods trained with monocular video sequences and stereo image pairs respectively.

\subsection{Self-supervised monocular training}
The existing methods which are trained with monocular video sequences simultaneously predict the scene depths and estimate the camera poses~\cite{Zhou2017Unsupervised, Wang2018Learning, Johnston2020Self-supervised, Guizilini20203d, Yang2018Unsupervised, Mahjourian2018Unsupervised, Godard2019Digging, Almalioglu2019Ganvo, Zhao2020Masked, Shu2020Feature-metric, Casser2019Depth, Guizilini2020Semantically-guided, Yin2018Geonet, Chen2019Self}.
Zhou et al.~\cite{Zhou2017Unsupervised} proposed an end-to-end approach comprised of two separate networks for predicting depths and camera poses. 
Godard et al.~\cite{Godard2019Digging} proposed the per-pixel minimum reprojection loss, the auto-mask loss, and the full-resolution sampling for self-supervised monocular depth estimation.
Guizilini et al.~\cite{Guizilini20203d} re-implemented upsample and downsample operations by 3D convolutions to preserve image details for depth predictions.
Casser et al.~\cite{Casser2019Depth} used instance segmentation maps to help model the object motions for handling the non-rigid scene problem.
Additionally, the frameworks which jointly learnt depth, optical flow, and camera pose in a self-supervised manner were investigated in~\cite{Yin2018Geonet, Chen2019Self}.

\subsection{Self-supervised stereo training}
\label{sec:related-stereo}
Unlike the methods trained with monocular video sequences, the existing methods which are trained with stereo image pairs generally estimate scene depths by predicting the disparities between stereo pairs~\cite{Garg2016Unsupervised, Godard2017Unsupervised, Wong2019Bilateral, Poggi2018Learning, Pilzer2019Refine, Tosi2019Learning, Watson2019Self, Chen2019Towards, Zhu2020The}.
Garg et al.~\cite{Garg2016Unsupervised} proposed a pioneering method, which reconstructed one image of a stereo pair with the other image using the predicted depths at its training stage.
Godard et al.~\cite{Godard2017Unsupervised} presented a left-right disparity consistency loss to improve the robustness of the proposed method.
To handle the occlusion problem, Poggi et al.~\cite{Poggi2018Learning} proposed the 3Net which was trained in a trinocular domain, while different types of occlusion masks were proposed in~\cite{Zhu2020The, Wong2019Bilateral} for indicating the occlusion regions.
Additionally, several methods used extra supervision information (e.g. disparities generated with Semi Global Matching~\cite{Watson2019Self,Tosi2019Learning, Zhu2020The}, semantic segmentation labels ~\cite{Zhu2020The, Chen2019Towards}) to improve the performance of self-supervised monocular depth estimation.
It is noted that all the aforementioned methods employed a continuous depth constraint (CDC) for depth estimation at their training stage, assuming that the disparity of each pixel is a continuous variable determined by the visual consistency between the input training stereo images.

Unlike the above methods that utilized the CDC, a few methods~\cite{Gonzalezbello2020Forget, Gonzalez2021Plade} employed a discrete depth constraint (DDC) at their training stage, assuming that the depth of each pixel is inversely proportional to a weighted sum of a set of discrete disparities determined by the visual consistency between the input training stereo images. 
Gonzalez and Kim~\cite{Gonzalezbello2020Forget} proposed a self-supervised monocular depth estimation network by utilizing the DDC with a mirrored exponential disparity discretization.

\section{Methodology}
\label{sec:method}
In this section, we propose the OCFD-Net for self-supervised monocular depth estimation.
Firstly, we give an empirical analysis on the effects of the continuous depth constraint (CDC) and discrete depth constraint (DDC) used in literature.
Then according to this analysis, we describe the proposed OCFD-Net in detail.

\subsection{Effects of CDC and DDC}
\label{sec:constraints}
As discussed in Section~\ref{sec:related}, most of the existing methods which are trained with stereo images learn depths by introducing either the CDC~\cite{Garg2016Unsupervised, Zhou2017Unsupervised, Godard2017Unsupervised, Poggi2018Learning, Pilzer2019Refine, Godard2019Digging, Watson2019Self, Guizilini20203d} or the DDC~\cite{Gonzalezbello2020Forget, Gonzalez2021Plade}.
However, it is still unclear what are the advantage and disadvantage of the CDC in comparison to the DDC.
Addressing this issue, we investigate the effects of the two depth constraints empirically here.

Specifically, under each of the two depth constraints, we evaluate the following two typical backbone architectures which are used in many existing self-supervised monocular depth estimation methods (rather than these original methods) on the KITTI dataset~\cite{Geiger2012We} with the raw Eigen splits~\cite{Eigen2014Depth}, in order to concentrate on the two constraints and simultaneously avoid possible disturbances of other modules involved in these original methods:

\textbf{FAL-Arc}: It has a 21-layer convolutional architecture as used in the DDC-based FAL-Net~\cite{Gonzalezbello2020Forget} and PLADE-Net~\cite{Gonzalez2021Plade}. 

\textbf{Res-Arc}: It has a ResNet-50~\cite{He2016Deep} based architecture as used in many CDC-based works, e.g.~\cite{Godard2017Unsupervised, Poggi2018Learning, Godard2019Digging, Watson2019Self, Zhu2020The}. 

The corresponding results are reported in Table~\ref{tab:analyze} (the metrics are introduced in Section~\ref{sec:experiments}).
As is seen, both the two architectures with DDC outperform those with CDC under all the metrics, demonstrating that DDC is probably more helpful for boosting the performances of the existing methods.

\begin{table*}
  \centering
  \small
  \caption{Quantitative comparison of FAL-Arc and Res-Arc with CDC and DDC on the raw KITTI Eigen test set~\cite{Eigen2014Depth}.
  $\downarrow/\uparrow$ denotes that lower / higher is better.}
  \begin{tabular}{|lc|cccc|ccc|}
  \hline
  \multicolumn{1}{|c}{Arc.} &
    Constraint &
    Abs Rel $\downarrow$ &
    Sq Rel $\downarrow$ &
    RMSE $\downarrow$ &
    logRMSE $\downarrow$ &
    A1 $\uparrow$ &
    A2 $\uparrow$ &
    A3 $\uparrow$\\ \hline
  FAL-Arc   & CDC & 0.135 & 0.915 & 4.705 & 0.212 & 0.834 & 0.937 & 0.975 \\
  FAL-Arc   & DDC   & 0.104 & 0.683 & 4.363 & 0.190 & 0.877 & 0.960 & 0.981 \\\hline
  Res-Arc & CDC & 0.126 & 0.912 & 4.592 & 0.204 & 0.851 & 0.944 & 0.977 \\
  Res-Arc & DDC   & 0.112 & 0.685 & 4.298 & 0.193 & 0.871 & 0.957 & 0.981 \\\hline
  \end{tabular}
  \label{tab:analyze}
\end{table*}

\begin{figure}
  \centering
    \begin{minipage}{0.27\linewidth}
      \small
      \leftline{Input image}
      \leftline{(Local regions)}
      \vspace{21pt}
      \leftline{FAL-Arc + CDC}
      \vspace{22pt}
      \leftline{FAL-Arc + DDC}
      \vspace{22pt}
      \leftline{Res-Arc + CDC}
      \vspace{22pt}
      \leftline{Res-Arc + DDC}
      \vspace{3pt}
    \end{minipage}
    \begin{minipage}{0.72\linewidth}
      \begin{minipage}{0.61\linewidth}
        \centerline{\includegraphics[width=\textwidth]{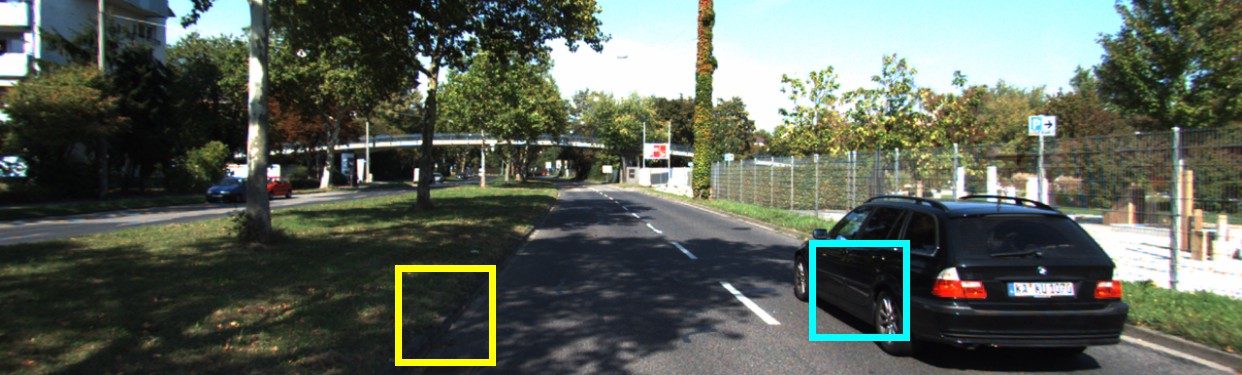}}
        \centerline{\includegraphics[width=\textwidth]{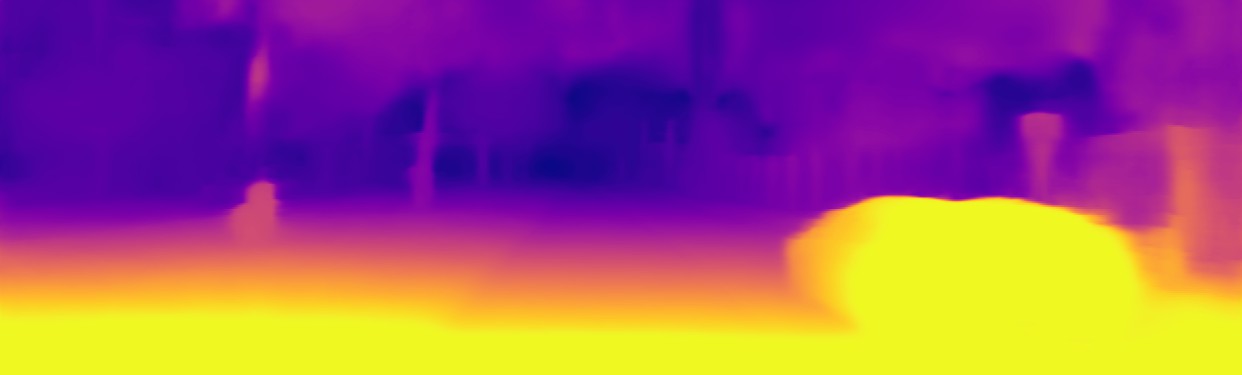}}
        \centerline{\includegraphics[width=\textwidth]{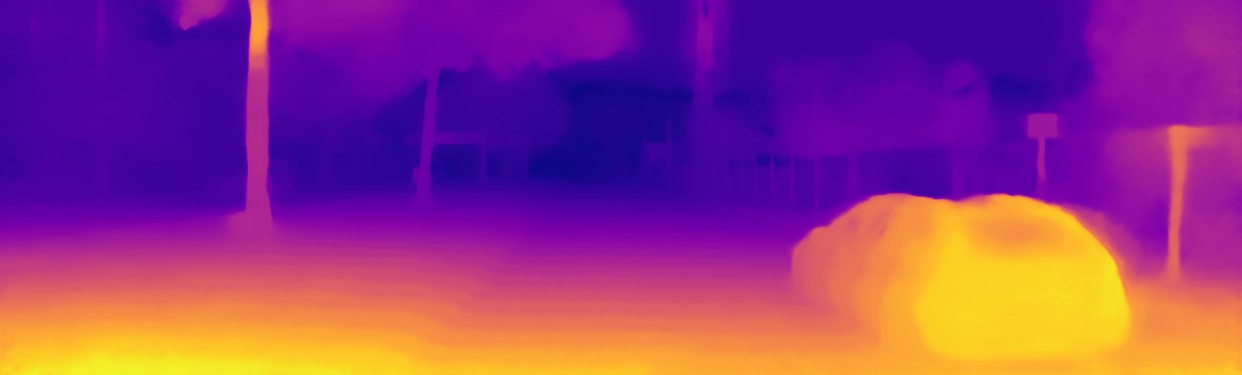}}
        \centerline{\includegraphics[width=\textwidth]{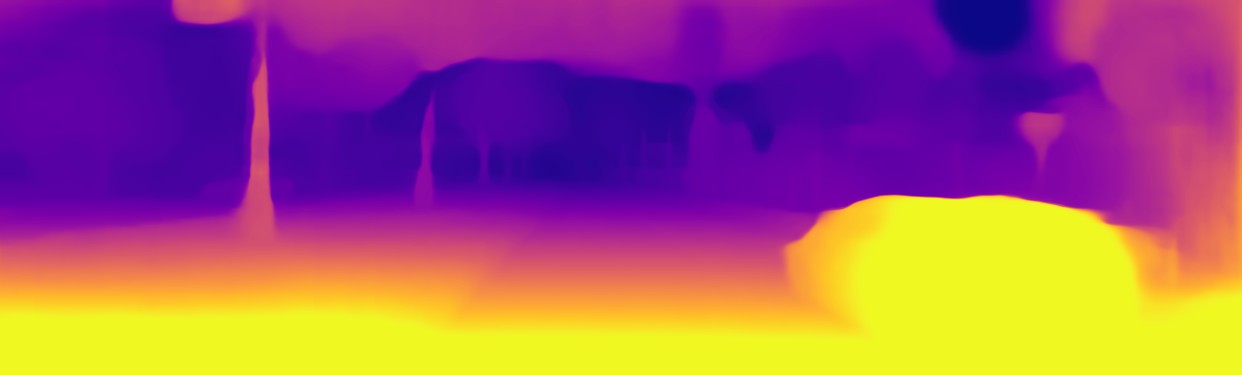}}
        \centerline{\includegraphics[width=\textwidth]{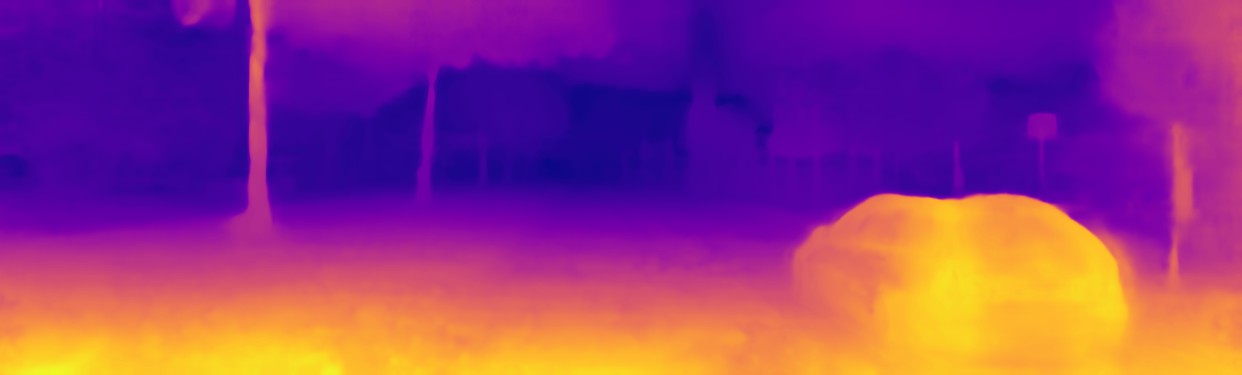}}
      \end{minipage}
      \begin{minipage}{0.183\linewidth}
        \centerline{\includegraphics[width=\textwidth]{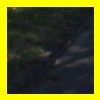}}
        \centerline{\includegraphics[width=\textwidth]{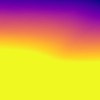}}
        \centerline{\includegraphics[width=\textwidth]{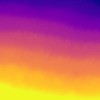}}
        \centerline{\includegraphics[width=\textwidth]{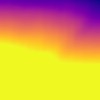}}
        \centerline{\includegraphics[width=\textwidth]{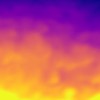}}
      \end{minipage}
      \begin{minipage}{0.183\linewidth}
        \centerline{\includegraphics[width=\textwidth]{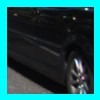}}
        \centerline{\includegraphics[width=\textwidth]{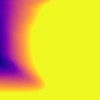}}
        \centerline{\includegraphics[width=\textwidth]{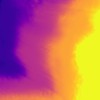}}
        \centerline{\includegraphics[width=\textwidth]{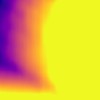}}
        \centerline{\includegraphics[width=\textwidth]{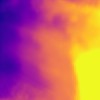}}
      \end{minipage}
    \end{minipage}
    \caption{Visualization results of FAL-Arc and Res-Arc with CDC and DDC on KITTI~\cite{Geiger2012We}.
    The left column shows the depth maps predicted by the architectures with CDC and DDC.
    The middle and right columns show the depth maps corresponding to two local regions from the input image (the yellow and cyan boxes) for comparing the performances of CDC and DDC in detail.
    Since the depth range in the local regions is much smaller than that in the input image, the depth maps in the middle and right columns are re-normalized so that they could be compared more clearly.}
    
    \label{fig:analyze}
  \end{figure}

In addition, the visualization results of the estimated depth maps by the two architectures with the two depth constraints are shown in Figure~\ref{fig:analyze}.
Two points are revealed from this figure:
(1) The depth maps estimated by the two architectures with DDC preserve more detailed information than those with CDC (e.g. the estimated depths on the cylindrical objects by the two architectures with DDC are visually more delicate than those by the two architectures with CDC as shown in the left column of Figure~\ref{fig:analyze})
(2) The estimated depth maps (particularly for flat regions such as the ground and car surfaces in the middle and right columns of Figure~\ref{fig:analyze}) by the two architectures with CDC are relatively smoother, while the estimated depth maps by the two architectures with DDC are relatively sharper.
More visualization results could be found in the supplemental material.

In sum, as noted from both the quantitative results in Table~\ref{tab:analyze} and the revealed points from Figure~\ref{fig:analyze}, the two constraints have their own advantage and disadvantage: \textbf{DDC is more helpful for preserving more detailed depth information and improving the depth estimation accuracy but fails to achieve a smooth estimation on flat regions, while CDC is helpful for maintaining the smoothness of the estimated depths, but it often achieves a lower depth estimation accuracy than DDC.}
These issues inspire us to propose the following OCFD-Net that does not only take the advantages of the two depth constraints but also alleviate their deficiencies.

\subsection{OCFD-Net}
Here, we propose the OCFD-Net, whose architecture is shown in Figure~\ref{fig:architec}(a).
It is trained with stereo image pairs, and it has a coarse-to-fine depth module, an image reconstruction module, and an occlusion-aware module. 
Considering the relative advantage of DDC for improving estimation accuracy, the coarse-to-fine depth module learns a coarse-level depth map under the imposed DDC by the image reconstruction module, which could provide a relatively accurate initial estimation of depth.
And considering the relative advantage of CDC for maintaining the smoothness of the estimated depths, the coarse-to-fine depth module learns a scene depth residual for providing a smooth depth compensation under the imposed CDC by the image reconstruction module, then it learns a fine-level depth map by integrating the obtained coarse-level depth map with the scene depth residual. 
Additionally, the occlusion-aware module is designed for alleviating the negative influence of occluded regions.
We introduce the three modules and the used loss function as follows:

\subsubsection{Coarse-to-fine depth module}
\begin{figure*}
  \small
  \centering
  \centerline{\includegraphics[width=0.85\textwidth]{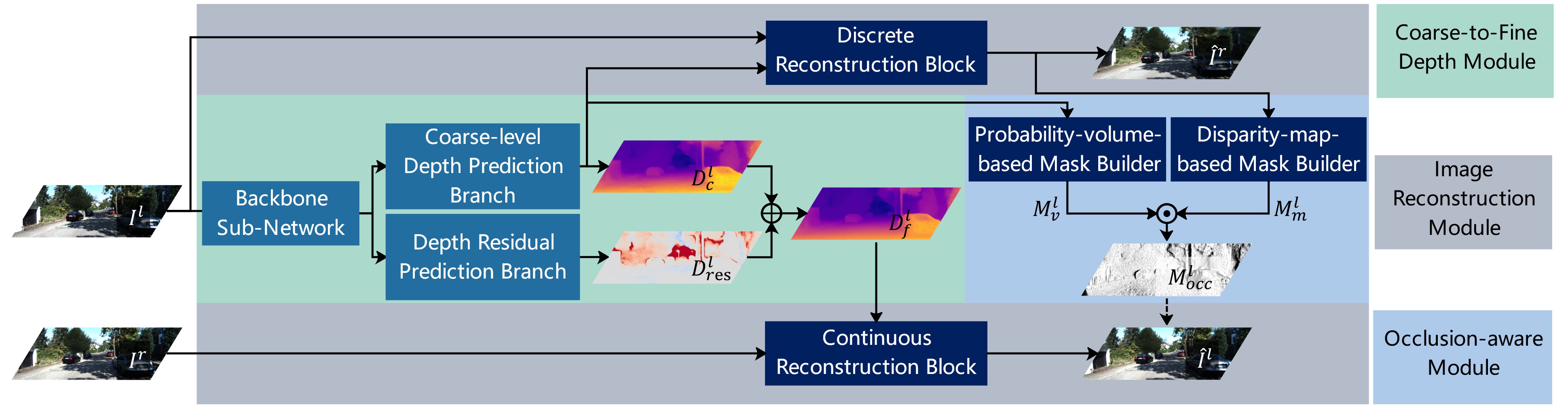}}
  \centerline{(a)}
  \centerline{\includegraphics[width=0.75\textwidth]{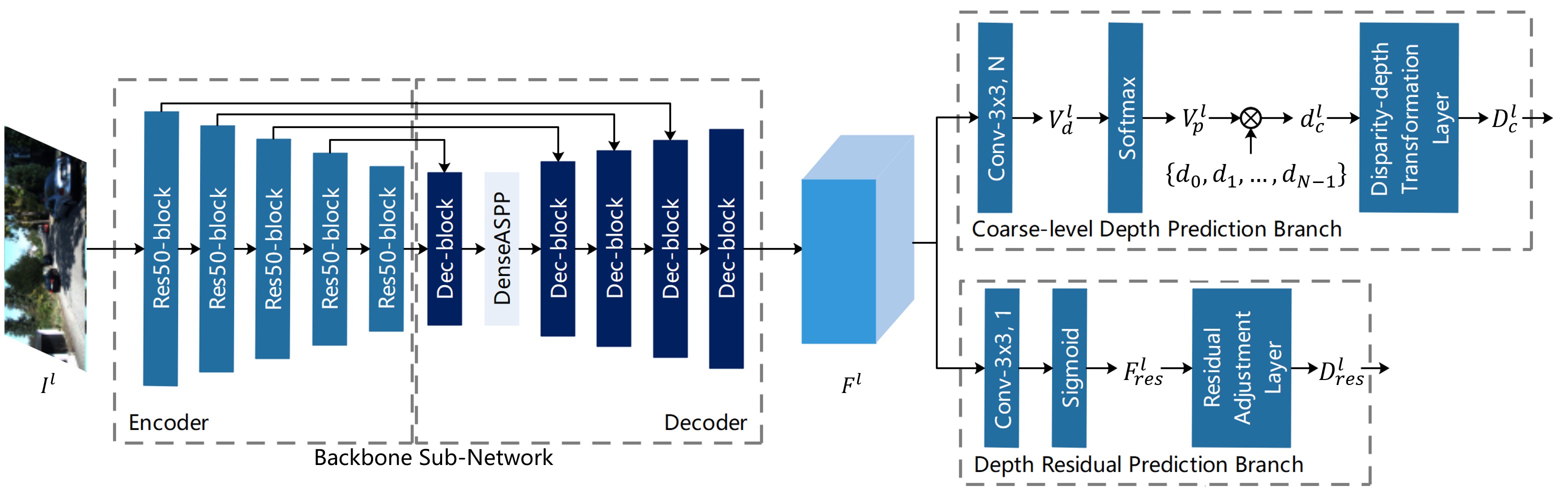}}
  \centerline{(b)}
  \caption{Architectures of OCFD-Net and its coarse-to-fine depth module.
           (a) Architecture of OCFD-Net. It has a coarse-to-fine depth module, an image reconstruction module, and an occlusion-aware module.
           `$\oplus$' denotes the element-wise addition and `$\odot$' denotes the element-wise multiplication.
           (b) Architecture of the coarse-to-fine depth module. `Conv-3x3, N' denotes a $3 \times 3$ convolutional layer with $N$ channels, and `$\otimes$' denotes the scalar multiplication.
          }
\label{fig:architec}
\end{figure*}

The coarse-to-fine depth module is to learn a fine-level depth map by simultaneously learning a coarse-level depth map and a scene depth residual from an input scene image.
It has a backbone sub-network for feature extraction, a coarse-level depth prediction branch, and a depth residual prediction branch, as shown in Figure~\ref{fig:architec}(b).

\paragraph{Backbone sub-network}
It employs an encoder-decoder architecture for extracting a visual feature $F^l \in \mathbb{R}^{W\times H \times C_f}$ from the input left image $I^l \in \mathbb{R}^{W\times H \times 3}$, where $\{W, H\}$ are the height and width of the image, and $C_f$ is the number of feature channels.
Similar to~\cite{Godard2017Unsupervised, Poggi2018Learning, Godard2019Digging}, this backbone sub-network simply uses the first five blocks of ResNet50~\cite{He2016Deep} as its encoder, and the 5-block decoder designed in~\cite{Godard2019Digging} as its decoder.
In addition, a DenseASPP module~\cite{Yang2018Denseaspp} with dilation rates $r\in\{3, 6, 12, 18, 24\}$ is inserted between the first two blocks of the decoder to extract a multi-scale feature.

\paragraph{Coarse-level depth prediction branch}
It uses the feature $F^l$ extracted from the backbone sub-network as its input, and pursues a coarse-level depth map $D_c^l$ for the left image by imposing the DDC.
This branch consists of a $3 \times 3$ convolutional layer with $N$ channels, a softmax operation, and a disparity-depth transformation layer.
The convolutional layer maps the feature $F^l$ to a density volume $V_d^l=[V_{dn}^l]_{n=0}^{N-1}$, where $V_{dn}^l\in \mathbb{R}^{W\times H \times 1}$ is the $n^{th}$ channel of $V_d^l$ and `$[\cdot]$' denotes a concatenation operation along the third dimension.
Then, a probability volume $V_p^l=[V_{pn}^l]_{n=0}^{N-1}$ is obtained by passing $V_d^l$ through the softmax operation along the third dimension.

Given a disparity range $[d_{\min}, d_{\max}]$ where $d_{\min}$ and $d_{\max}$ are the predefined minimum and maximum disparities respectively, a set of discrete disparity values $\{d_n\}$ is generated by the mirrored exponential disparity discretization~\cite{Gonzalezbello2020Forget} as:
\begin{equation}
    d_n = d_{\max}\left(\frac{d_{\min}}{d_{\max}}\right)^{ \frac{n}{N-1}}, \quad n = 0, 1, ..., N-1 \quad.
\end{equation}
According to the described DDC in Section~\ref{sec:related-stereo} as well as the obtained probability volume $V_p^l$ above, a disparity map $d_c^l$ for the left image is obtained by calculating a weighted sum of $\{d_n\}_{n=0}^{N-1}$ with the corresponding weights $\{V_{pn}^l\}_{n=0}^{N-1}$:
\begin{equation}
    d^l_{c} = \sum_{n=0}^{N-1}{V_{pn}^l d_n}\quad.
\end{equation}
According to the obtained $d_c^l$, our coarse-level depth map $D_c^l$ for the left image is calculated via the following operation at the disparity-depth transformation layer:
\begin{equation}
    D_{c}^l = \frac{Bf_x}{d_{c}^l} \quad,
\end{equation}
where $B$ is the baseline length of the stereo pair and $f_x$ is the horizontal focal length of the left camera.

\paragraph{Residual depth prediction branch}
It uses the feature $F^l$ extracted from the backbone sub-network as its input, and outputs a scene depth residual $D_{res}^l\in \mathbb{R}^{W\times H \times 1}$ for the left image by imposing the CDC for refining the coarse-level depth $D_c^l$.
This branch consists of a $3 \times 3$ convolutional layer with 1 channel, a sigmoid operation used as the activation function, and a residual adjustment layer.
A feature residual map $F_{res}^l$ whose elements vary in $[0, 1]$ is firstly calculated by passing the feature $F^l$ through the convolutional layer with the sigmoid activation.
Then, considering that a depth residual should be able to provide an either positive or negative compensation for the coarse-level depth map predicted from the coarse-level depth prediction branch, the feature residual $F_{res}^l$ is transformed into a range $[-0.5w, 0.5w]$ (where $w$ is a preseted compensation parameter) via the following linear transformation at the residual adjustment layer:
\begin{equation}
   D_{res}^l = w(F_{res}^l - 0.5) \quad.
   \label{equ:resw}
\end{equation}

Once the scene depth residual $D_{res}^l$ and the coarse-level depth map $D_c^l$ are obtained, a fine-level depth map $D_f^l$ is obtained as:
\begin{equation}
   D_f^l = D_c^l + D_{res}^l\quad.
\end{equation}

\subsubsection{Image reconstruction module}
The image reconstruction module uses one image from each input stereo pair to reconstruct its partner with the predicted depth maps for network training.
As shown in Figure~\ref{fig:architec}(a), this module contains two parts: a discrete reconstruction block for imposing the DDC and a continuous reconstruction block for imposing the CDC.

\paragraph{Discrete reconstruction block}
It takes the left image $I^l$ and the predicted density volume $V_d^l(=[V_{dn}^l]_{n=0}^{N-1})$ as its input, and it reconstructs the right image under the DDC.
As done in~\cite{Gonzalezbello2020Forget}, the density volume  $\hat{V}^r_d=[\hat{V}_{dn}^{r}]_{n=0}^{N-1}$ for the right view is firstly generated by shifting each channel $V_{dn}^l$  of $V_{d}^l$  with the disparity $d_n$.
Then, $\hat{V}^r_d$ is passed thought a softmax operation along the third dimension to obtain the right-view probability volume $\hat{V}_{p}^{r}=[\hat{V}_{pn}^{r}]_{n=0}^{N-1}$.
According to the DDC, the reconstructed right image $\hat{I}^r$ is obtained by calculating a weighted sum of the shifted $N$ versions of the left image $I^l$ with the corresponding probabilities $\hat{V}_{pn}^{r}$:
\begin{equation}
    \hat{I}^r = \sum_{n=0}^{N-1}{\hat{V}_{pn}^{r} \odot I^l_{n}}\quad,
\end{equation}
where `$\odot$' denotes the element-wise multiplication, and $I^l_{n}$ is the left image shifted with $d_n$.

\paragraph{Continuous reconstruction block}
It takes the right image $I^r$ and the fine-level depth map $D_f^l$ as its input, and it reconstructs the corresponding left image under the CDC.
Specifically, for an arbitrary pixel coordinate $p \in \mathbb{R}^2 $ in the left image, its corresponding coordinate $p'$ in the right image is obtained with the fine-level depth map $D_{f}^l$:
\begin{equation}
    p'=p-\left[\frac{Bf_x}{D_f^l(p)},0\right]^\top
    \quad.
\end{equation}
Accordingly, the reconstructed left image $\hat{I}^l$ is obtained by assigning the RGB value of the right image pixel $p'$ to the pixel $p$ of $\hat{I}^l$.
Please see the supplemental material for more details about the geometric transformations used in this module.

\subsubsection{Occlusion-aware module}
As shown in Figure~\ref{fig:architec}(a), the explored occlusion-aware module contains a probability-volume-based mask builder and a disparity-map-based mask builder for learning two masks $M^l_{v}$ and $M^l_{m}$.
The two masks have the same size as the input images, and each element in them varies from 0 to 1 and indicates the probability of whether the corresponding pixel in the left view image is still visible in the right view.
Then, the occlusion-aware module builds an occlusion mask $M^l_{occ}$ by element-wisely multiplying $M^l_{v}$ with $M^l_{m}$:

\paragraph{Probability-volume-based mask builder}
This builder takes the probability volume $\hat{V}_{p}^{r}$ (obtained by the discrete reconstruction block) as its input, and it builds a probability-volume-based mask $M^l_{v}$ as done in~\cite{Gonzalezbello2020Forget}.
Under the DDC, a cyclic probability volume $\hat{V}_{p}^{r\rightarrow l}$ is obtained by shifting $\hat{V}_{p}^{r}$ back into the left view.
Accordingly, for each pixel that is visible in the left view but invisible in the right view, its corresponding elements in all the channels of $\hat{V}_{p}^{r\rightarrow l}$ should be equal or close to 0 in the ideal or noisy case.
For each pixel that is visible in both the two views, its corresponding element in some one of the $N$ channels of $\hat{V}_{p}^{r\rightarrow l}$ should be much larger than 0. 
Hence, the probability-volume-based mask is defined as:
\begin{equation}
    M^l_{v} = \min\left(\sum_{n=0}^{N-1}{\hat{V}_{pn}^{r\rightarrow l}}, 1\right) \quad.
\end{equation}

\paragraph{Disparity-map-based mask builder}
This builder takes the coarse-level disparity map $d_c^l$ as its input, and it builds a disparity-map-based mask $M^l_{m}$ based on the following observation: for an arbitrary pixel location $p = [p_x,p_y]^\top$ and its horizontal right neighbor $p_i = [p_x+i,p_y]^\top (i=1,2,...,K)$ in the left image, if the corresponding location of $p$ is occluded by that of $p_i$ in the right image, the difference between their disparities $d^l_c(p)$ and $d^l_c(p_i)$ should be close or equal to the difference of their horizontal coordinates~\cite{Zhu2020The}. Hence, this mask builder is formulated as:
\begin{equation}
  M^l_{m}(p)=\min{
    \left(
      \min_{i}
      \left(
        \left|d_{c}^l(p_i) - d_{c}^l(p) - i \right|
      \right)
      ,1
    \right)}\quad.
\end{equation}

\subsubsection{Loss function}
The total loss function for training the OCFD-Net contains the following 4 loss terms:

\paragraph{Coarse-level reconstruction loss $L_{CR}$}
As done in~\cite{Gonzalezbello2020Forget}, it is formulated as a weighted sum of the $L_1$ loss and the perceptual loss~\cite{Johnson2016Perceptual} for reflecting the similarity between the reconstructed right image $\hat{I}_r$ and the input right image $I_r$:
\begin{equation}
    L_{CR}= \left \| \hat{I}^r - I^r \right \|_1 + \alpha_1 \sum_{i=1,2,3}{\left \| f^i_{R18}(\hat{I}^r) - f^i_{R18}(I^r) \right \|}_2 \quad,
\end{equation}
where `$\|\cdot\|_1$' and `$\|\cdot\|_2$' represent the $L_1$ norm and the $L_2$ norm, $f^i_{R18}(\cdot)$ denotes the output of the $i^{th}$ block of ResNet18~\cite{He2016Deep} pretrained on the ImageNet dataset~\cite{Russakovsky2015Imagenet}, and $\alpha_1$ is a tuning parameter.

\paragraph{Fine-level reconstruction loss $L_{FR}$}
It is formulated as a weighted sum of the $L_1$ loss and the structural similarity (SSIM) loss ~\cite{Wang2004Image} for reflecting the photometric difference between the reconstructed left image $\hat{I}_l$ and the input left image $I_l$, with the occlusion mask $M^l_{occ}$ for alleviating the negative influence of occlusions and the edge mask $M^l_{edge}$~\cite{Mahjourian2018Unsupervised} for filtering out the pixels whose reprojected coordinates are out of the image:
\begin{equation}
  \begin{aligned}
    L_{FR} = &
     M^l_{occ} \odot M^l_{edge} \odot\\
     & \left(\alpha_2 \left \| \hat{I}^l -  I^l \right \|_1
    +(1 - \alpha_2) {\rm SSIM}{(\hat{I}^l, I^l)} \right) \quad,
  \end{aligned}
  \label{equ:cr}
\end{equation}
where $\alpha_2$ is a balance parameter.

\paragraph{Coarse-level smoothness loss $L_{CS}$ and fine-level smoothness loss $L_{FS}$}
As done in~\cite{Godard2017Unsupervised, Gonzalezbello2020Forget}, we adopt the edge-aware smoothness loss to constrain the continuity of both the coarse-level and fine-level disparity maps.
The coarse-level smoothness loss is formulated as:
\begin{equation}
    L_{CS}= \left \| \partial_x d_{c}^l \right \|_1 e^{-\beta_{c} \left \| \partial_x I^l \right \|_1}
    + \left \| \partial_y d_{c}^l \right \|_1 e^{-\beta_{c} \left \| \partial_y I^l \right \|_1}\ \quad,
\end{equation}
where `$\partial_x$', `$\partial_y$' are the differential operators in the horizontal and vertical directions respectively, and $\beta_c$ is a parameter for adjusting the degree of edge preservation.
The fine-level smoothness loss uses an additional weight matrix $W = 1 + (1 - M^l_{occ} \odot M^l_{edge})$ to enforce the smoothness in occluded and edge regions as:
\begin{equation}
    L_{FS}= W \odot \left( \left \| \partial_x d_{f}^l \right \|_1 e^{-\beta_{f} \left \| \partial_x I^l \right \|_1}
    + \left \| \partial_y d_{f}^l \right \|_1 e^{-\beta_{f} \left \| \partial_y I^l \right \|_1} \right) \quad,
    \label{equ:df}
\end{equation}
where $\beta_f$ is the edge preservation parameter.

\begin{table*}
  \centering
  \small
  \caption{Quantitative comparison on both the raw and improved KITTI Eigen test sets.
  The best and the second best results are in \textbf{bold} and \underline{underlined} in each metric.}
  \begin{tabular}{|lccc|cccc|ccc|}
  \hline
  \multicolumn{1}{|c}{Method} &
    PP. &
    Data. &
    Sup. &
    Abs Rel $\downarrow$ &
    Sq Rel $\downarrow$ &
    RMSE $\downarrow$ &
    logRMSE $\downarrow$ &
    A1 $\uparrow$&
    A2 $\uparrow$&
    A3 $\uparrow$\\ \hline
    \multicolumn{11}{|c|}{Raw Eigen test set~\cite{Eigen2014Depth}} \\\hline
  Zhao et al.~\cite{Zhao2020Masked}       &   & K    & M      & 0.139 & 1.034 & 5.264 & 0.214 & 0.821 & 0.942 & 0.978 \\
  DualNet~\cite{Zhou2019Unsupervised} &   & K    & M      & 0.121 & 0.837 & 4.945 & 0.197 & 0.853 & 0.955 & 0.982 \\
  PackNet~\cite{Guizilini20203d} &   & K    & M      & 0.107 & 0.802 & 4.538 & 0.186 & 0.889 & 0.962 & 0.981 \\
  Johnston and Carneiro~\cite{Johnston2020Self-supervised}   &   & K    & M      & 0.106 & 0.861 & 4.699 & 0.185 & 0.889 & 0.962 & 0.982 \\
  Shu et al.~\cite{Shu2020Feature-metric}       &   & K    & M      & 0.104 & 0.729 & 4.481 & 0.179 & 0.893 & 0.965 & \underline{0.984} \\
  3Net~\cite{Poggi2018Learning}     & \checkmark & K    & S      & 0.126 & 0.961 & 5.205 & 0.220  & 0.835 & 0.941 & 0.974 \\
  Peng et al.~\cite{Peng2020A}       &  \checkmark & K    & S      & 0.107 & 0.908 & 4.877 & 0.202 & 0.862 & 0.945 & 0.975 \\
  monoResMatch~\cite{Tosi2019Learning}     & \checkmark & K    & S(d)   & 0.111 & 0.867 & 4.714 & 0.199 & 0.864 & 0.954 & 0.979 \\
  Monodepth2~\cite{Godard2019Digging}       &   & K    & S      & 0.107 & 0.849 & 4.764 & 0.201 & 0.874 & 0.953 & 0.977 \\
  Pilzer et al.~\cite{Pilzer2019Refine}     &   & K    & S      & 0.098 & 0.831 & 4.656 & 0.202 & 0.882 & 0.948 & 0.973 \\
  DepthHints~\cite{Watson2019Self}       & \checkmark & K    & S(d)   & 0.096 & 0.710  & 4.393 & 0.185 & 0.890  & 0.962 & 0.981 \\
  FAL-Net~\cite{Gonzalezbello2020Forget} & \checkmark & K    & S      & 0.093    & \underline{0.564} & \textbf{3.973} & \underline{0.174}    & 0.898 & \textbf{0.967} & \textbf{0.985} \\
  Zhu et al.~\cite{Zhu2020The}       & \checkmark & K    & S(s,d) & 0.091 & 0.646    & 4.244 & 0.177 & 0.898 & \underline{0.966}    & 0.983 \\
  PLADE-Net~\cite{Gonzalez2021Plade} & \checkmark & K & S  & \textbf{0.089} & 0.590 & 4.008 & \textbf{0.172} & 0.900 & \textbf{0.967} & \textbf{0.985} \\
  OCFD-Net (our)     &   & K    & S      & 0.091 & 0.576 & 4.036    & \underline{0.174} & \underline{0.901} & \textbf{0.967} & \underline{0.984}    \\
  OCFD-Net (our)      & \checkmark & K    & S      & \underline{0.090} & \textbf{0.563} & \underline{4.005}    & \textbf{0.172} & \textbf{0.903} & \textbf{0.967} & \underline{0.984} \\
  \hline\hline
  Zhao et al.~\cite{Zhao2020Masked}       &   & CS+K    & M      & 0.135 & 1.026 & 5.153 & 0.210 & 0.833 & 0.945 & 0.979 \\
  PackNet~\cite{Guizilini20203d}  &   & CS+K & M      & 0.104 & 0.758 & 4.386 & 0.182 & 0.895 & 0.964 & 0.982 \\
  Guizilini et al.~\cite{Guizilini2020Semantically-guided}  &   & CS+K    & M(s)   & 0.100   & 0.761 & 4.270  & 0.175 & 0.902 & 0.965 & 0.982 \\
  3Net~\cite{Poggi2018Learning}    & \checkmark & CS+K & S      & 0.111 & 0.849 & 4.822 & 0.202 & 0.865 & 0.952 & 0.978 \\
  Peng et al.~\cite{Peng2020A}       &  \checkmark & CS+K    & S      & 0.100 & 0.767 & 4.455 & 0.189 & 0.881 & 0.956 & 0.980 \\
  monoResMatch~\cite{Tosi2019Learning}     & \checkmark & CS+K & S(d)   & 0.096 & 0.673 & 4.351 & 0.184 & 0.890  & 0.961 & 0.981    \\
  FAL-Net~\cite{Gonzalezbello2020Forget} & \checkmark & CS+K & S      & 0.088 & \underline{0.547}    & 4.004    & 0.175    & 0.898    & 0.966    & \underline{0.984} \\
  PLADE-Net~\cite{Gonzalez2021Plade} & \checkmark & CS+K & S & \underline{0.087} & 0.550 & \textbf{3.837} & \textbf{0.167} & \underline{0.908} & \textbf{0.970} & \textbf{0.985} \\
  OCFD-Net (our)      &   & CS+K & S      & 0.088 & 0.554 & 3.944 & 0.171 & 0.906 & 0.967 & \underline{0.984} \\
  OCFD-Net (our)      & \checkmark & CS+K & S      & \textbf{0.086} & \textbf{0.536} & \underline{3.889} & \underline{0.169} & \textbf{0.909} & \underline{0.969} & \textbf{0.985} \\\hline \hline
  \multicolumn{11}{|c|}{Improved Eigen test set~\cite{Uhrig2017Sparsity}} \\\hline

  PackNet~\cite{Guizilini20203d}    &    & K    & M    & 0.078       & 0.420 & 3.485 & 0.121 & 0.931 & 0.986       & \underline{0.996} \\
  Monodepth2~\cite{Godard2019Digging} & \checkmark  & K    & S    & 0.085       & 0.537 & 3.868 & 0.139 & 0.912 & 0.979       & 0.993 \\
  FAL-Net~\cite{Gonzalezbello2020Forget}    & \checkmark  & K    & S    & 0.071 & 0.281 & 2.912 & 0.108 & 0.943 & 0.991       & \textbf{0.998} \\
  PLADE-Net~\cite{Gonzalez2021Plade} & \checkmark & K & S & \textbf{0.066} & 0.272 & 2.918 & \underline{0.104} & \underline{0.945} & \underline{0.992} & \textbf{0.998} \\
  OCFD-Net (our) &    & K    & S & 0.070 & \underline{0.270}    & \underline{2.821}    & \underline{0.104}    & \underline{0.949}    & \underline{0.992}    & \textbf{0.998} \\
  OCFD-Net (our) & \checkmark  & K    & S & \underline{0.069} & \textbf{0.262} & \textbf{2.785} & \textbf{0.103} & \textbf{0.951} & \textbf{0.993} & \textbf{0.998} \\
  \hline \hline
  PackNet~\cite{Guizilini20203d}    & \multicolumn{1}{l}{} & CS+K & M    & 0.071       & 0.359 & 3.153 & 0.109 & 0.944 & 0.990       & 0.997 \\
  FAL-Net~\cite{Gonzalezbello2020Forget}    & \checkmark  & CS+K & S    & 0.068 & 0.276 & 2.906 & 0.106 & 0.944 & 0.991 & \underline{0.998}    \\
  PLADE-Net~\cite{Gonzalez2021Plade} & \checkmark & CS+K & S & \textbf{0.065} & 0.253 & 2.710 & 0.100 & 0.950 & \underline{0.992} & \underline{0.998} \\
  OCFD-Net (our) & \multicolumn{1}{l}{} & CS+K & S & 0.068    & \underline{0.246}   & \underline{2.669}   & \underline{0.099}    & \underline{0.955}    & \textbf{0.994} & \textbf{0.999} \\
  OCFD-Net (our) & \checkmark  & CS+K & S & \underline{0.066} & \textbf{0.236} & \textbf{2.612} & \textbf{0.096} & \textbf{0.957} & \textbf{0.994} & \textbf{0.999} \\\hline
  \end{tabular}
  \label{tab:res}
\end{table*}

Finally, the total loss is a weighted sum of the above four loss terms, which is formulated as:
\begin{equation}
    L=L_{CR} + \lambda_1 L_{FR} + \lambda_2 L_{CS} +  \lambda_3 L_{FS} \quad,
    \label{equ:total}
\end{equation}
where $\{\lambda_1, \lambda_2, \lambda_3\}$ are three preseted weight parameters.

\section{Experiments}
\label{sec:experiments}

\subsection{Datasets and metrics}
We train OCFD-Net on the KITTI dataset~\cite{Geiger2012We} with the Eigen split~\cite{Eigen2014Depth}, which consists of 22600 stereo image pairs.
Additionally, the Cityscapes dataset~\cite{Cordts2016The}, which consists of 22972 stereo pairs, is used for jointly training OCFD-Net as done in~\cite{Gonzalezbello2020Forget}.
The raw and improved KITTI Eigen test sets~\cite{Eigen2014Depth} are used to evaluate OCFD-Net, which consist of 697 and 652 images respectively.
At both the training and inference stages, the images are resized into the resolution of $1280 \times 384$, while we assume that the intrinsics of all the images are identical.  
We also test OCFD-Net on the Make3D~\cite{Saxena2008Make3d} test set, which includes 134 images. At the inference stage on Make3D, we crop and resize the input images as done in~\cite{Godard2019Digging}.

For the evaluation on the KITTI dataset~\cite{Geiger2012We} (also jointly trained with the Cityscapes dataset~\cite{Geiger2012We}), we use the following metrics as done in~\cite{Zhou2017Unsupervised, Godard2017Unsupervised, Godard2019Digging, Gonzalezbello2020Forget}: Abs Rel, Sq Rel, RMSE, logRMSE, A1~$=\delta < 1.25$, A2~$=\delta < 1.25^2$, and A3~$=\delta < 1.25^3$. 
For the evaluation on Make3D~\cite{Saxena2008Make3d}, we use the following metrics as done in~\cite{Godard2017Unsupervised, Godard2019Digging, Gonzalezbello2020Forget}: Abs Rel, Sq Rel, RMSE, and $log_{10}$.
Please see the supplemental material for more details about the datasets and metrics.

\begin{figure*}
  \small
  \begin{minipage}{0.23\linewidth}
    \begin{minipage}{0.7527\linewidth}
      \small
      \centerline{\includegraphics[width=1\textwidth]{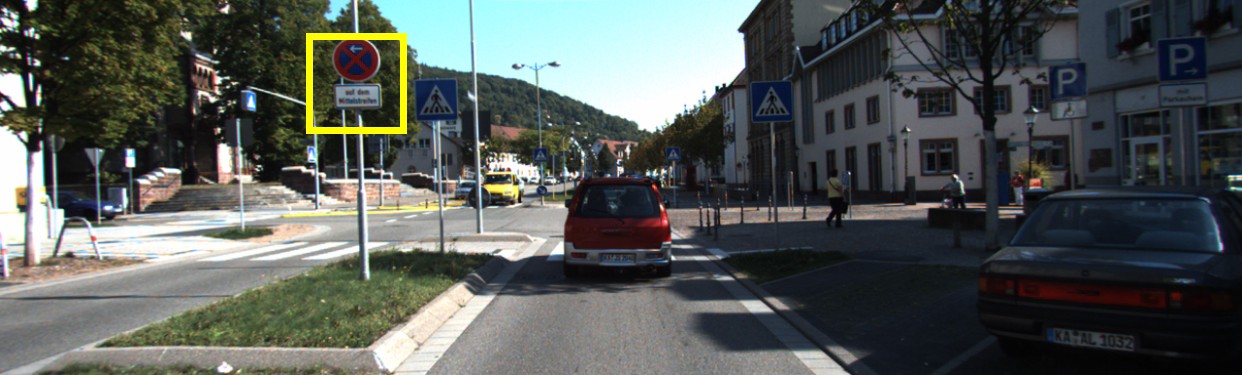}}
      \centerline{\includegraphics[width=1\textwidth]{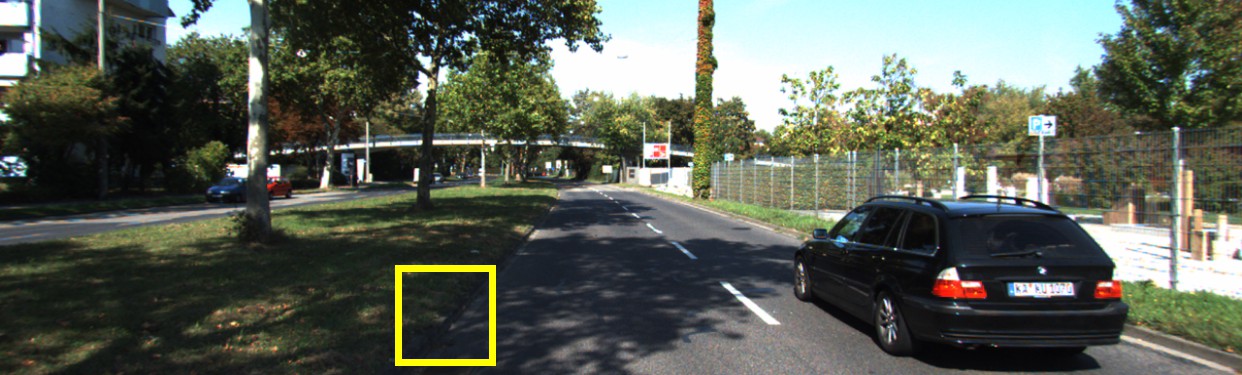}}
      \centerline{\includegraphics[width=1\textwidth]{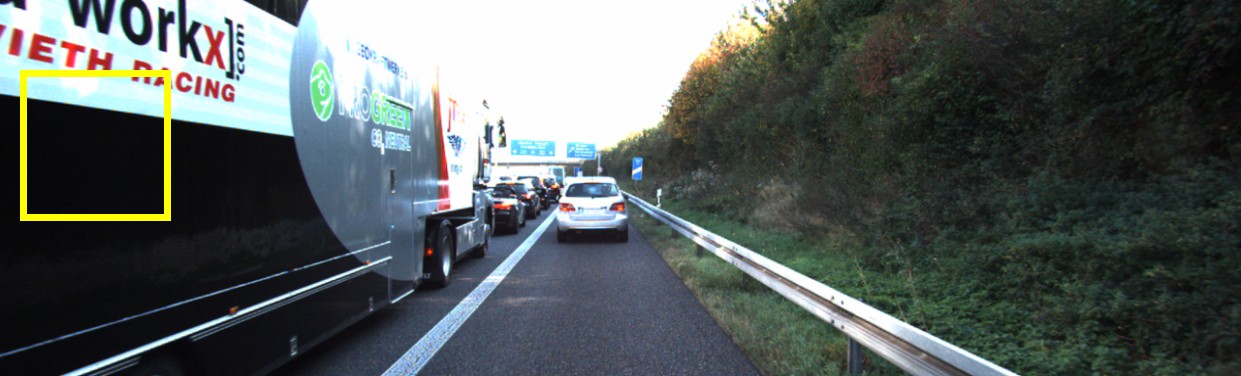}}
    \end{minipage}
    \begin{minipage}{0.2273\linewidth}
      \small
      \centerline{\includegraphics[width=1\textwidth]{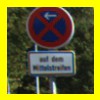}}
      \centerline{\includegraphics[width=1\textwidth]{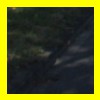}}
      \centerline{\includegraphics[width=1\textwidth]{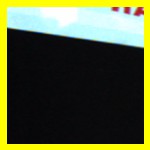}}
    \end{minipage}
    \centerline{Input images (Local regions)}
  \end{minipage}
  \begin{minipage}{0.23\linewidth}
    \begin{minipage}{0.7527\linewidth}
      \small
      \centerline{\includegraphics[width=\textwidth]{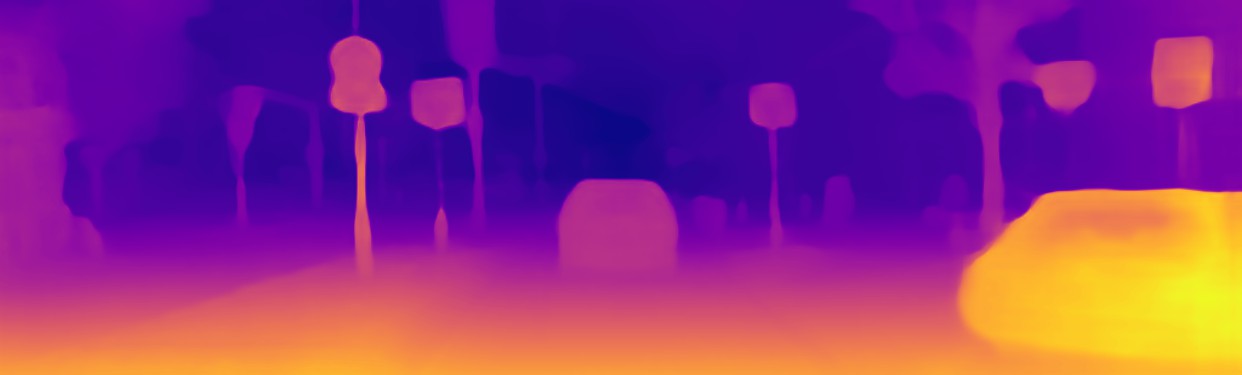}}
      \centerline{\includegraphics[width=\textwidth]{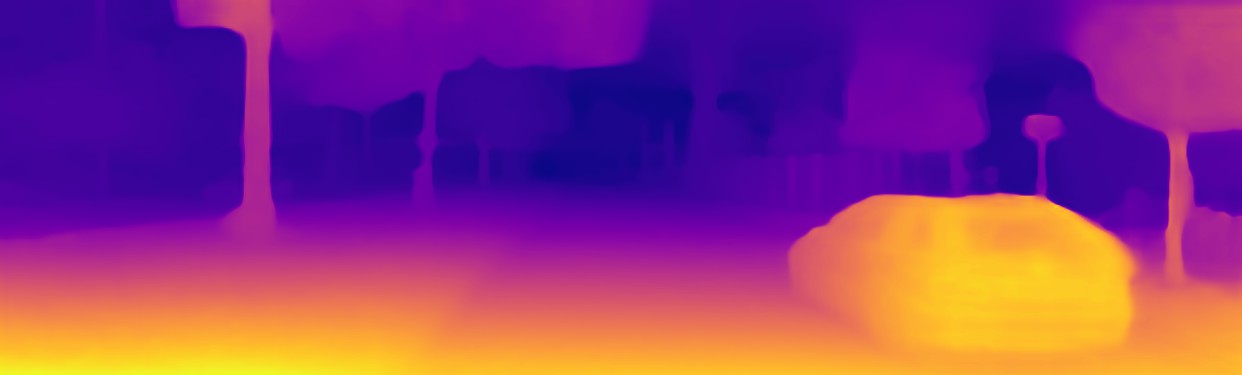}}
      \centerline{\includegraphics[width=\textwidth]{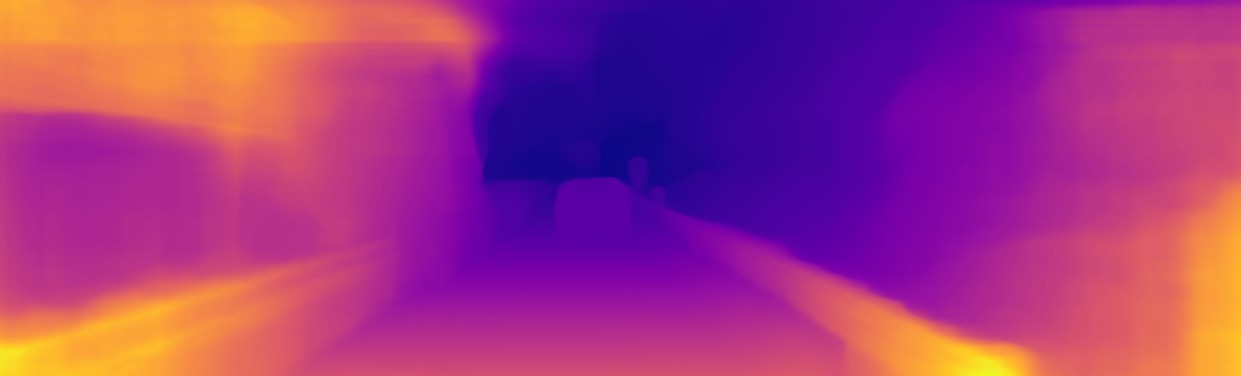}}
    \end{minipage}
    \begin{minipage}{0.2273\linewidth}
      \small
      \centerline{\includegraphics[width=1\textwidth]{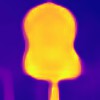}}
      \centerline{\includegraphics[width=1\textwidth]{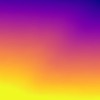}}
      \centerline{\includegraphics[width=1\textwidth]{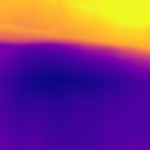}}
    \end{minipage}
    \centerline{DepthHints~\cite{Watson2019Self}}
  \end{minipage}
  \begin{minipage}{0.23\linewidth}
    \begin{minipage}{0.7527\linewidth}
      \small
      \centerline{\includegraphics[width=\textwidth]{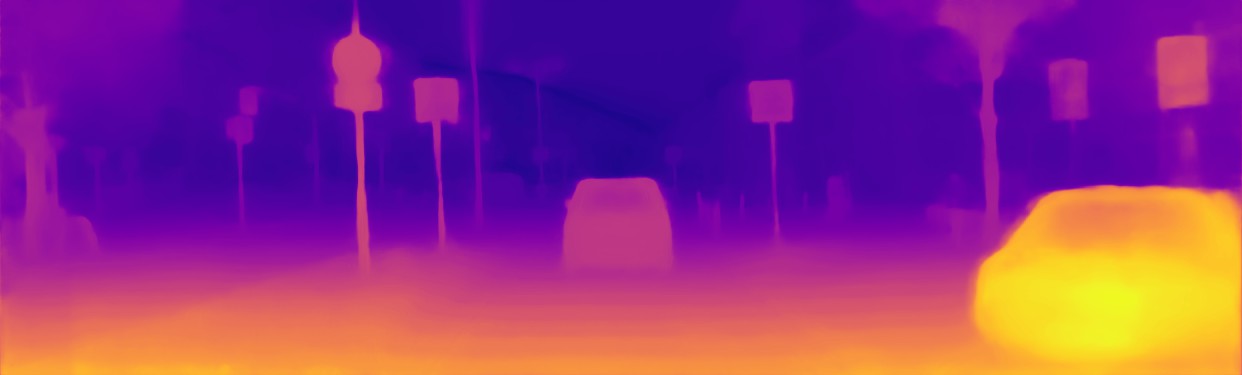}}
      \centerline{\includegraphics[width=\textwidth]{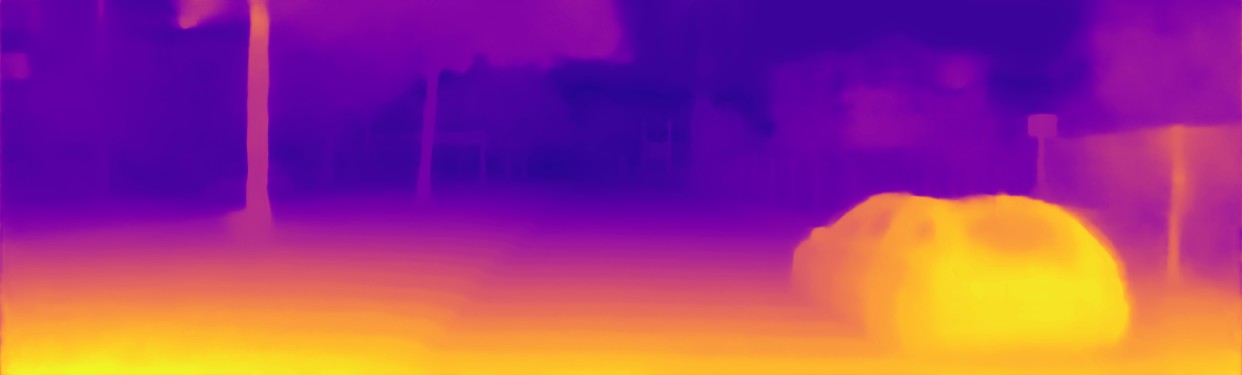}}
      \centerline{\includegraphics[width=\textwidth]{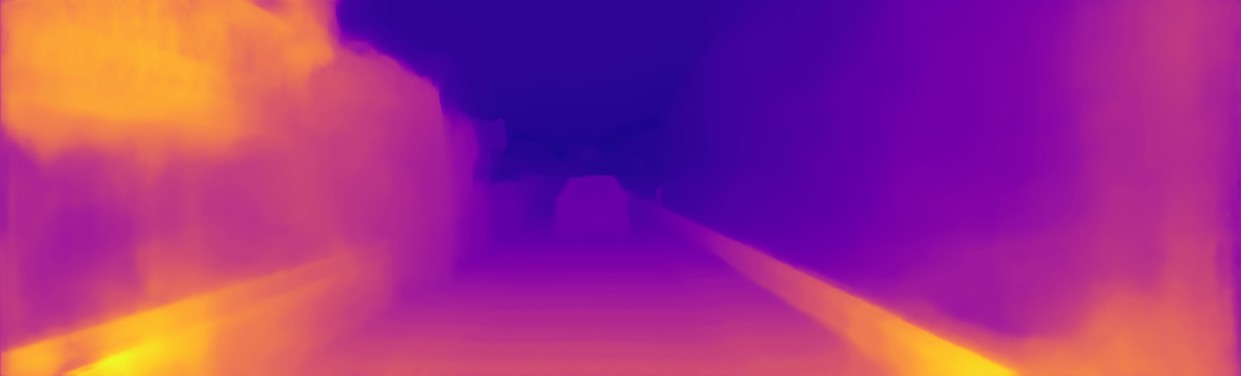}}
    \end{minipage}
    \begin{minipage}{0.2273\linewidth}
      \small
      \centerline{\includegraphics[width=1\textwidth]{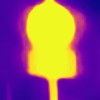}}
      \centerline{\includegraphics[width=1\textwidth]{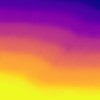}}
      \centerline{\includegraphics[width=1\textwidth]{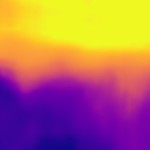}}
    \end{minipage}
    \centerline{FAL-Net~\cite{Gonzalezbello2020Forget}}
  \end{minipage}
  \begin{minipage}{0.23\linewidth}
    \begin{minipage}{0.7527\linewidth}
      \small
      \centerline{\includegraphics[width=\textwidth]{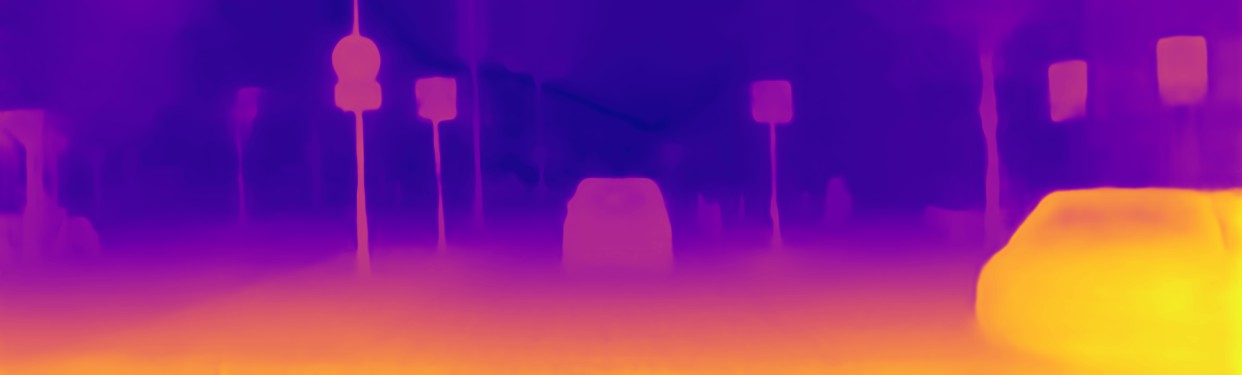}}
      \centerline{\includegraphics[width=\textwidth]{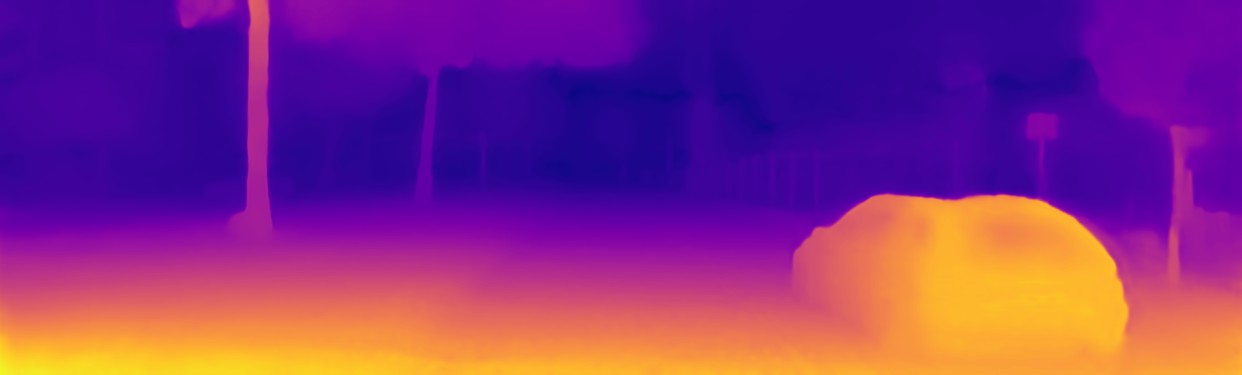}}
      \centerline{\includegraphics[width=\textwidth]{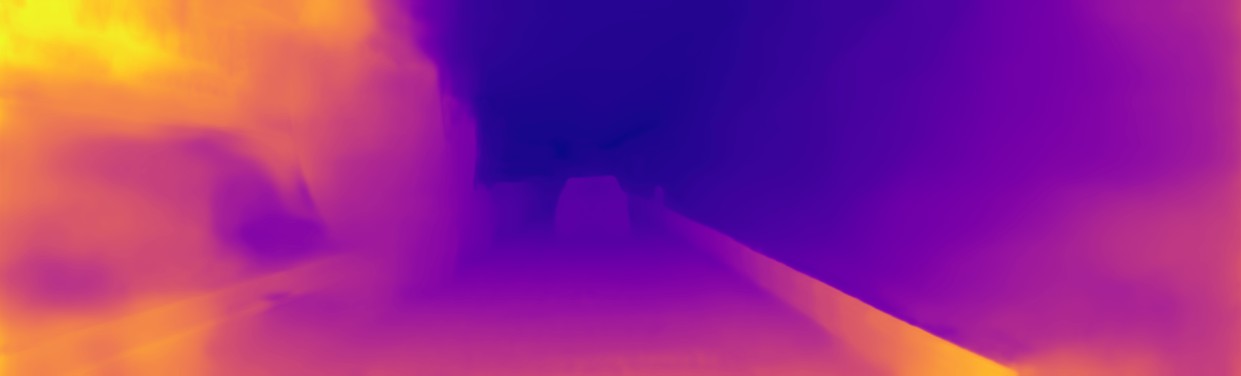}}
    \end{minipage}
    \begin{minipage}{0.2273\linewidth}
      \small
      \centerline{\includegraphics[width=1\textwidth]{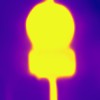}}
      \centerline{\includegraphics[width=1\textwidth]{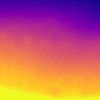}}
      \centerline{\includegraphics[width=1\textwidth]{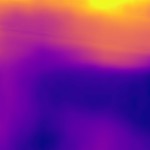}}
    \end{minipage}
    \centerline{OCFD-Net}
  \end{minipage}
  \caption{Visualization results of DepthHints~\cite{Watson2019Self} , FAL-Net~\cite{Gonzalezbello2020Forget}, and our OCFD-Net on KITTI~\cite{Geiger2012We}.
  The images in the even columns are the enlarged versions of the yellow rectangle regions selected from the images in the odd columns, and they are used to show differences of the predicted depth maps by the three methods more clearly (the depth maps in the even
  columns are re-normalized for clearer comparison).}
  \label{fig:compare}
\end{figure*}

\subsection{Implementation details}
We implement the OCFD-Net with PyTorch~\cite{Paszke2019Pytorch}. 
The encoder of the backbone sub-network is pretrained on the ImageNet dataset~\cite{Russakovsky2015Imagenet}.
For disparity discretization, we set the minimum and the maximum disparities to $d_{\min}=2,d_{\max}=300$, and the number of the discrete levels is set to $N=49$.
The weight of the scene depth residual is set to $w=10$, and we set $K=41$ for the disparity-map-based mask.
The weight parameters for the loss function are set to $\lambda_1=1,\lambda_2=0.0008$, and $\lambda_3=0.001$, while we set $\alpha_1=0.1, \alpha_2=0.15, \beta_c=2$ and $\beta_f=1$.
The Adam optimizer~\cite{Kingma2014Adam} with $\beta_1=0.5$ and $\beta_2=0.999$ is used to train the OCFD-Net for 50 epochs with a batch size of 8.
The initial learning rate is firstly set to $10^{-4}$, and is downgraded by half at epoch 30 and 40.
The on-the-fly data augmentations are performed in training, including random resizing (from 0.75 to 1.5) and cropping (640$\times$192), random horizontal flipping, and random color augmentation.

\subsection{Comparative evaluation}
\label{sec:experiments-compare}

\begin{table}
  \centering
  \small
  \renewcommand\tabcolsep{2.0pt}
  \caption{Quantitative comparison on Make3D~\cite{Saxena2008Make3d}.
  Note that all the methods benefit from the median scaling.
  The methods marked with `+PP.'  benefit from the post-processing step.}
  \begin{tabular}{|lc|cccc|}
      \hline 
      \multicolumn{1}{|c}{Method} & Sup. & Abs Rel $\downarrow$  & Sq Rel $\downarrow$   & RMSE $\downarrow$  & $log_{10}$ $\downarrow$        \\\hline
      DDVO~\cite{Wang2018Learning}  & M & 0.387 & 4.720 & 8.090 & 0.204 \\
      Monodepth2~\cite{Godard2019Digging}     & M & 0.322 & 3.589 & 7.417 & 0.163 \\
      Johnston and Carneiro~\cite{Johnston2020Self-supervised}        & M & 0.297 & 2.902 & 7.013 & 0.158     \\
      FAL-Net + PP.~\cite{Gonzalezbello2020Forget}        & S & 0.284 & 2.803 & 6.643 & -     \\
      PLADE-Net + PP.~\cite{Gonzalez2021Plade}  & S & 0.265 & 2.469 & 6.373 & -     \\
      PLADE-Net(CS+K) + PP.~\cite{Gonzalez2021Plade}  & S & \textbf{0.253} & \textbf{2.100} & \underline{6.031} & -     \\
      OCFD-Net       & S & 0.279 & 2.573 & 6.421 & 0.145 \\
      OCFD-Net + PP.        & S & 0.275 & 2.515 & 6.354 & \underline{0.144} \\
      OCFD-Net(CS+K) + PP.        & S & \underline{0.256} & \underline{2.187} & \textbf{5.856} & \textbf{0.135} \\\hline
  \end{tabular}
  \label{tab:m3d}
\end{table}

\begin{figure}
  \centering
  \begin{minipage}{0.20\linewidth}
      \small
      \leftline{Input images}
      \vspace{14.5pt}
      \leftline{OCFD-Net}
      \vspace{14.5pt}
      \leftline{Ground truth}
  \end{minipage}
  \begin{minipage}{0.19\linewidth}
      \small
      \centerline{\includegraphics[width=\textwidth]{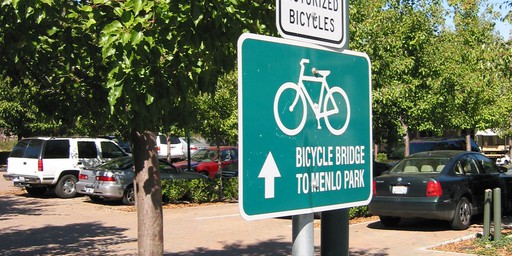}}
      \centerline{\includegraphics[width=\textwidth]{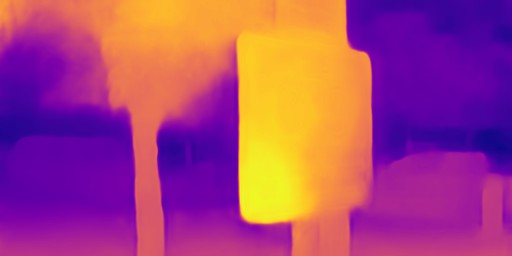}}
      \centerline{\includegraphics[width=\textwidth]{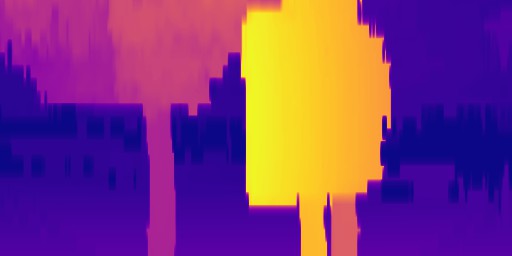}}
  \end{minipage}
  \begin{minipage}{0.19\linewidth}
      \small
      \centerline{\includegraphics[width=\textwidth]{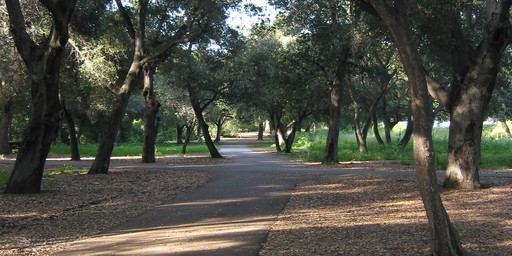}}
      \centerline{\includegraphics[width=\textwidth]{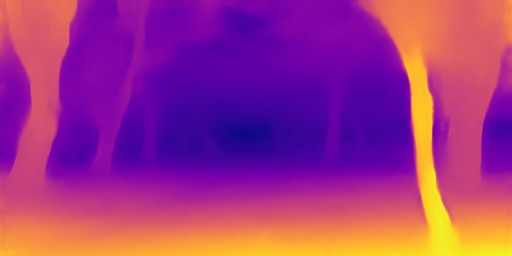}}
      \centerline{\includegraphics[width=\textwidth]{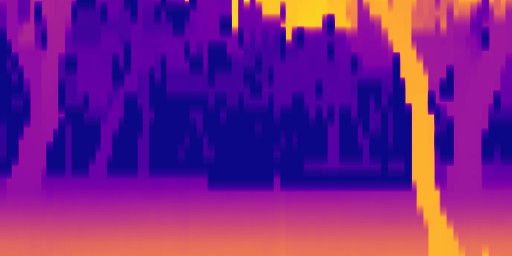}}
  \end{minipage}
  \begin{minipage}{0.19\linewidth}
      \small
      \centerline{\includegraphics[width=\textwidth]{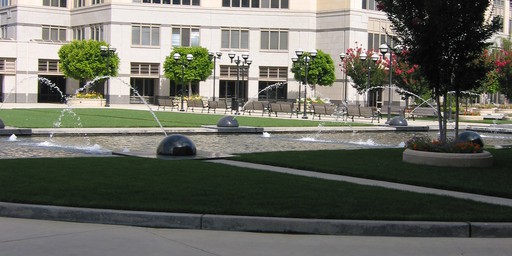}}
      \centerline{\includegraphics[width=\textwidth]{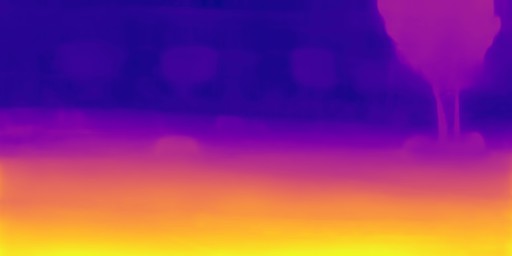}}
      \centerline{\includegraphics[width=\textwidth]{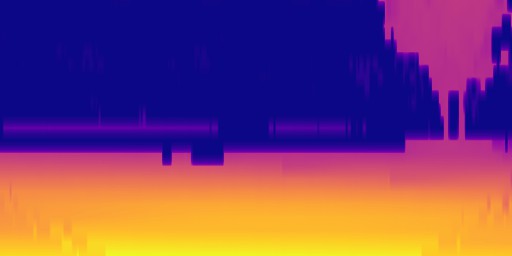}}
  \end{minipage}
  \begin{minipage}{0.19\linewidth} 
      \small
      \centerline{\includegraphics[width=\textwidth]{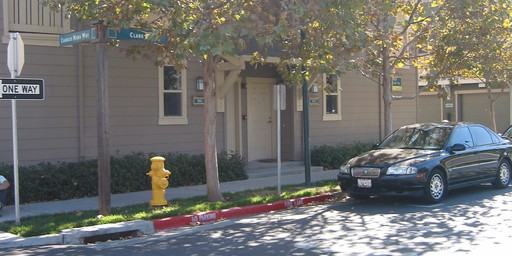}}
      \centerline{\includegraphics[width=\textwidth]{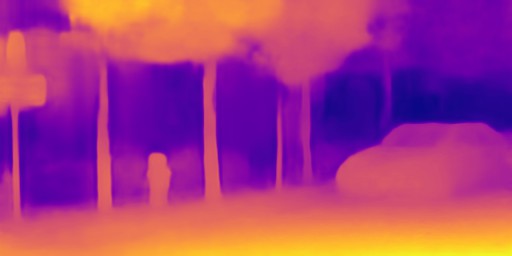}}
      \centerline{\includegraphics[width=\textwidth]{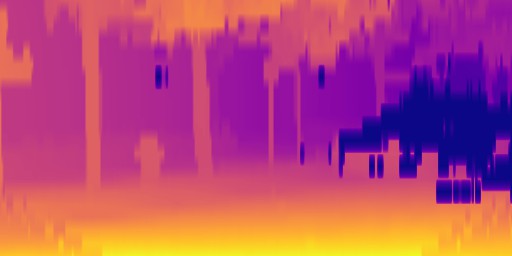}}
  \end{minipage}
  \caption{Visualization results of OCFD-Net on Make3D~\cite{Saxena2008Make3d}.}
  \label{fig:m3d}
\end{figure}

\begin{table*}
  \centering
  \small
  \caption{Quantitative comparison on the raw KITTI Eigen test set~\cite{Eigen2014Depth} in the ablation study.}
  \begin{tabular}{|l|cccc|ccc|}
  \hline
  \multicolumn{1}{|c|}{Method} &
    Abs Rel $\downarrow$ &
    Sq Rel $\downarrow$ &
    RMSE $\downarrow$ &
    logRMSE $\downarrow$ &
    A1 $\uparrow$ &
    A2 $\uparrow$ &
    A3 $\uparrow$\\ \hline
  Baseline &
    0.097 &
    0.602 &
    4.214 &
    0.183 &
    0.889 &
    0.963 &
    \underline{0.983} \\
  Baseline+DRB &
    0.095 &
    0.621 &
    4.162 &
    0.181 &
    0.891 &
    0.962 &
    0.982 \\
    Baseline+DRB+$M^l_v$ &
    0.094 &
    0.591 &
    4.102 &
    0.178 &
    0.895 &
    0.964 &
    \underline{0.983} \\
    Baseline+DRB+$M^l_m$ &
    \underline{0.093} &
    \underline{0.589} &
    \underline{4.079} &
    \underline{0.175} &
    \underline{0.898} &
    \underline{0.966} &
    \textbf{0.984} \\
    OCFD-Net &
    \textbf{0.091} &
    \textbf{0.576} &
    \textbf{4.036} &
    \textbf{0.174} &
    \textbf{0.901} &
    \textbf{0.967} &
    \textbf{0.984} \\
    \hline
  \end{tabular}
  \label{tab:represent}
\end{table*}

We firstly evaluate the OCFD-Net with/without a post-processing step (PP.)~\cite{Godard2017Unsupervised} on the raw KITTI Eigen test set~\cite{Eigen2014Depth} in comparison to 15 state-of-the-art methods, including 6 methods trained with monocular video sequences (M)~\cite{Zhao2020Masked, Zhou2019Unsupervised, Guizilini20203d, Johnston2020Self-supervised, Shu2020Feature-metric, Guizilini2020Semantically-guided} and 9 methods trained with stereo image pairs (S)~\cite{Poggi2018Learning, Peng2020A, Tosi2019Learning, Godard2019Digging, Pilzer2019Refine, Watson2019Self, Gonzalezbello2020Forget, Zhu2020The, Gonzalez2021Plade}. 
As done in \cite{Guizilini20203d, Gonzalezbello2020Forget, Gonzalez2021Plade}, we also evaluate the OCFD-Net on the improved KITTI Eigen test set~\cite{Uhrig2017Sparsity}.
The corresponding results by all the referred methods are cited from their original papers and reported in Table~\ref{tab:res}.
It is noted that some methods are trained with additional supervision, such as the semantic segmentation label (s)~\cite{Guizilini2020Semantically-guided, Zhu2020The}, and the offline computed disparity (d)~\cite{Tosi2019Learning, Watson2019Self, Zhu2020The}.

As seen from Table~\ref{tab:res}, when only the KITTI dataset~\cite{Geiger2012We} is used for training (K), our OCFD-Net without post processing outperforms the comparative methods without post processing under all the evaluation metrics.
The performance of our method is improved by adopting the post processing step, which simply averages the depths of the input image and the flipped depths of a flipped copy of the image.
And our method performs best under 4 metrics and second-best under the other 3 metrics on the raw KITTI Eigen test set~\cite{Eigen2014Depth}.
When both Cityscapes~\cite{Cordts2016The} and KITTI~\cite{Geiger2012We} are jointly used for training (CS+K) as done in~\cite{Zhao2020Masked, Guizilini20203d, Guizilini2020Semantically-guided, Poggi2018Learning, Tosi2019Learning, Gonzalezbello2020Forget, Gonzalez2021Plade}, the performance of OCFD-Net is further boosted.
On the improved KITTI Eigen test set~\cite{Uhrig2017Sparsity}, our method performs better than all the comparative methods in most cases.
These results demonstrate that the OCFD-Net is able to achieve more effective depth estimation.

In Figure~\ref{fig:compare}, we also give several visualization results of OCFD-Net as well as two comparative methods, DepthHints~\cite{Watson2019Self} and FAL-Net~\cite{Gonzalezbello2020Forget}, which perform without extra semantic supervision as done in our method and achieve better performances than the other comparative methods in most cases.
Their visualization results are generated with their open-source pretrained models.
It can be seen that DepthHints predicts inaccurate depths on the regions close to object boundaries (first row of Figure~\ref{fig:compare}), FAL-Net predicts unsmooth depths on the flat regions (second row of Figure~\ref{fig:compare}), but our OCFD-Net could handle both the two cases effectively.
As seen from the yellow rectangle in the last row of Figure~\ref{fig:compare}, all the three methods generate unreliable depths on the black region, indicating that it is still hard for them to handle texture-less regions, and it would be one of our future works to improve the proposed method for handling texture-less regions more effectively.
More visualization results could be found in the supplemental material.

Furthermore, we train the OCFD-Net on KITTI~\cite{Geiger2012We} (or on both KITTI and Cityscapes~\cite{Cordts2016The}) and evaluate it on Make3D~\cite{Saxena2008Make3d} for testing its cross-dataset generalization ability. 
The corresponding results of the OCFD-Net and 5 comparative methods~\cite{Wang2018Learning, Godard2019Digging, Johnston2020Self-supervised, Gonzalezbello2020Forget,Gonzalez2021Plade} are reported in Table~\ref{tab:m3d}, where the results of these methods are cited from their original papers.
It can be seen that the OCFD-Net outperforms 4 comparative methods and is competitive with the state-of-the-art PLADE-Net~\cite{Gonzalez2021Plade}, demonstrating its generalization ability on the unseen dataset.
Several visualization results on Make3D shown in Figure~\ref{fig:m3d} further demonstrate that the OCFD-Net could estimate scene depths effectively and maintain detailed structures of scenes.

\begin{figure}
  \small
  \centering
  \begin{minipage}{0.37\linewidth}
    \small
    \centerline{\includegraphics[width=1\textwidth]{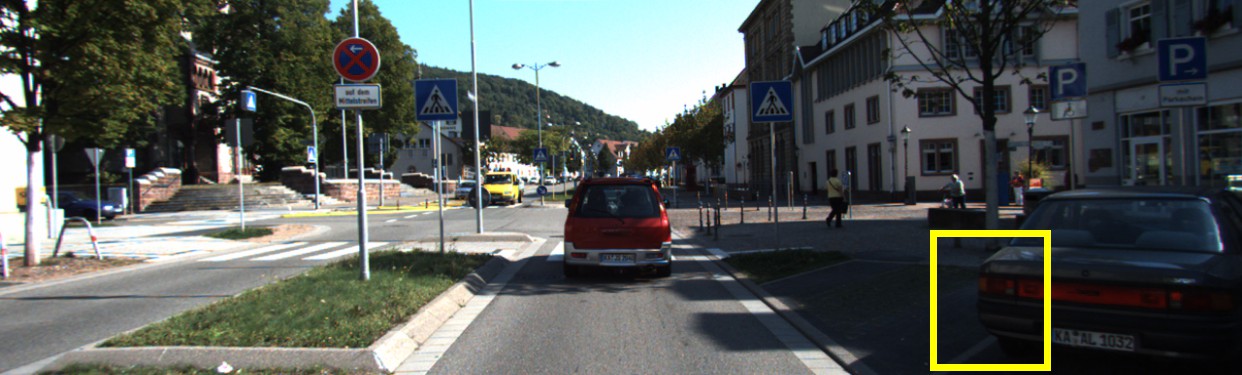}}
    \centerline{Input image}
    \centerline{\includegraphics[width=1\textwidth]{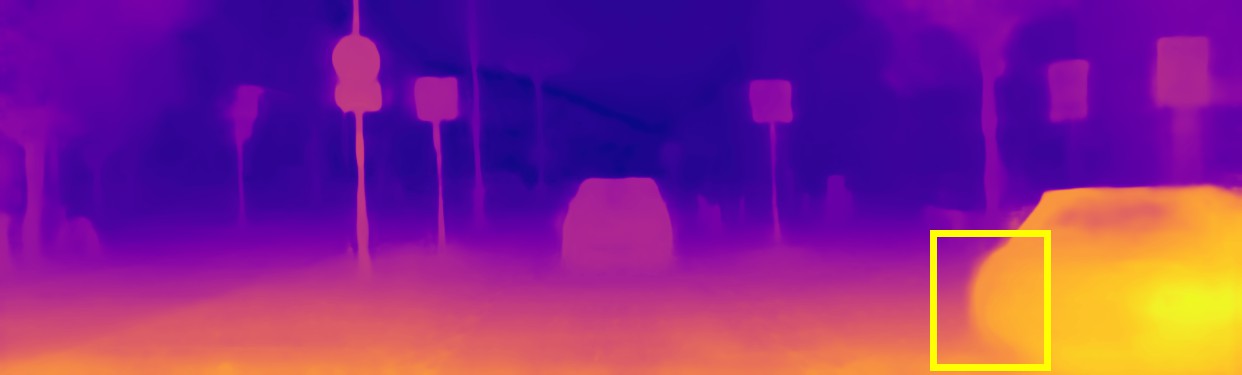}}
    \vspace{0.1em}
    \centerline{Baseline}
    \vspace{0.2em}
    \centerline{\includegraphics[width=1\textwidth]{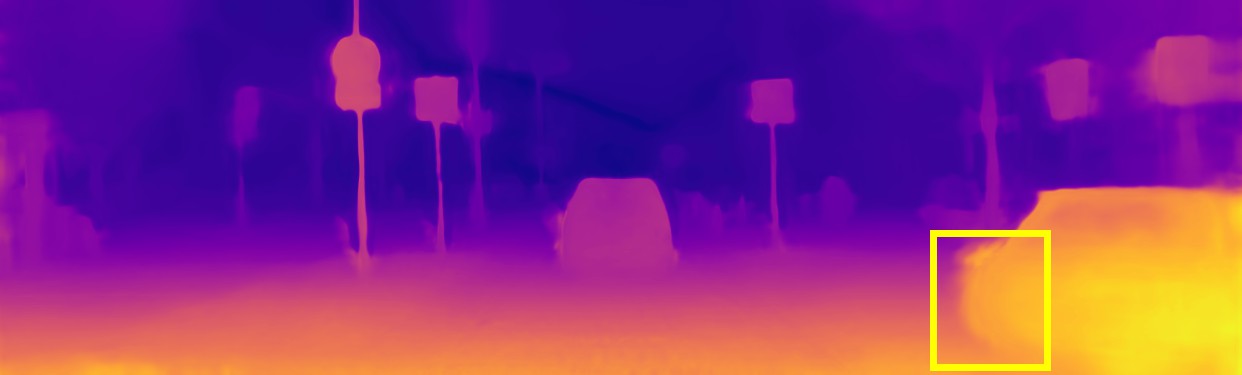}}
    \vspace{0.1em}
    \centerline{Baseline+DRB}
    \vspace{0.2em}
    \centerline{\includegraphics[width=1\textwidth]{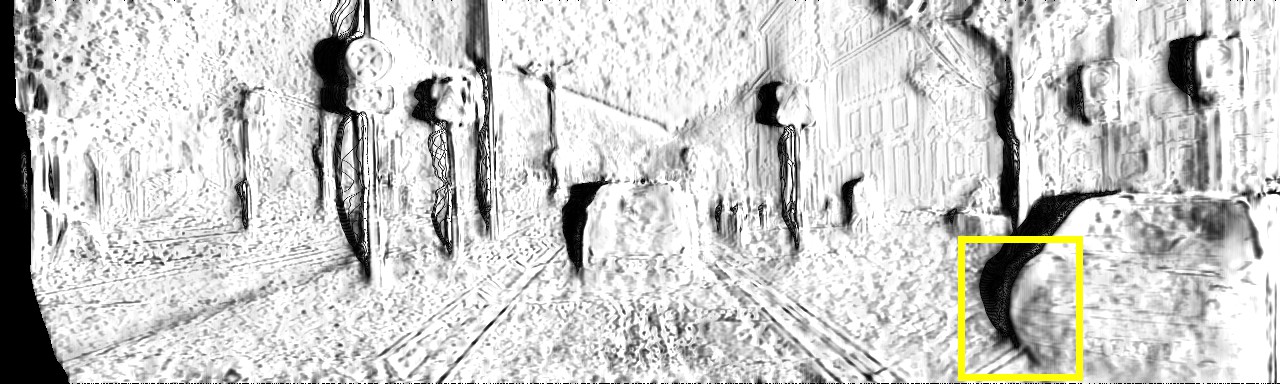}}
    \centerline{Occlusion mask $M^l_{occ}$}
  \end{minipage}
  \begin{minipage}{0.02\linewidth}
    \centerline{\includegraphics[width=0.245\textwidth]{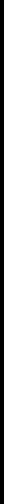}}
  \end{minipage}
  \begin{minipage}{0.491\linewidth}
    \begin{minipage}{0.753\linewidth}
      \small
      \centerline{\includegraphics[width=1\textwidth]{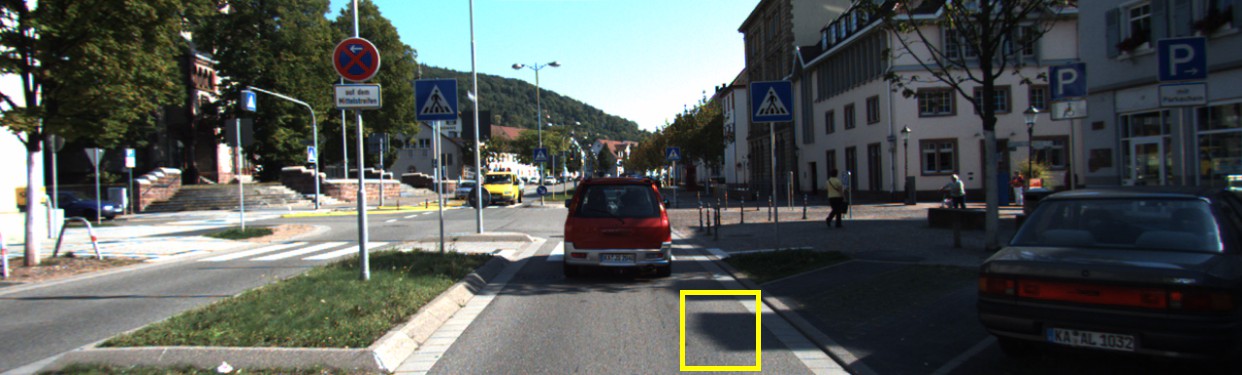}}
    \end{minipage}
    \begin{minipage}{0.23\linewidth}
      \small
      \centerline{\includegraphics[width=1\textwidth]{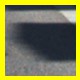}}
    \end{minipage}
    \vspace{0.2em}
    \centerline{Input image (Local regions)}
    \vspace{0.2em}
    \begin{minipage}{0.753\linewidth}
      \small
      \centerline{\includegraphics[width=1\textwidth]{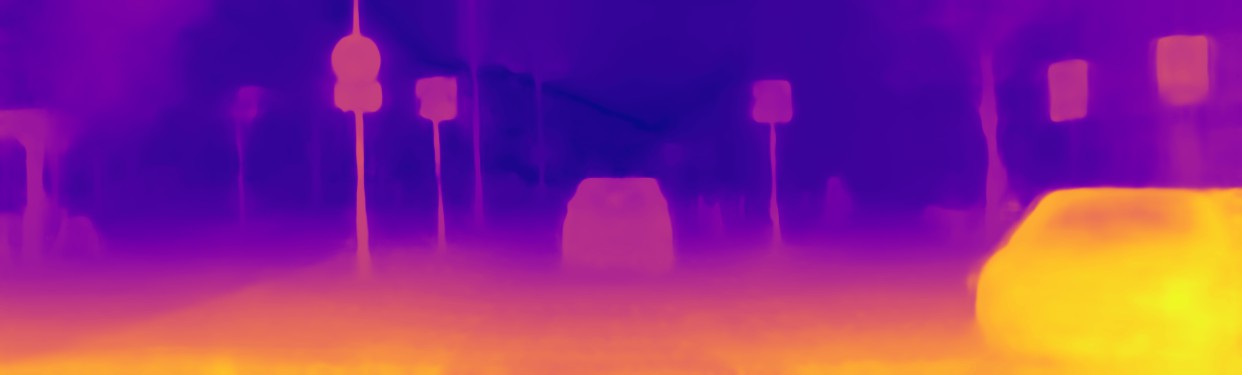}}
    \end{minipage}
    \begin{minipage}{0.23\linewidth}
      \small
      \centerline{\includegraphics[width=1\textwidth]{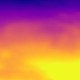}}
    \end{minipage}
    \centerline{$D^l_c$ of OCFD-Net}
    \begin{minipage}{0.753\linewidth}
      \small
      \centerline{\includegraphics[width=1\textwidth]{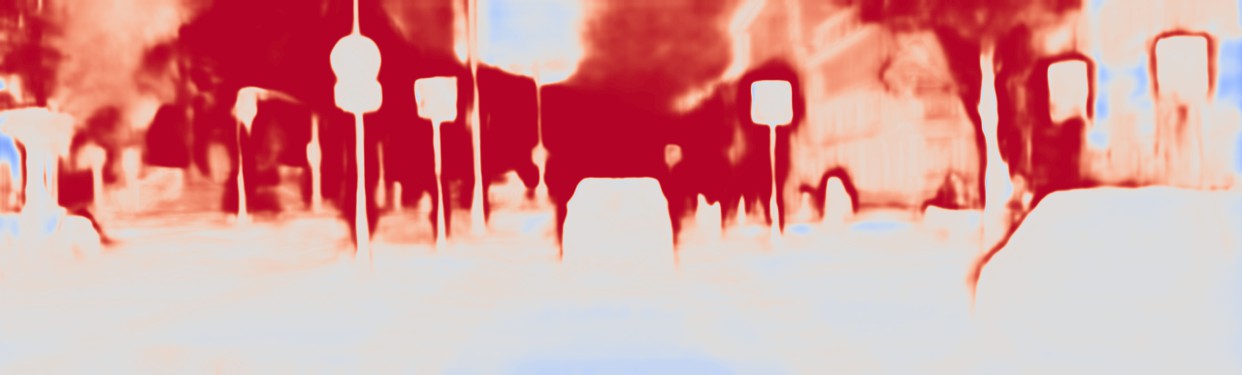}}
    \end{minipage}
    \begin{minipage}{0.23\linewidth}
      \small
      \centerline{\includegraphics[width=1\textwidth]{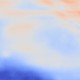}}
    \end{minipage}
    \centerline{$D^l_{res}$ of OCFD-Net}
    \begin{minipage}{0.753\linewidth}
      \small
      \centerline{\includegraphics[width=1\textwidth]{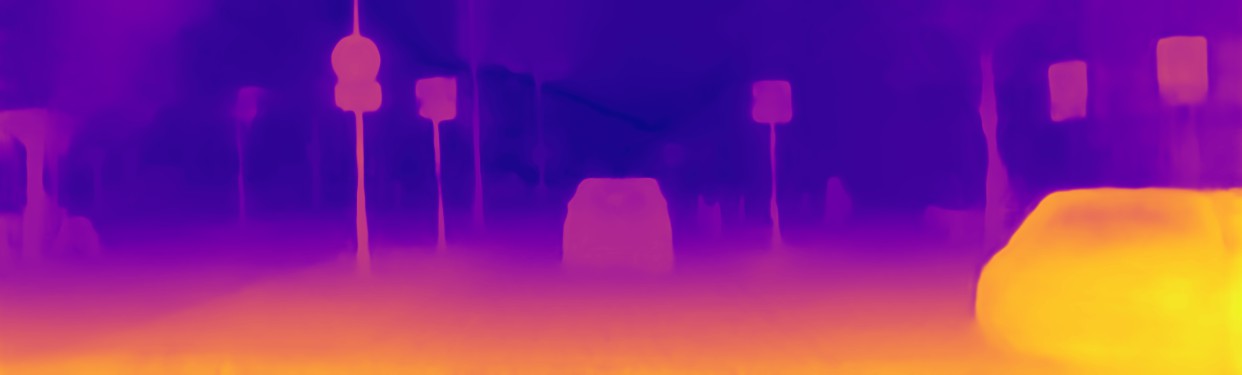}}
    \end{minipage}
    \begin{minipage}{0.23\linewidth}
      \small
      \centerline{\includegraphics[width=1\textwidth]{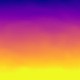}}
    \end{minipage}
    \centerline{$D^l_f$ of OCFD-Net (final prediction)}
  \end{minipage}
  \caption{Visualization results by the different modules on KITTI~\cite{Geiger2012We}.
  For the scene depth residual $D^l_{res}$, different colors indicate different residuals (red indicates positive, and blue indicates negative).
  For the occlusion mask $M^l_{occ}$, black indicates zero, and white indicates one.}
  \label{fig:abla}
\end{figure}

\subsection{Ablation studies} 
This subsection verifies the effectiveness of each key element in OCFD-Net by conducting ablation studies on the KITTI dataset~\cite{Geiger2012We}. 
We firstly train a simplest version of OCFD-Net (denoted as Baseline), consisting of the proposed backbone sub-network and the coarse-level depth prediction branch.
Then, we sequentially add the Depth Residual prediction Branch (DRB), the probability-volume-based mask ($M^l_v$), and the disparity-map-based mask ($M^l_m$) into the model.

The results are reported in Table~\ref{tab:represent}. It is noted that `Baseline+DRB' performs better than `Baseline' under 4 metrics, probably because the DRB improves the smoothness of the estimated depths on flat regions, but it performs poorer under the metric `Sq Rel', mainly because the depths of occluded regions are simultaneously wrongly smoothed, as illustrated by the corresponding visualization result in the left column of Figure~\ref{fig:abla}. Additionally, by singly utilizing the occlusion mask $M^l_v$ (also $M^l_m$), the depth estimation accuracy is further improved.
Our full model (OCFD-Net) with $M^l_{occ}$ could detect occluded regions effectively, as illustrated on the bottom left of Figure~\ref{fig:abla}, and it performs best under all the metrics.

To further understand the effect of the depth residual prediction branch, we visualize the depth maps and the residual map generated by OCFD-Net in the right column of Figure~\ref{fig:abla}.
It can be seen that the intensities of the depth residual map $D^l_{res}$ are large on the relatively far regions, and the enlarged versions of the yellow rectangle regions further show that a smoother fine-level depth map $D^l_f$ is obtained by integrating $D^l_{res}$ with the coarse-level depth map $D^l_c$.
Please see the supplemental material for more visualization results.

We also evaluate the influence of the residual weight $w$ in Equation~(\ref{equ:resw}) by training the OCFD-Net with $w=\{1, 5, 10, 20, 100\}$ respectively.
The corresponding results are reported in Table~\ref{tab:weight}.
As seen from this table, when $w$ ranges from 5 to 20, the corresponding results are close, and the OCFD-Net with $w=10$ achieves a trade-off among all the evaluation metrics.
It demonstrates that the performance of OCFD-Net is not sensitive to the residual weight $w$.

\begin{table}
  \centering
  \small
  \renewcommand\tabcolsep{3.0pt}
  \caption{Quantitative comparison of the OCFD-Net with different values of the residual weight $w$ on the raw KITTI Eigen test set~\cite{Eigen2014Depth}.}
  \begin{tabular}{|c|cccc|ccc|}
  \hline
 $w$ & Abs Rel $\downarrow$ & Sq Rel $\downarrow$ & RMSE $\downarrow$ & logRMSE $\downarrow$ & A1 $\uparrow$ & A2 $\uparrow$& A3 $\uparrow$\\\hline
    1   & 0.094       & 0.597       & 4.225       & 0.179       & 0.890       & 0.963       & \underline{0.983}    \\
  5   & \underline{0.093} & \underline{0.578} & 4.115       & \underline{0.175} & 0.896       & \underline{0.965} & \textbf{0.984} \\
  10 & \textbf{0.091} & \textbf{0.576} & \textbf{4.036} & \textbf{0.174} & \textbf{0.901} & \textbf{0.967} & \textbf{0.984} \\
  20  & \underline{0.093} & 0.597       & \underline{4.046} & \textbf{0.174}       & \underline{0.899}       & \underline{0.965} & \textbf{0.984} \\
  100 & 0.097       & 0.734       & 4.287       & 0.179       & \underline{0.899} & \underline{0.965} & \underline{0.983}    \\\hline  
  \end{tabular}
  \label{tab:weight}
\end{table}

\section{Conclusion}
\label{sec:conclusion}
In this paper, we propose the OCFD-Net for self-supervised monocular depth estimation.
Firstly, we empirically find that both the discrete and continuous depth constraints widely used in literature have their own advantage and disadvantage: the discrete depth constraint is relatively more effective for improving estimation accuracy, while the continuous one maintains relatively better depth smoothness.
Inspired by this finding, we design the OCFD-Net to learn a coarse-to-fine depth map with stereo image pairs by jointly utilizing both the continuous and discrete depth constraints.
Moreover, we explore an occlusion-aware module for handling occlusions under the OCFD-Net.
Experimental results show the effectiveness of the proposed OCFD-Net.

In the future, we will further investigate how to make use of both the continuous and discrete constraints more effectively for improving depth estimation accuracy, as well as how to effectively handle texture-less regions as indicated in Section~\ref{sec:experiments-compare}.

\begin{acks}
This work was supported by the National Key R\&D Program of China (Grant No. 2021ZD0201600), the National Natural Science Foundation of China (Grant Nos. U1805264 and 61991423), the Strategic Priority Research Program of the Chinese Academy of Sciences (Grant No. XDB32050100), the Beijing Municipal Science and Technology Project (Grant No. Z211100011021004).
\end{acks}

\bibliographystyle{ACM-Reference-Format}
\bibliography{base}
\clearpage

\appendix
\renewcommand{\maketitle}{
 \
 \null
 \begin{center}
  {\Huge Supplementary Material \par}
 \end{center}%
 \par}
\maketitle

\setcounter{figure}{0}
\setcounter{equation}{0}

\renewcommand{\theequation}{A.\arabic{equation}}
\renewcommand{\thetable}{A\arabic{table}}
\renewcommand{\thefigure}{A\arabic{figure}}

\section{Mathematical description of the geometric transformations}
\textbf{Notation}: In this paper, the proposed method (also the used comparative methods) is trained with stereo image pairs. As done in~\cite{Godard2017Unsupervised,Watson2019Self,Gonzalezbello2020Forget}, it is assumed that the intrinsic matrices of the left and right cameras are identical and their relative pose is only up to a pure translation along the X-axis. Hence, let $t=\left[-B, 0, 0\right]^\top$ be the translation from the left camera to the right one ($B$ denotes the baseline length),  and let $R$ be the rotation matrix between them ($R$ is indeed a $3 \times 3$ identity matrix $I$). Let $K$ denote the intrinsic matrix of the two cameras as:
\begin{equation}
  K=\begin{bmatrix}f_x & 0 & c_x \\ 0 & f_y & c_y \\0 & 0 & 1 \\\end{bmatrix}.
  \label{eq:k}
\end{equation}
where $f_x, f_y$ denote the focal lengths of the camera, and $(c_x, c_y)$ denotes the principal point of the camera.

There are the following two geometric transformations in Section 3.2.2 "Image reconstruction module": 

\textbf{(1)} Geometric transformation  ${\rm T_1}(\cdot)$ in the "Continuous reconstruction block": This block takes the right image $I^r$ and the predicted fine-level depth map $D^l_f$ of the left image as the input. And it aims to reconstruct the left image $\hat{I}^l$ by assigning the value of the right image pixel $p'$ to the pixel $p$ of  $\hat{I}^l$:
\begin{equation}
  \hat{I}^l(p)=I^r\left<p'\right>.
  \label{eq:con_rec}
\end{equation}
where "$<\cdot>$" denotes the bilinear sampling operator. Based on Eq.(\ref{eq:con_rec}), the transformation ${\rm T_1}(\cdot)$ is used to convert the homogeneous coordinate $p$  of an arbitrary pixel in the left image to its corresponding coordinate $p'$ in the right image according to both the camera intrinsic/extrinsic matrices and a predicted depth $D^l_f(p)$ on $p$, which is formulated as:
\begin{equation}
  p'\sim{\rm T_1}(p)= K[R|t] \left[ \begin{matrix}
    D^l_f(p)K^{-1}p \\
    1
   \end{matrix}
   \right].
   \label{eq:t1}
\end{equation}
Since $R$ is a  $3 \times 3$ identity matrix and $t=\left[-B, 0, 0\right]^\top$, Eq.(\ref{eq:t1}) could be re-formulated by introducing Eq.(\ref{eq:k}) into Eq.(\ref{eq:t1}):
\begin{equation}
  p'\sim {\rm T_1}(p) = D^l_f(p)\left(p-\left[\frac{Bf_x}{D^l_f(p)},0,0\right]^\top\right).
  \label{eq:t1-2}
\end{equation}
It can be seen that Eq.(\ref{eq:t1-2}) is the homogeneous form of Eq.(7).

\textbf{(2)} Geometric transformation  ${\rm T_2}(\cdot)$ in the *Discrete reconstruction block*: For reconstructing the right image $\hat{I}^r$ based on the discrete depth constraint, this block firstly takes the $\rm n^{th}$ channel of the predicted density volume  $V_{dn}^l$ ,  the left image $I^l$  and one of predefined disparity $d_n$ as the input and generates the right-view density volume  $\hat{V}_{dn}^r$ and the shifted left image $I^l_n$ as:
\begin{equation}
  I^l_n(p)=I^l\left<p'\right>,
  \label{eq:dis-rec1}
\end{equation}
\begin{equation}
  \hat{V}_{dn}^r(p)=V_{dn}^l\left<p'\right>.
  \label{eq:dis-rec2}
\end{equation}
Based on Eq.(\ref{eq:dis-rec1}) and Eq.(\ref{eq:dis-rec2}), the transformation ${\rm T_2}(\cdot)$ is used to convert the homogeneous coordinate $p$  of an arbitrary pixel in the right image to its corresponding coordinate $p'$ in the left image according to both the camera intrinsic/extrinsic matrices and a predefined disparity $d_n$. Let $D_n=\frac{Bf_x}{d_n}$ be the depth converted from $d_n$ based on the definition of disparity, ${\rm T_2}(\cdot)$ is formulated as:
\begin{equation}
  p'\sim {\rm T_2}(p) 
= K[R|-t] \left[ \begin{matrix}
 D_nK^{-1}p \\
 1
\end{matrix}
\right]
= D_n\left(p+\left[d_n,0,0\right]^\top\right).
\end{equation}
After obtaining $\hat{V}_{dn}^r,I^l_n$ by the transformation ${\rm T_2}(\cdot)$, $\hat{V}_p^r$ is generated by passing $\hat{V}_d^r$ through a softmax operation and $\hat{I}^r$ is calculated with  $I^l_n$ and  $\hat{V}_p^r$ according to Eq.(6).

\section{Details on datasets and metrics}
\subsection{Datasets}
The three datasets used in this work are introduced in detail as follows: 
\begin{itemize}
\item KITTI~\cite{Geiger2012We} contains the rectified stereo image pairs captured from a driving car.
We use the Eigen splits~\cite{Eigen2014Depth} to train and evaluate the OCFD-Net, which consist of 22600 stereo image pairs for training and 697 images for testing.
Additionally, we also evaluate the OCFD-Net on the improved Eigen test set, which consists of 652 images and adopts the high-quality ground-truth depth maps generated with the method in~\cite{Uhrig2017Sparsity}.
At both the training and inference stages, the images are resized into the resolution of $1280 \times 384$, while we assume that the intrinsics of all the images are identical.  

\item Cityscapes~\cite{Cordts2016The} contains the stereo pairs of urban driving scenes, and we take 22972 stereo pairs from it for jointly training the OCFD-Net.
When the OCFD-Net is trained on both the KITTI and Cityscapes datasets, we crop and resize the images from Cityscapes into the resolution of $1280 \times 384$.
Considering that the baseline length in Cityscapes is different from that in KITTI, we scale the predicted disparities on Cityscapes by the rough ratio of the baseline lengths in the two datasets.

\item Make3D~\cite{Saxena2008Make3d} is a commonly used dataset for depth estimation in outdoor scenes.
Since self-supervised depth estimation methods could not be trained on Make3D, the test set of it including 134 images could be utilized to test the cross-dataset generalization ability.
For a fair comparison, we crop and resize the input images as done in~\cite{Godard2019Digging} at the inference stage.
\end{itemize}

\subsection{Metrics}
For the evaluation on the KITTI dataset~\cite{Geiger2012We}, we use the center crop proposed in~\cite{Garg2016Unsupervised} and the standard cap of 80m.
The following metrics are used:
\begin{itemize}
    \item Abs Rel: $\frac{1}{N}\sum_i{\frac{\left| \hat{D}_i - D^{gt}_i \right|}{D^{gt}_i}}$
    
    \item Sq Rel: $\frac{1}{N}\sum_i{\frac{\left| \hat{D}_i - D^{gt}_i \right|^2}{D^{gt}_i}}$
    
    \item RMSE: $\sqrt{\frac{1}{N}\sum_i{\left| \hat{D}_i - D^{gt}_i \right|^2}}$
    \item logRMSE: $\sqrt{\frac{1}{N}\sum_i{\left| \log\left(\hat{D}_i\right) - \log\left(D^{gt}_i\right) \right|^2}}$
    \item Threshold (A$j$):  $ \% \quad s.t. \quad \max{\left( \frac{\hat{D}_i}{D^{gt}_i}, \frac{D^{gt}_i}{\hat{D}_i} \right)}< a^{j}$
\end{itemize}
where $\{\hat{D}_i, D^{gt}_i\}$ are the predicted depth and the ground-truth depth at pixel $i$, and $N$ denotes the total number of the pixels with the ground truth.
In practice, we use $a^{j} = 1.25, 1.25^{2}, 1.25^{3}$, which are denoted as A1, A2, and A3 in all the tables.

For the evaluation on the Make3D dataset~\cite{Saxena2008Make3d}, we use the per-image median scaling and the standard cap of 70m.
The following metrics are used: Abs Rel, Sq Rel, RMSE, and 
\begin{equation}
  log_{10}=\sqrt{\frac{1}{N}\sum_i{\left| \log_{10}\left(\hat{D}_i\right) - \log_{10}\left(D^{gt}_i\right) \right|^2}}.
\end{equation}

\section{Visualization results on the effects of the continuous and discrete depth constraints}
\begin{figure*}[h]
  \centering
  \begin{minipage}{0.13\linewidth}
      \small
      \leftline{Input images}
      \leftline{(Local regions)}
      \vspace{18pt}

      \leftline{FAL-Arc+CDC}
      \vspace{23pt}

      \leftline{FAL-Arc+DDC}
      \vspace{23pt}

      \leftline{Res-Arc+CDC}
      \vspace{23pt}

      \leftline{Res-Arc+DDC}
      \vspace{36pt}

      \leftline{Input images}
      \leftline{(Local regions)}
      \vspace{18pt}

      \leftline{FAL-Arc+CDC}
      \vspace{23pt}

      \leftline{FAL-Arc+DDC}
      \vspace{23pt}

      \leftline{Res-Arc+CDC}
      \vspace{23pt}

      \leftline{Res-Arc+DDC}
      \vspace{15pt}

  \end{minipage}
  \begin{minipage}{0.28\linewidth}
      \begin{minipage}{0.7527\linewidth}
        \centerline{\includegraphics[width=\textwidth]{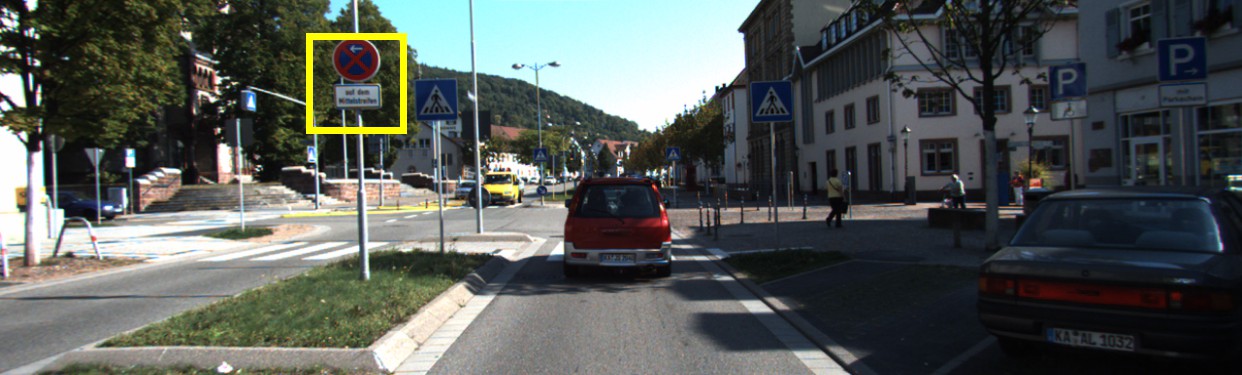}}
        \centerline{\includegraphics[width=\textwidth]{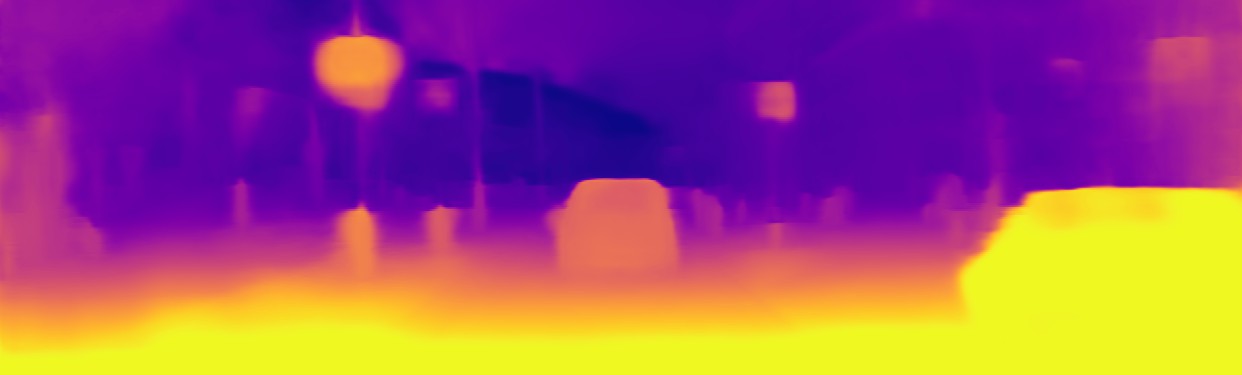}}
        \centerline{\includegraphics[width=\textwidth]{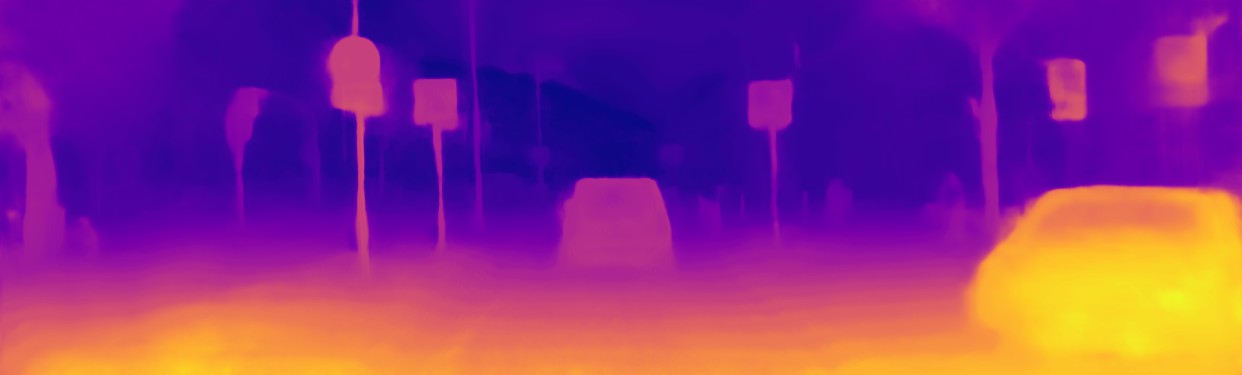}}
        \centerline{\includegraphics[width=\textwidth]{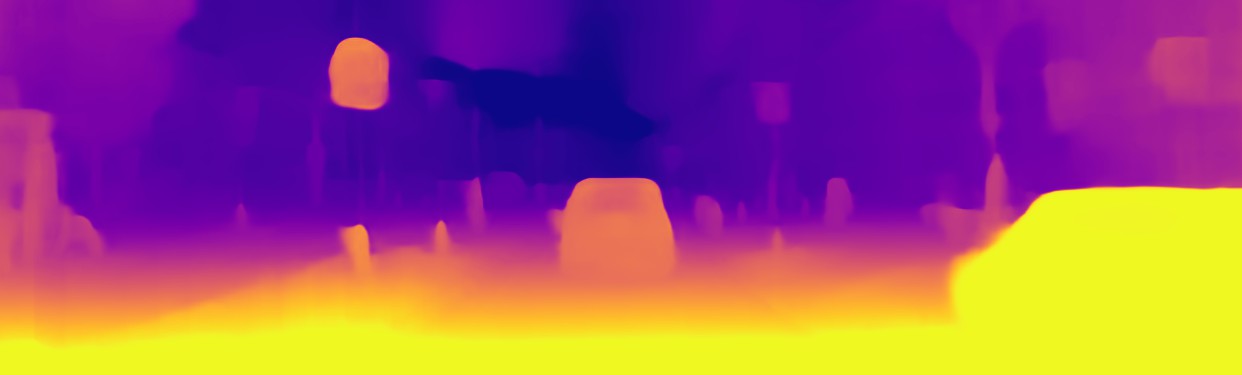}}
        \centerline{\includegraphics[width=\textwidth]{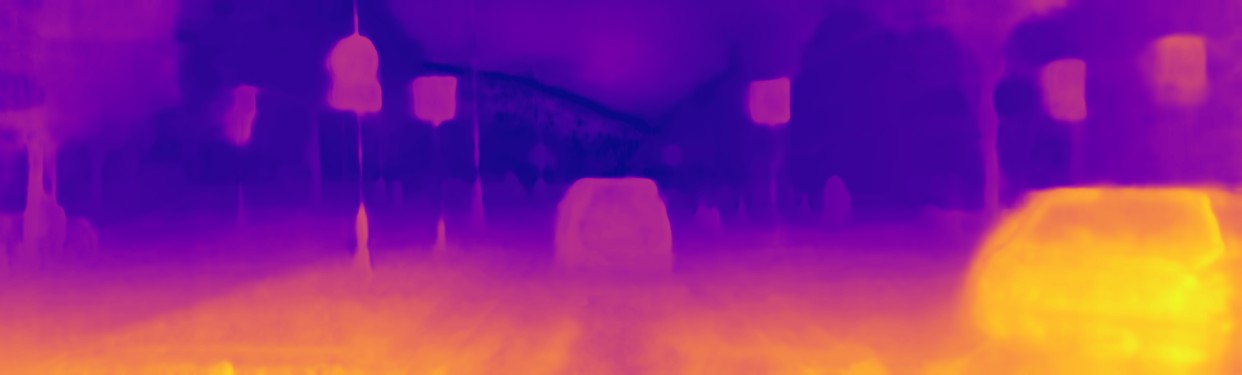}}
    \end{minipage}
    \begin{minipage}{0.2273\linewidth}
        \centerline{\includegraphics[width=\textwidth]{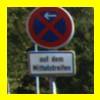}}
        \centerline{\includegraphics[width=\textwidth]{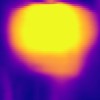}}
        \centerline{\includegraphics[width=\textwidth]{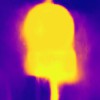}}
        \centerline{\includegraphics[width=\textwidth]{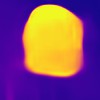}}
        \centerline{\includegraphics[width=\textwidth]{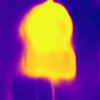}}
    \end{minipage}
    \centerline{(a)}
    \vspace{1pt}

    \begin{minipage}{0.7527\linewidth}
      \centerline{\includegraphics[width=\textwidth]{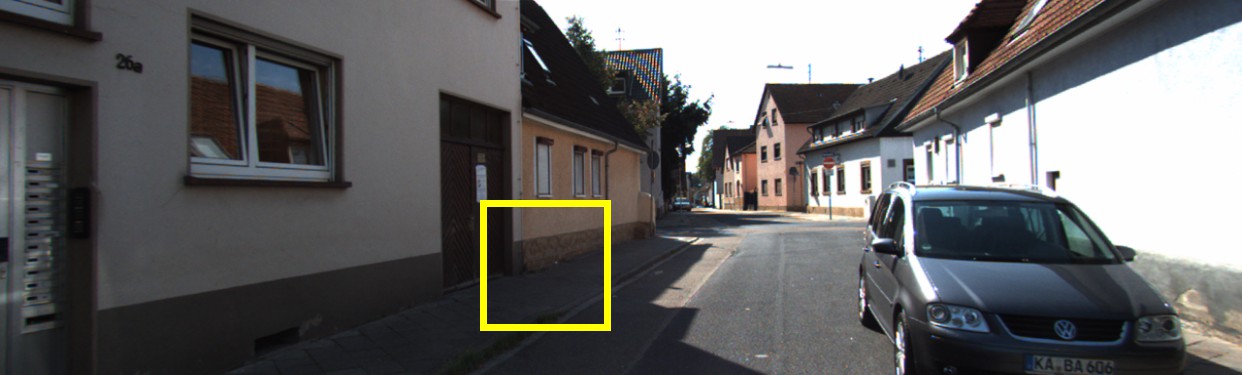}}
      \centerline{\includegraphics[width=\textwidth]{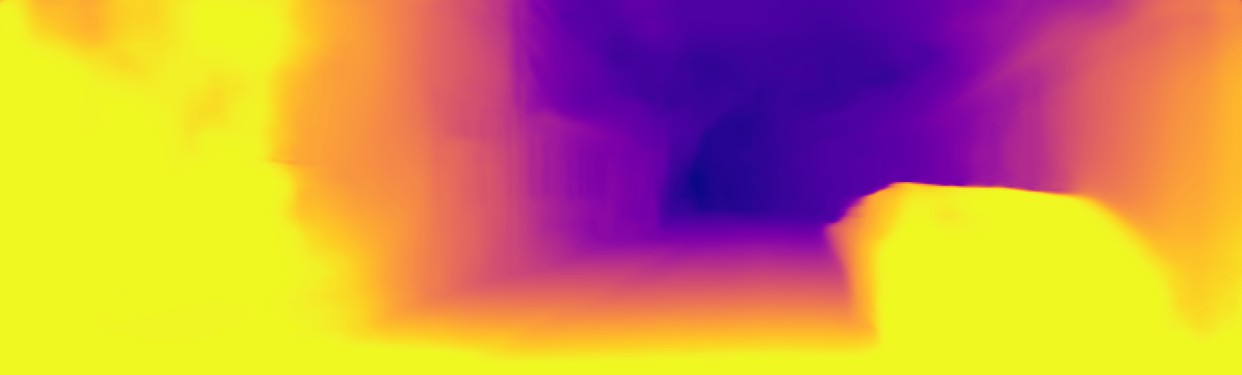}}
      \centerline{\includegraphics[width=\textwidth]{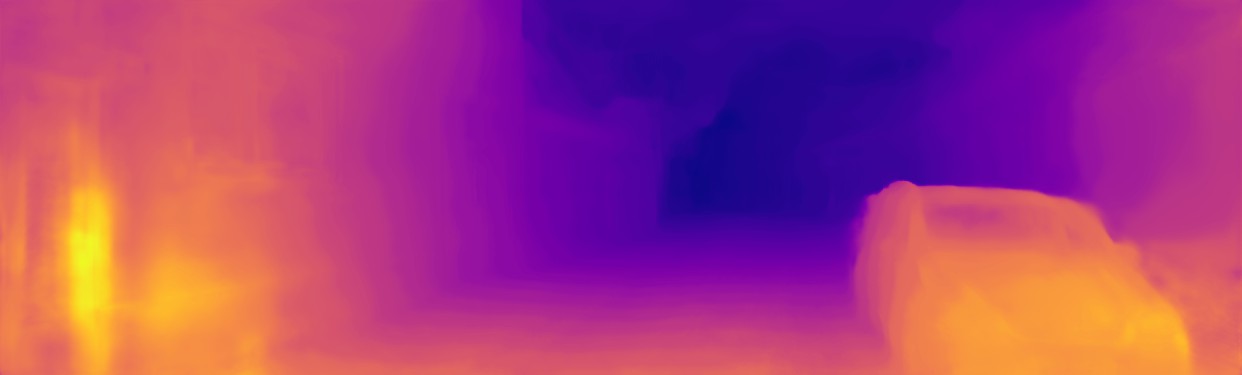}}
      \centerline{\includegraphics[width=\textwidth]{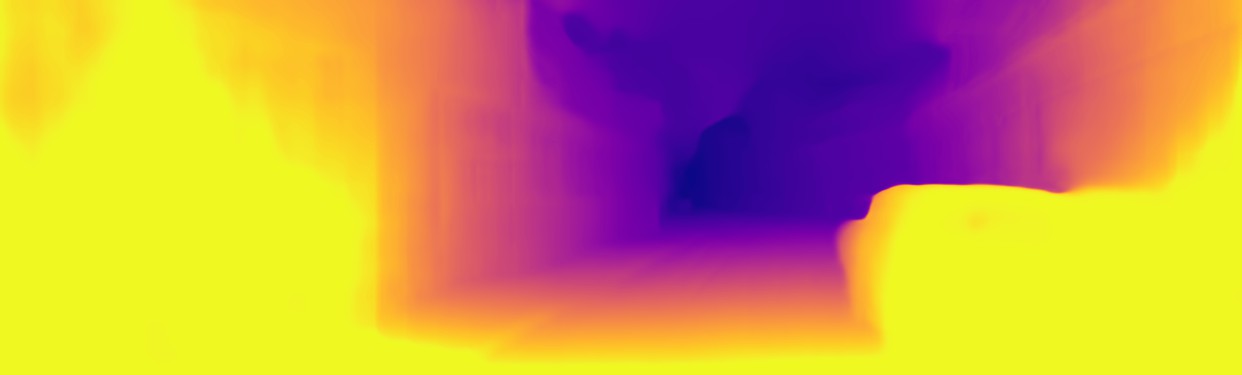}}
      \centerline{\includegraphics[width=\textwidth]{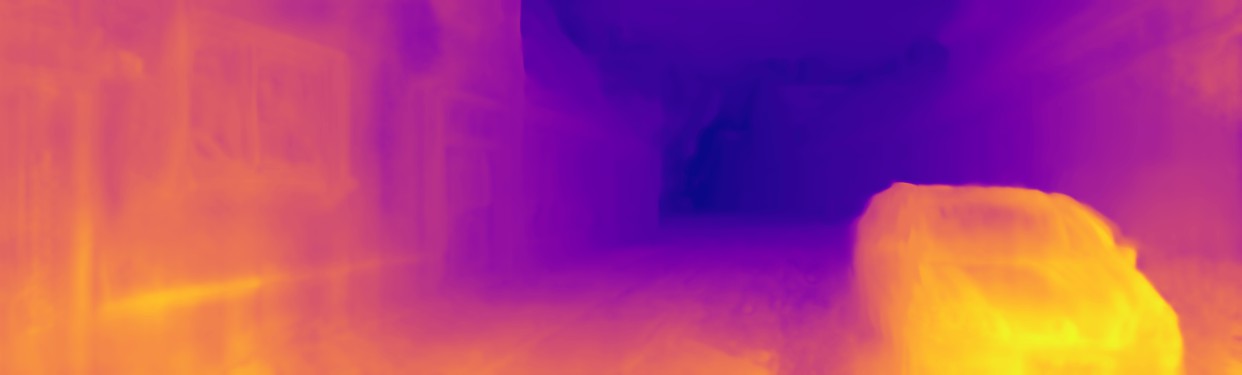}}
    \end{minipage}
    \begin{minipage}{0.2273\linewidth}
      \centerline{\includegraphics[width=\textwidth]{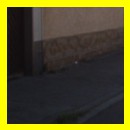}}
      \centerline{\includegraphics[width=\textwidth]{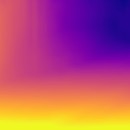}}
      \centerline{\includegraphics[width=\textwidth]{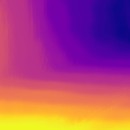}}
      \centerline{\includegraphics[width=\textwidth]{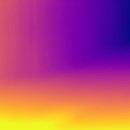}}
      \centerline{\includegraphics[width=\textwidth]{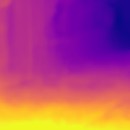}}
    \end{minipage}
    \centerline{(d)}
  \end{minipage}
  \begin{minipage}{0.28\linewidth}
    \begin{minipage}{0.7527\linewidth}
      \centerline{\includegraphics[width=\textwidth]{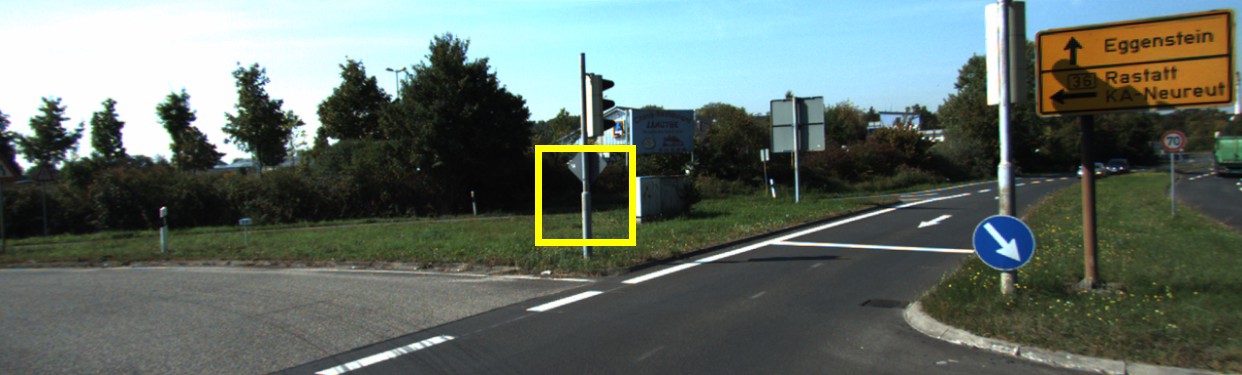}}
      \centerline{\includegraphics[width=\textwidth]{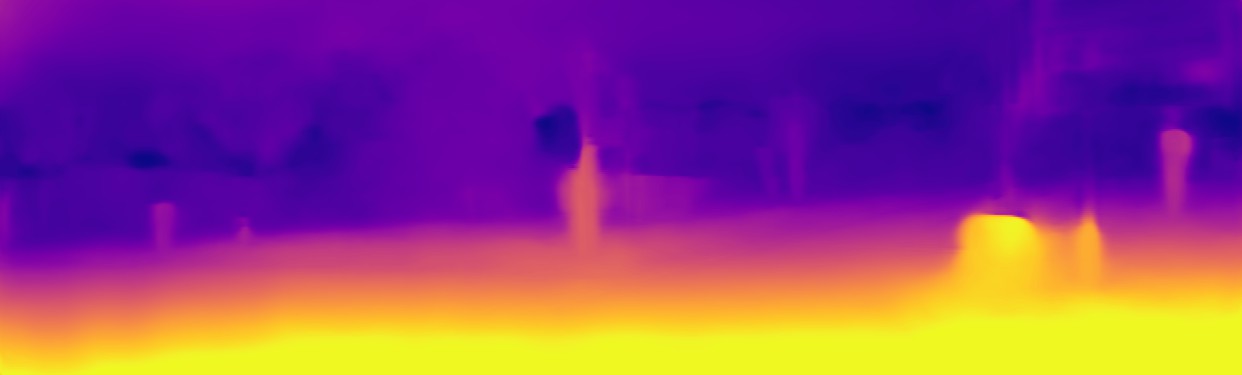}}
      \centerline{\includegraphics[width=\textwidth]{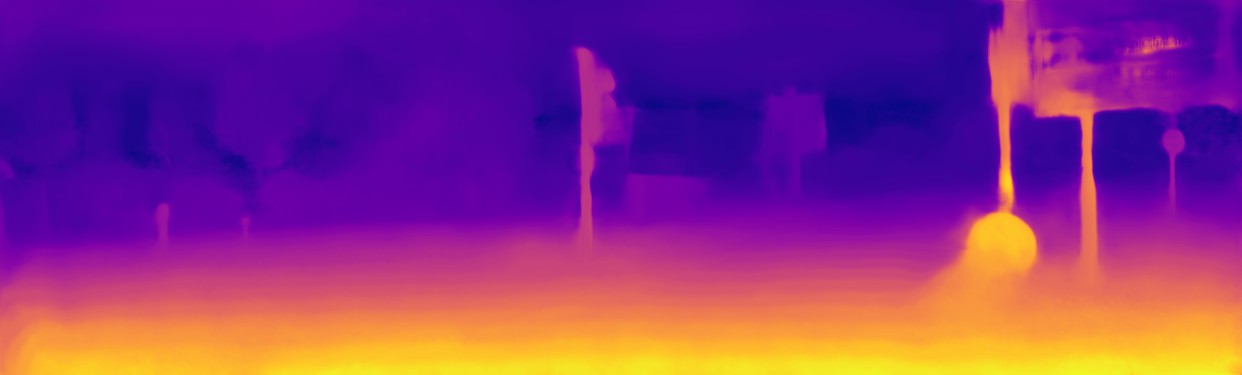}}
      \centerline{\includegraphics[width=\textwidth]{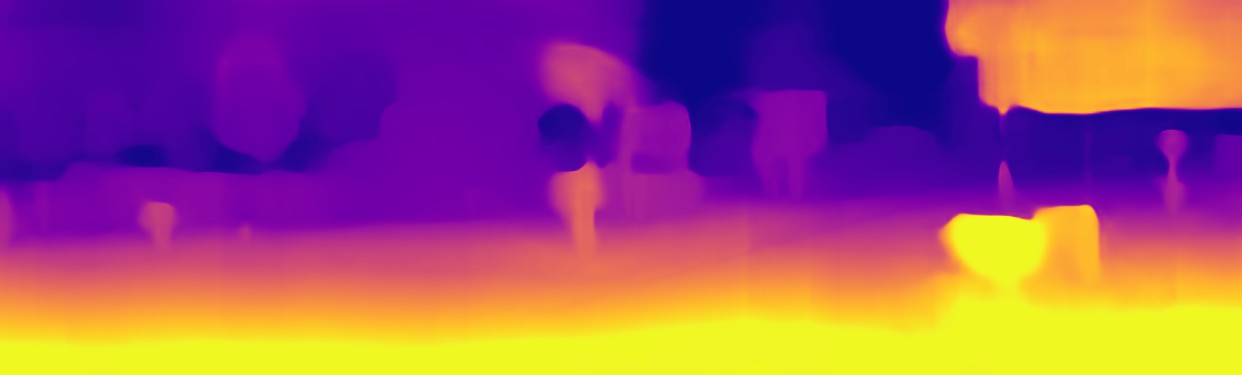}}
      \centerline{\includegraphics[width=\textwidth]{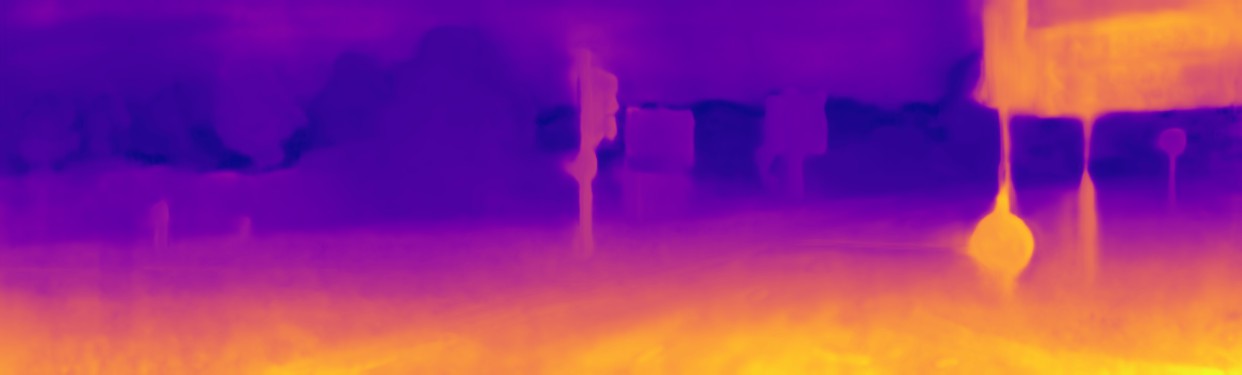}}
    \end{minipage}
    \begin{minipage}{0.2273\linewidth}
      \centerline{\includegraphics[width=\textwidth]{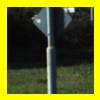}}
      \centerline{\includegraphics[width=\textwidth]{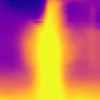}}
      \centerline{\includegraphics[width=\textwidth]{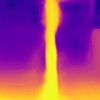}}
      \centerline{\includegraphics[width=\textwidth]{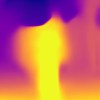}}
      \centerline{\includegraphics[width=\textwidth]{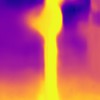}}
    \end{minipage}
    \centerline{(b)}
    \vspace{1pt}

    \begin{minipage}{0.7527\linewidth}
      \centerline{\includegraphics[width=\textwidth]{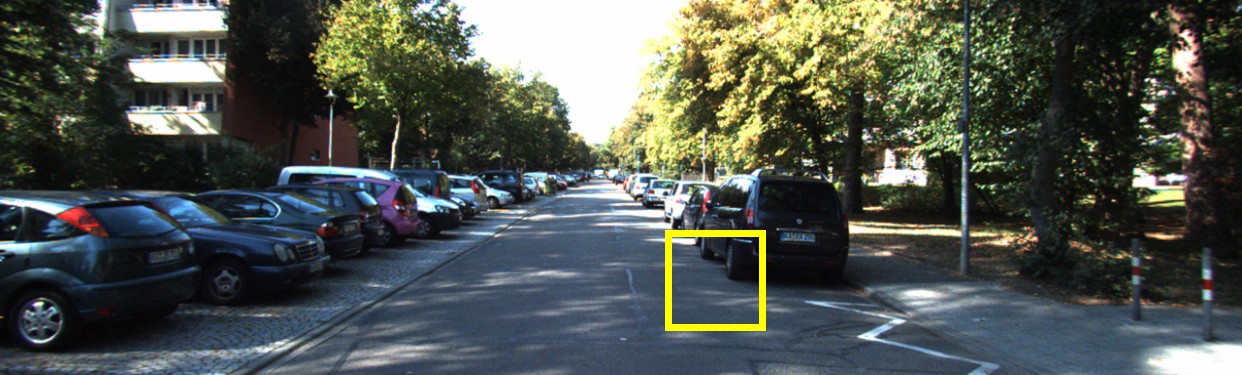}}
      \centerline{\includegraphics[width=\textwidth]{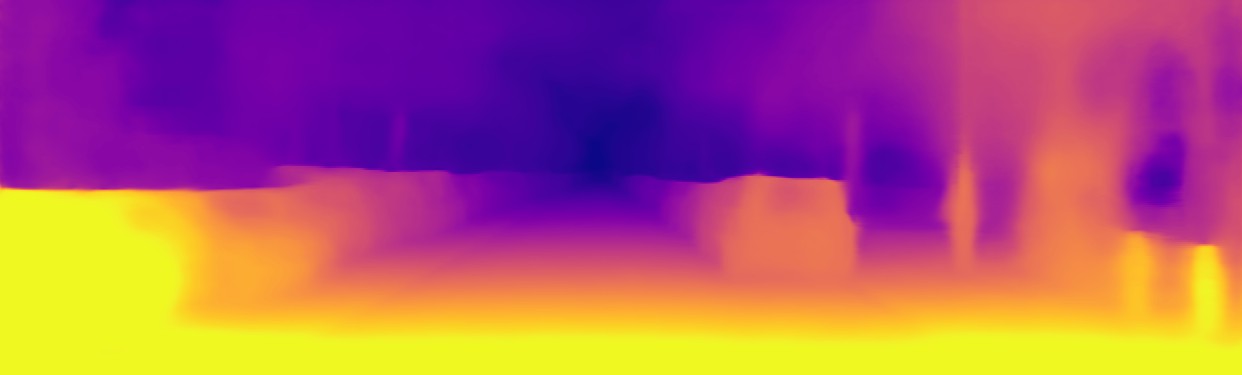}}
      \centerline{\includegraphics[width=\textwidth]{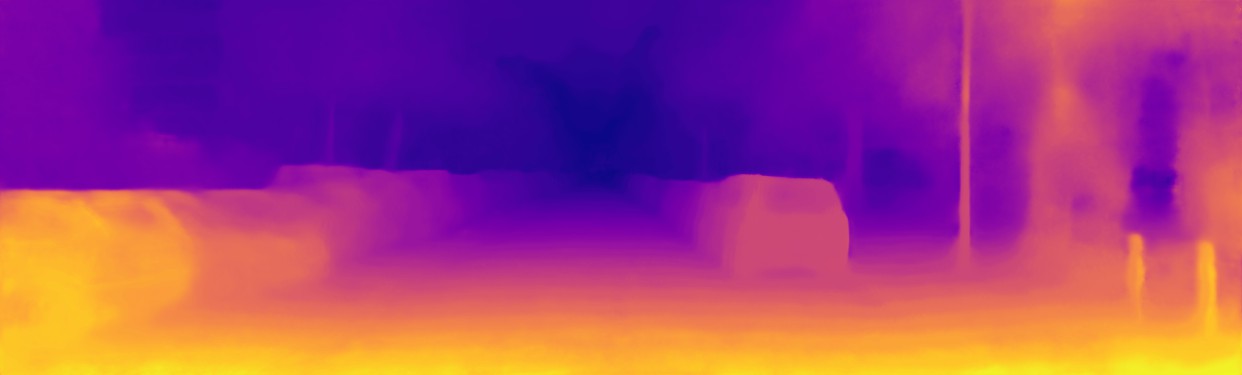}}
      \centerline{\includegraphics[width=\textwidth]{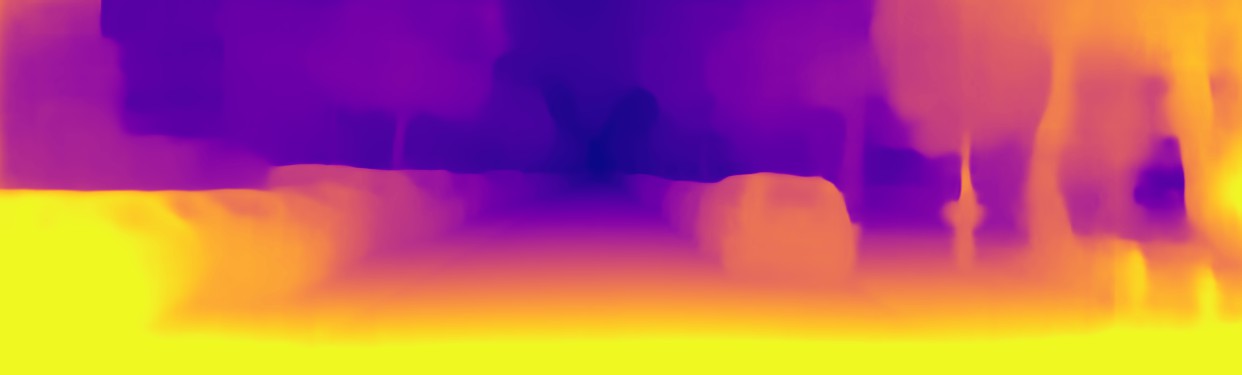}}
      \centerline{\includegraphics[width=\textwidth]{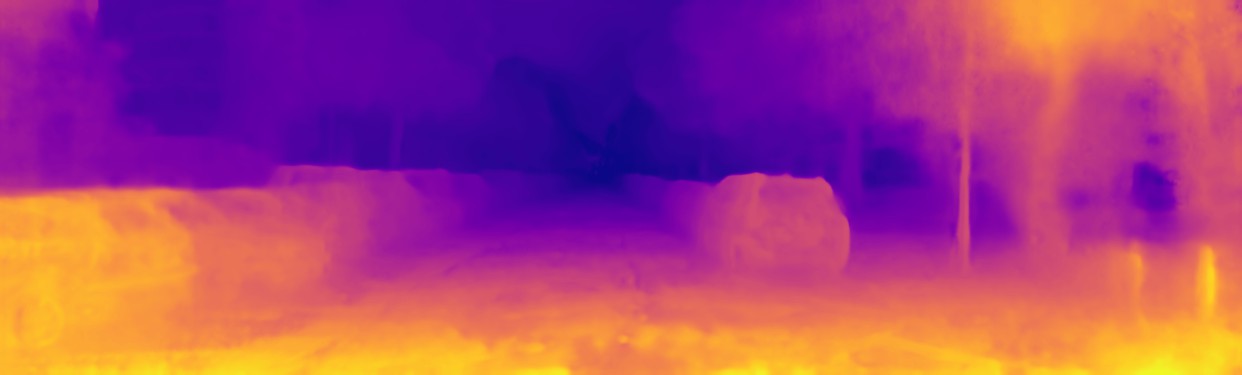}}
    \end{minipage}
    \begin{minipage}{0.2273\linewidth}
      \centerline{\includegraphics[width=\textwidth]{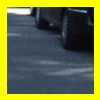}}
      \centerline{\includegraphics[width=\textwidth]{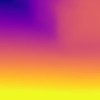}}
      \centerline{\includegraphics[width=\textwidth]{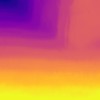}}
      \centerline{\includegraphics[width=\textwidth]{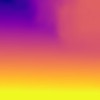}}
      \centerline{\includegraphics[width=\textwidth]{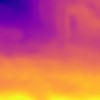}}
    \end{minipage}
    \centerline{(e)}
\end{minipage}
\begin{minipage}{0.28\linewidth}
  \begin{minipage}{0.7527\linewidth}
    \centerline{\includegraphics[width=\textwidth]{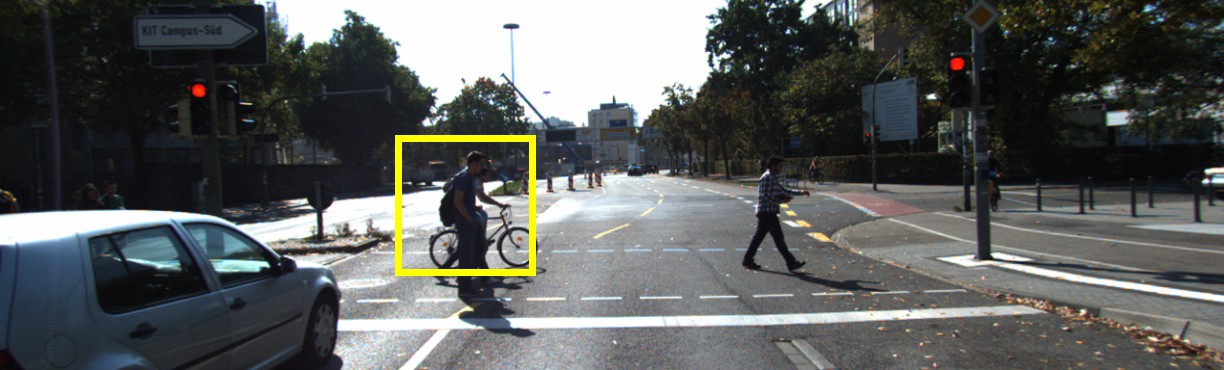}}
    \centerline{\includegraphics[width=\textwidth]{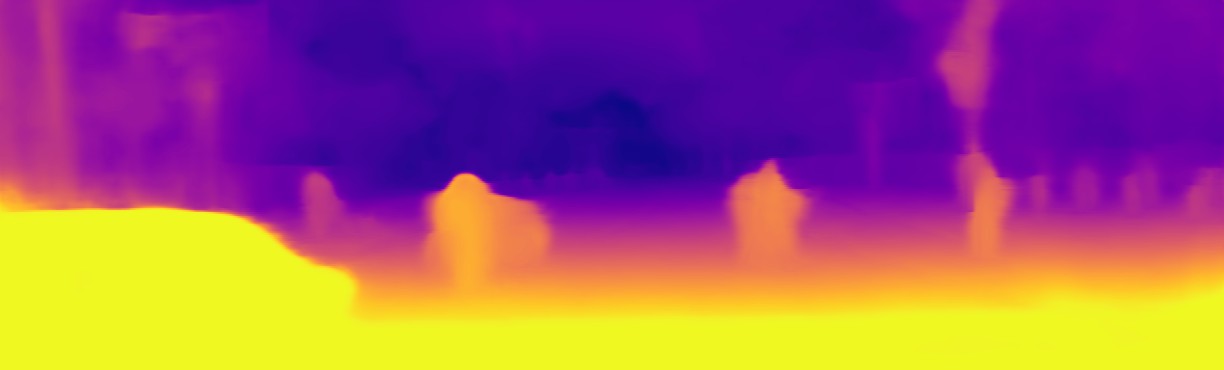}}
    \centerline{\includegraphics[width=\textwidth]{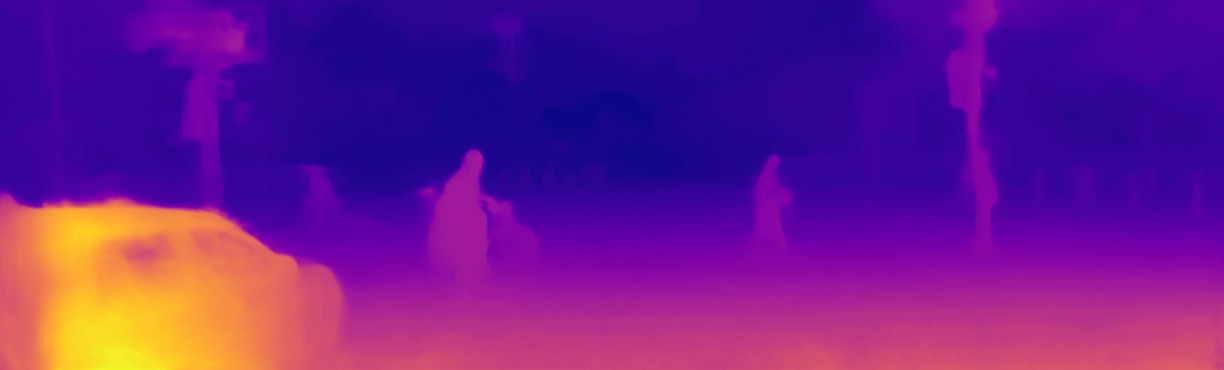}}
    \centerline{\includegraphics[width=\textwidth]{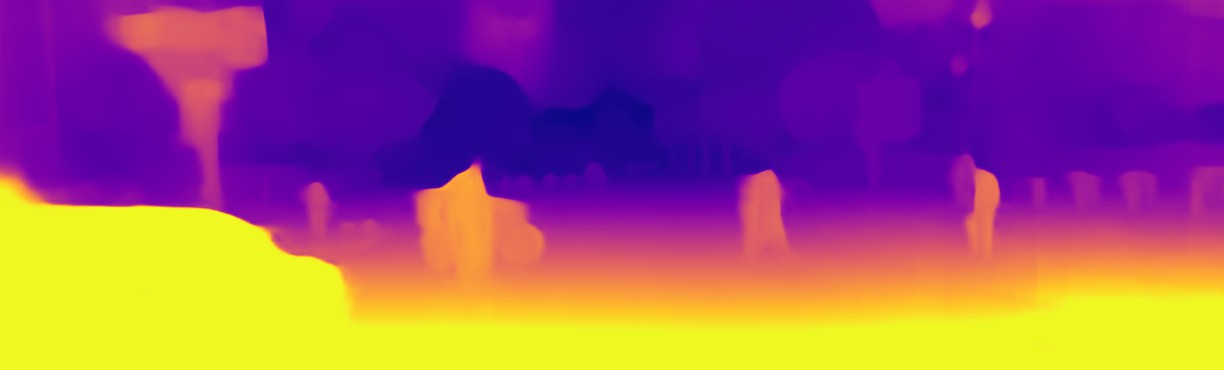}}
    \centerline{\includegraphics[width=\textwidth]{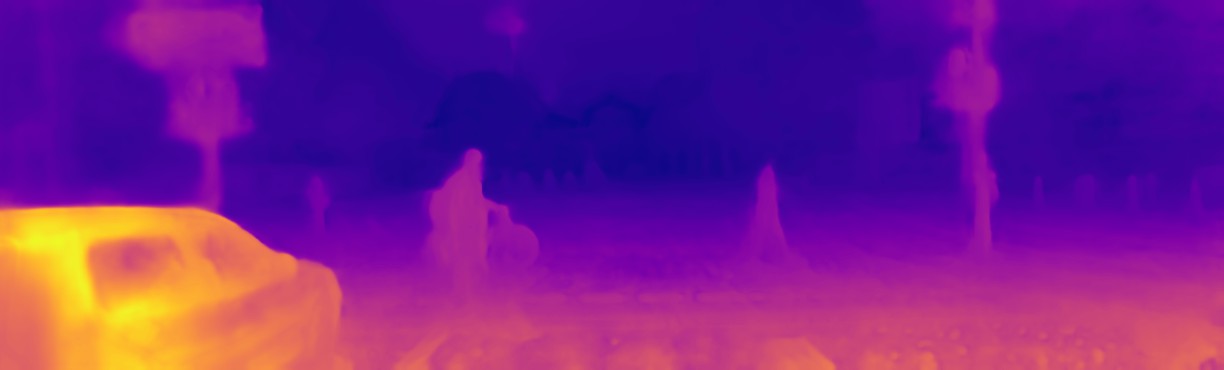}}
  \end{minipage}
  \begin{minipage}{0.2273\linewidth}
    \centerline{\includegraphics[width=\textwidth]{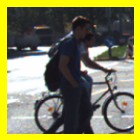}}
    \centerline{\includegraphics[width=\textwidth]{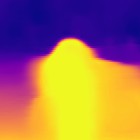}}
    \centerline{\includegraphics[width=\textwidth]{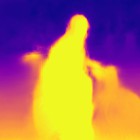}}
    \centerline{\includegraphics[width=\textwidth]{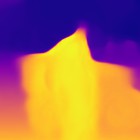}}
    \centerline{\includegraphics[width=\textwidth]{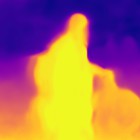}}
  \end{minipage}
  \centerline{(c)}
  \vspace{1pt}
  
  \begin{minipage}{0.7527\linewidth}
    \centerline{\includegraphics[width=\textwidth]{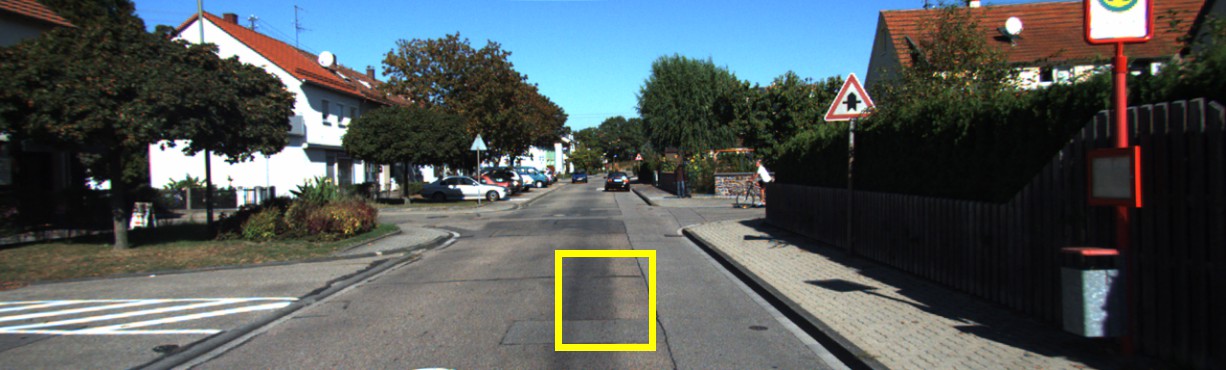}}
    \centerline{\includegraphics[width=\textwidth]{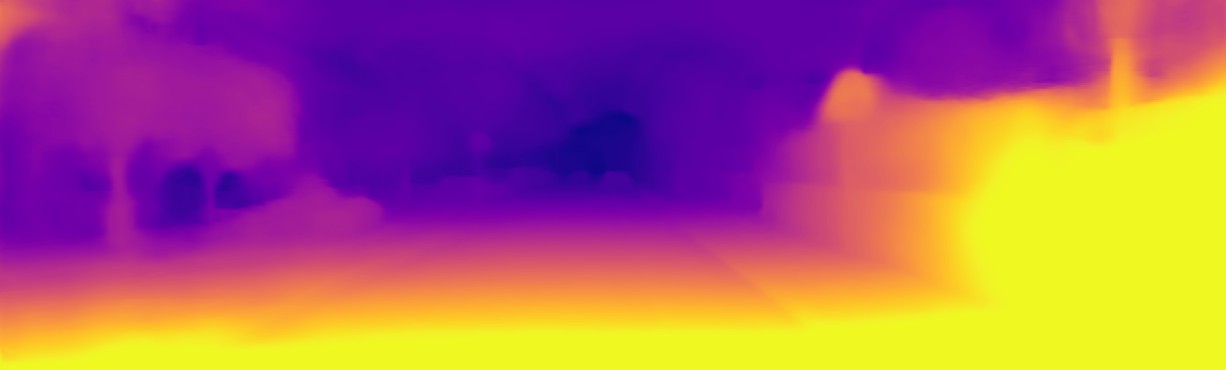}}
    \centerline{\includegraphics[width=\textwidth]{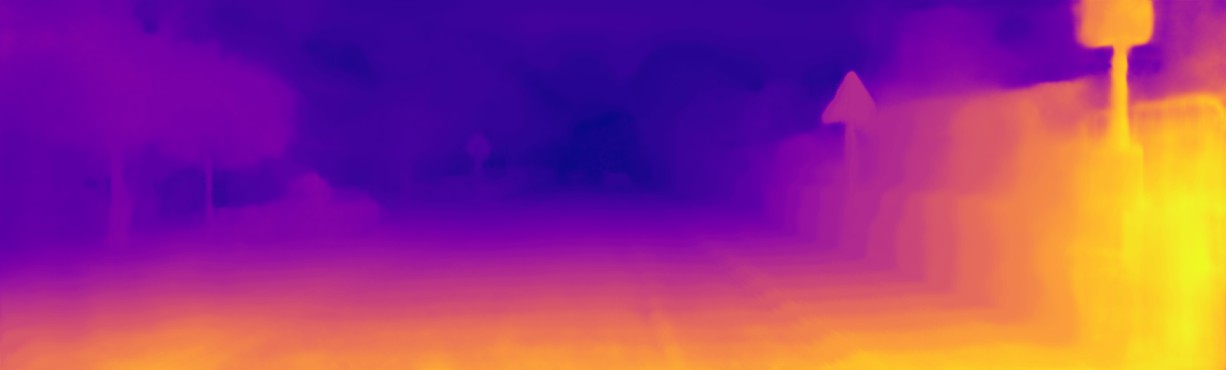}}
    \centerline{\includegraphics[width=\textwidth]{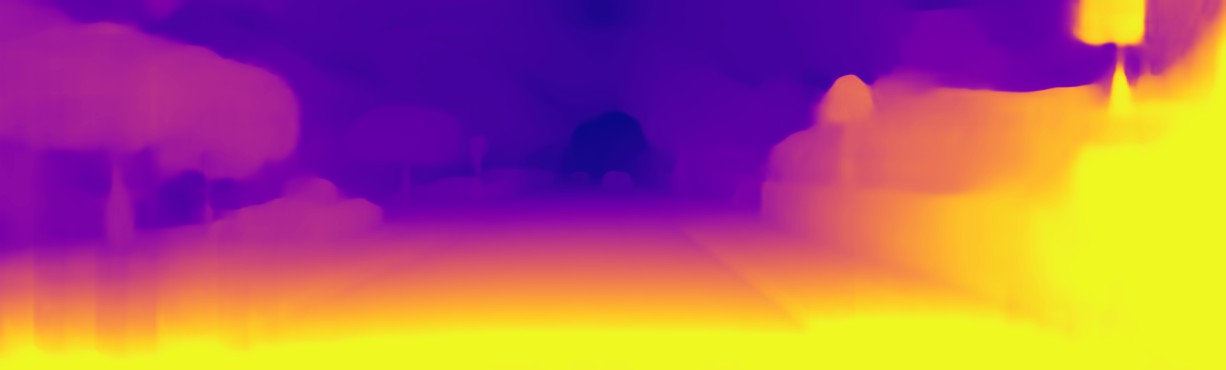}}
    \centerline{\includegraphics[width=\textwidth]{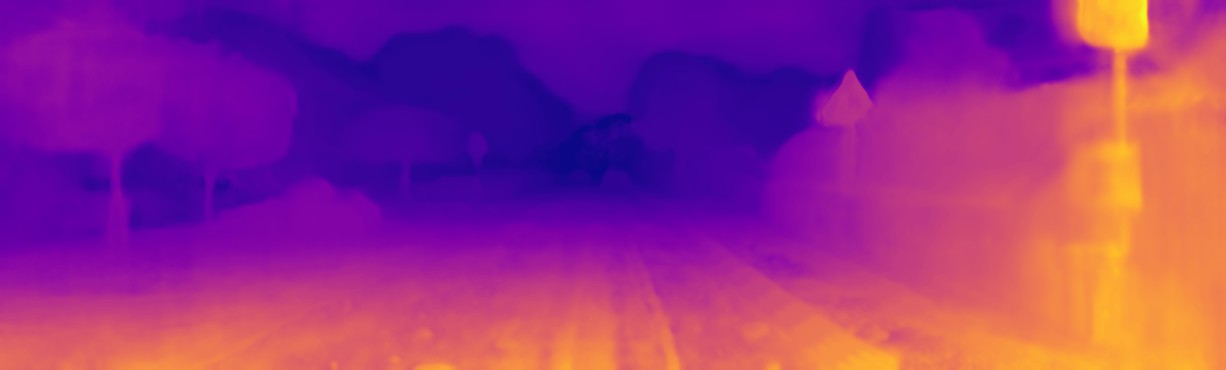}}
  \end{minipage}
  \begin{minipage}{0.2273\linewidth}
    \centerline{\includegraphics[width=\textwidth]{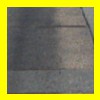}}
    \centerline{\includegraphics[width=\textwidth]{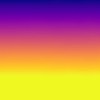}}
    \centerline{\includegraphics[width=\textwidth]{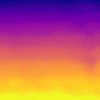}}
    \centerline{\includegraphics[width=\textwidth]{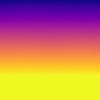}}
    \centerline{\includegraphics[width=\textwidth]{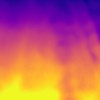}}
  \end{minipage}
  \centerline{(f)}
\end{minipage}

  \caption{
  Visualization results of FAL-Arc and Res-Arc with CDC and DDC on KITTI~\cite{Geiger2012We}.
  For showing the differences of the predicted depth maps more clearly, the images in the even columns are the enlarged versions of the yellow rectangle regions selected from the images in the odd columns, and the depth maps in the even columns are re-normalized.}
  \label{fig:add1}
\end{figure*}

Figure~\ref{fig:add1} shows the visualization results of the estimated depth maps by FAL-Arc and Res-Arc with the continuous depth constraint (CDC) and discrete depth constraint (DDC) on KITTI~\cite{Geiger2012We}.
These results reveal that the depth maps estimated by the two architectures with DDC preserve more detailed information than those with CDC ((a)-(c) in Figure~\ref{fig:add1}).
It could be also seen that the estimated depth maps by the two architectures with CDC are relatively smoother, while the estimated depth maps by the two architectures with DDC are relatively sharper ((d)-(f) in Figure~\ref{fig:add1}).

\section{Visualization results on the comparative evaluation}
\begin{figure*}[t]
  \centering
  \begin{minipage}{0.13\linewidth}
      \small
      \leftline{Input images}
      \leftline{(Local regions)}
      \vspace{20pt}

      \leftline{DepthHints~\cite{Watson2019Self}}
      \vspace{23pt}

      \leftline{FAL-Net~\cite{Gonzalezbello2020Forget}}
      \vspace{23pt}

      \leftline{OCFD-Net}
      \vspace{36pt}

      \leftline{Input images}
      \leftline{(Local regions)}
      \vspace{18pt}

      \leftline{DepthHints~\cite{Watson2019Self}}
      \vspace{23pt}

      \leftline{FAL-Net~\cite{Gonzalezbello2020Forget}}
      \vspace{23pt}

      \leftline{OCFD-Net}
      \vspace{36pt}

      \leftline{Input images}
      \leftline{(Local regions)}
      \vspace{20pt}

      \leftline{DepthHints~\cite{Watson2019Self}}
      \vspace{23pt}

      \leftline{FAL-Net~\cite{Gonzalezbello2020Forget}}
      \vspace{23pt}

      \leftline{OCFD-Net}
      \vspace{36pt}

      \leftline{Input images}
      \leftline{(Local regions)}
      \vspace{18pt}

      \leftline{DepthHints~\cite{Watson2019Self}}
      \vspace{23pt}

      \leftline{FAL-Net~\cite{Gonzalezbello2020Forget}}
      \vspace{23pt}

      \leftline{OCFD-Net}
      \vspace{15pt}

  \end{minipage}
  \begin{minipage}{0.28\linewidth}
    \begin{minipage}{0.7527\linewidth}
      \centerline{\includegraphics[width=\textwidth]{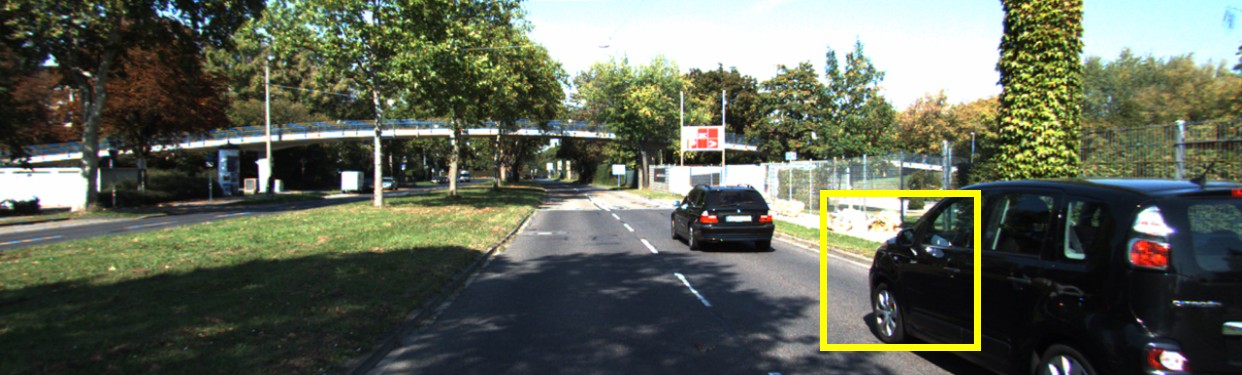}}
      \centerline{\includegraphics[width=\textwidth]{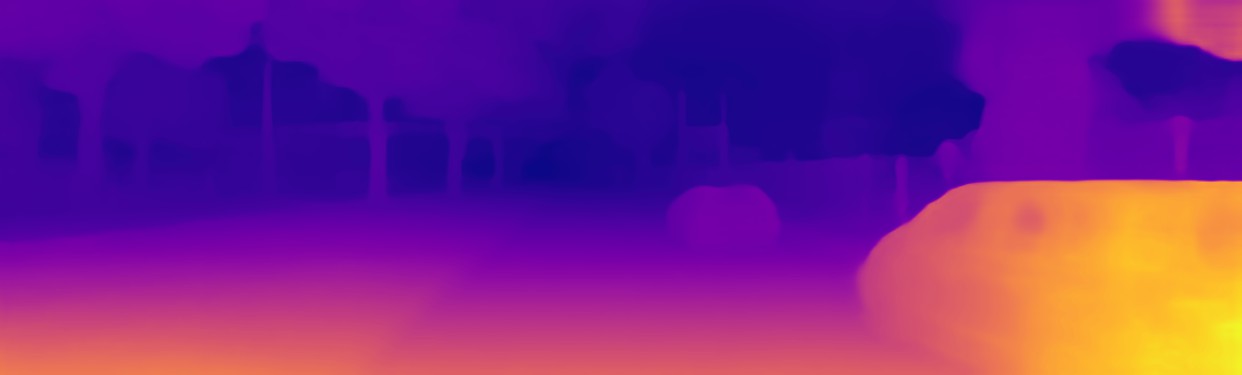}}
      \centerline{\includegraphics[width=\textwidth]{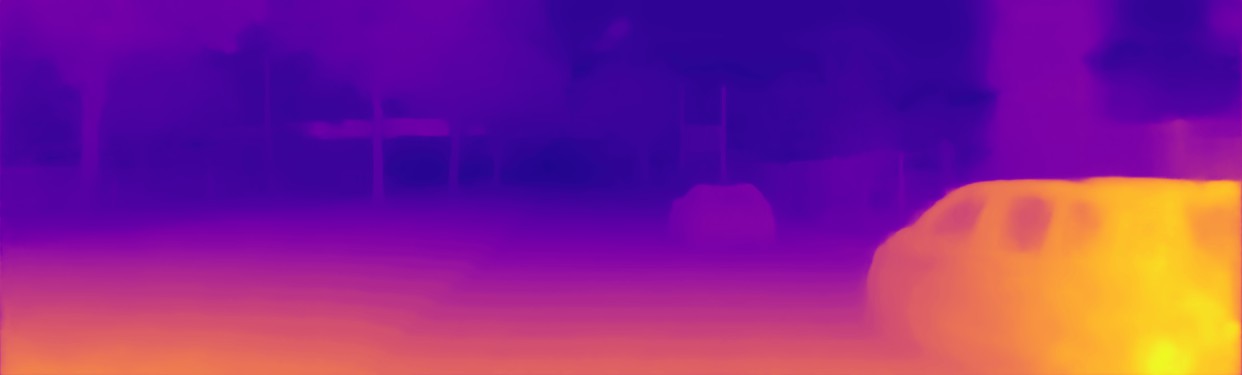}}
      \centerline{\includegraphics[width=\textwidth]{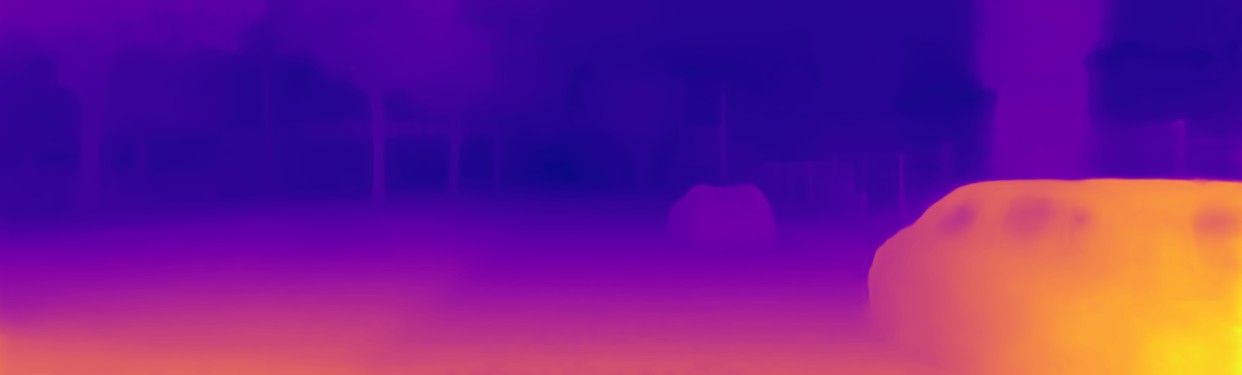}}
    \end{minipage}
    \begin{minipage}{0.2273\linewidth}
      \centerline{\includegraphics[width=\textwidth]{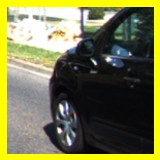}}
      \centerline{\includegraphics[width=\textwidth]{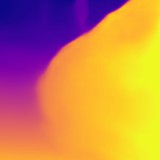}}
      \centerline{\includegraphics[width=\textwidth]{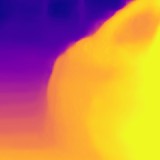}}
      \centerline{\includegraphics[width=\textwidth]{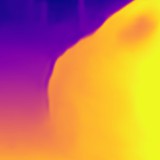}}
    \end{minipage}
    \centerline{(a)}
    \vspace{1pt}

    \begin{minipage}{0.7527\linewidth}
      \centerline{\includegraphics[width=\textwidth]{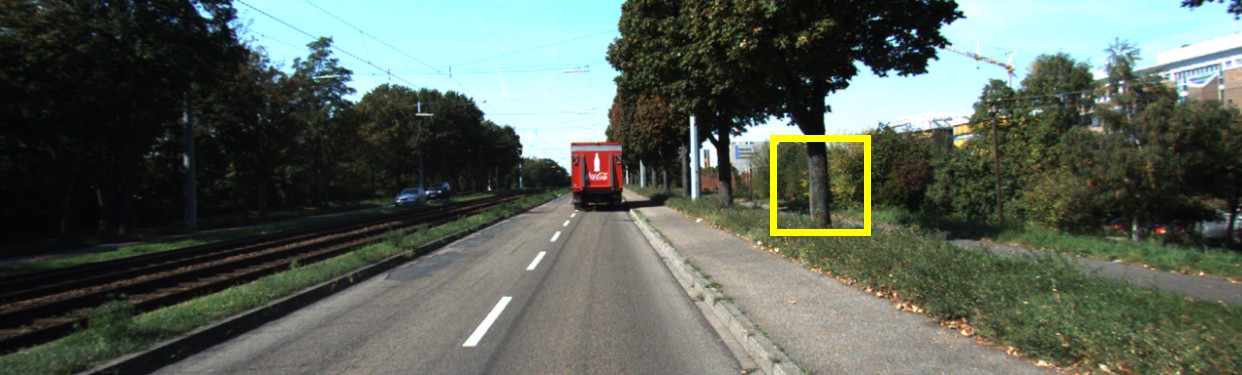}}
      \centerline{\includegraphics[width=\textwidth]{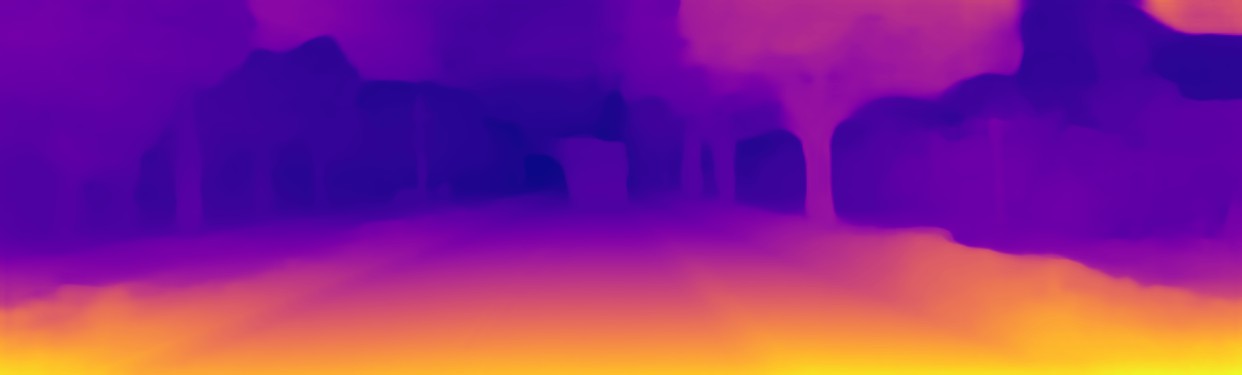}}
      \centerline{\includegraphics[width=\textwidth]{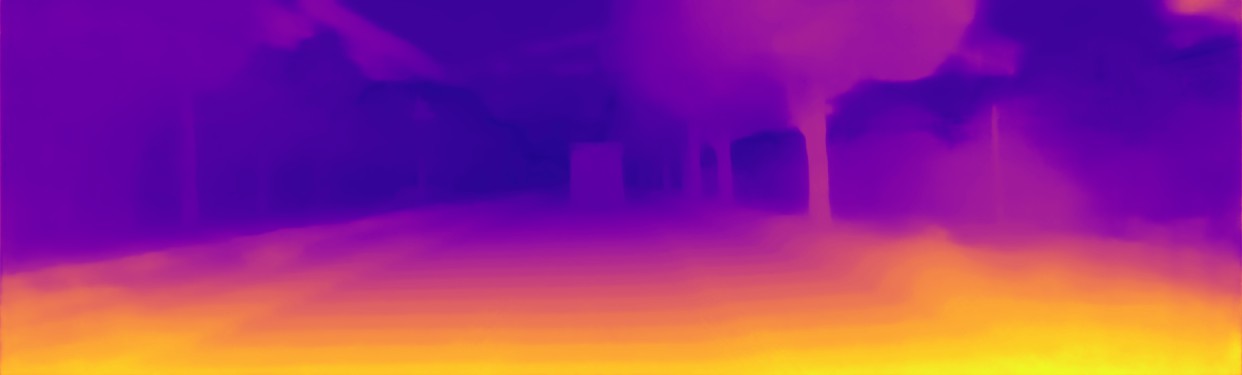}}
      \centerline{\includegraphics[width=\textwidth]{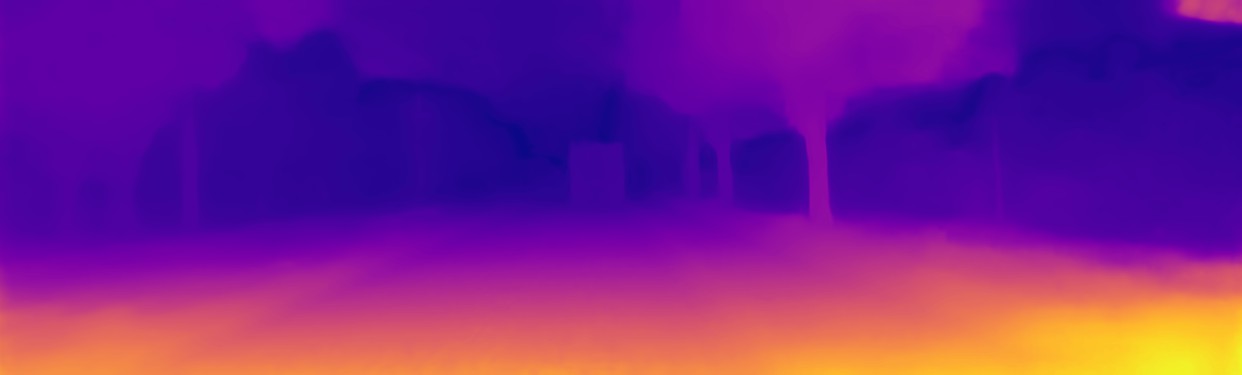}}
    \end{minipage}
    \begin{minipage}{0.2273\linewidth}
      \centerline{\includegraphics[width=\textwidth]{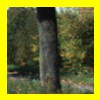}}
      \centerline{\includegraphics[width=\textwidth]{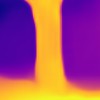}}
      \centerline{\includegraphics[width=\textwidth]{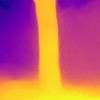}}
      \centerline{\includegraphics[width=\textwidth]{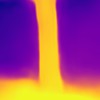}}
    \end{minipage}
    \centerline{(d)}
    \vspace{1pt}

    \begin{minipage}{0.7527\linewidth}
      \centerline{\includegraphics[width=\textwidth]{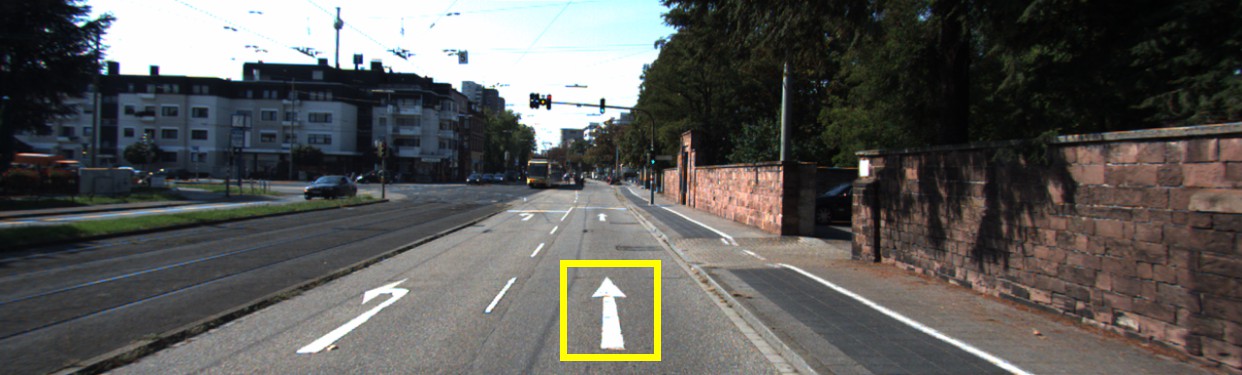}}
      \centerline{\includegraphics[width=\textwidth]{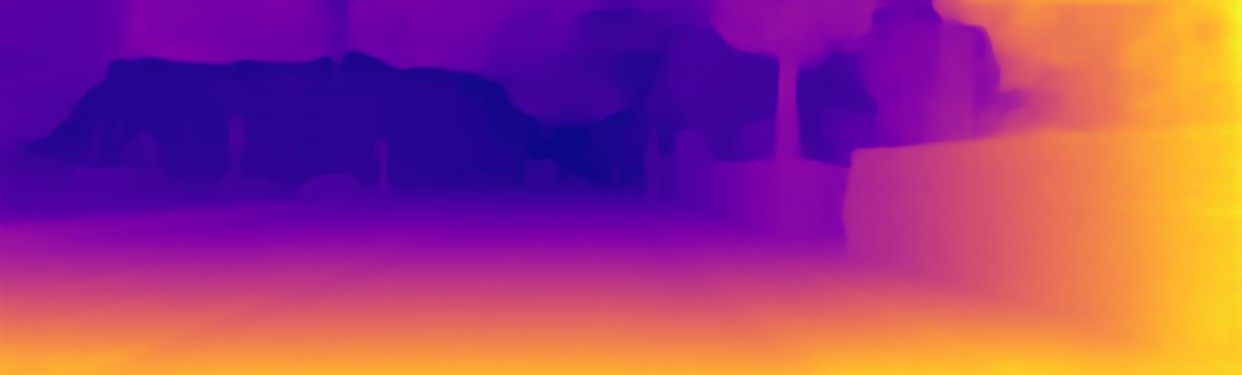}}
      \centerline{\includegraphics[width=\textwidth]{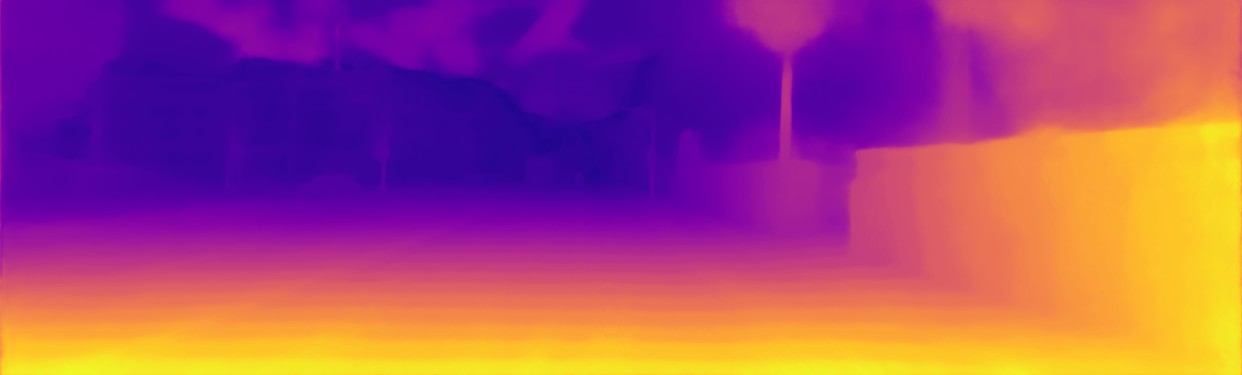}}
      \centerline{\includegraphics[width=\textwidth]{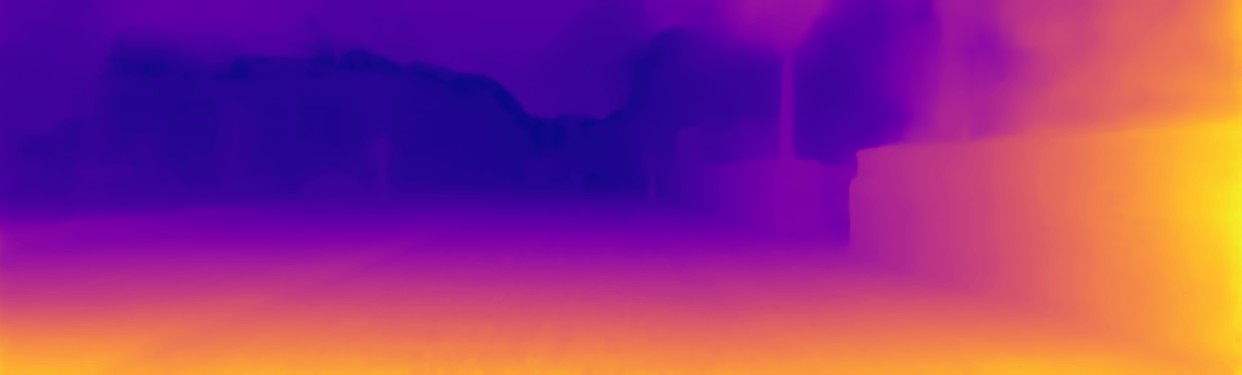}}
    \end{minipage}
    \begin{minipage}{0.2273\linewidth}
      \centerline{\includegraphics[width=\textwidth]{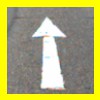}}
      \centerline{\includegraphics[width=\textwidth]{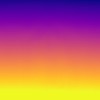}}
      \centerline{\includegraphics[width=\textwidth]{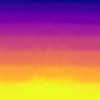}}
      \centerline{\includegraphics[width=\textwidth]{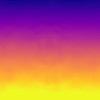}}
    \end{minipage}
    \centerline{(g)}
    \vspace{1pt}

    \begin{minipage}{0.7527\linewidth}
      \centerline{\includegraphics[width=\textwidth]{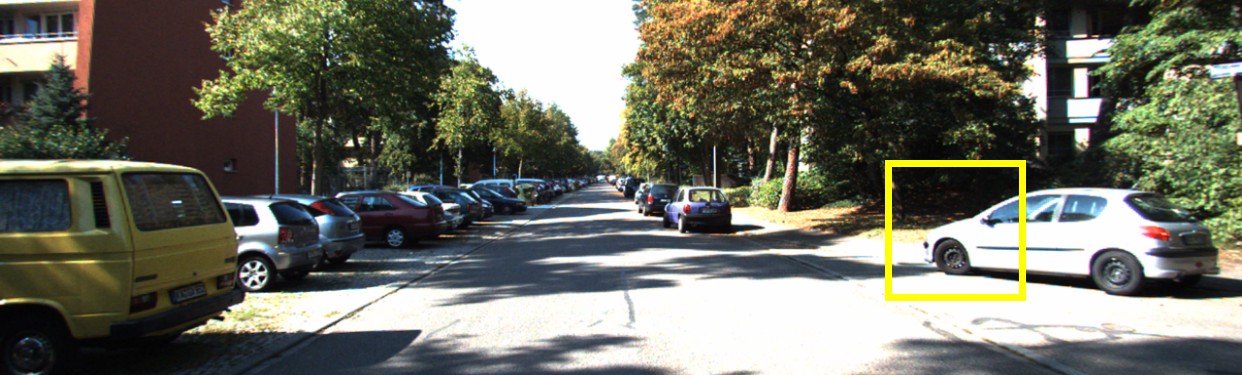}}
      \centerline{\includegraphics[width=\textwidth]{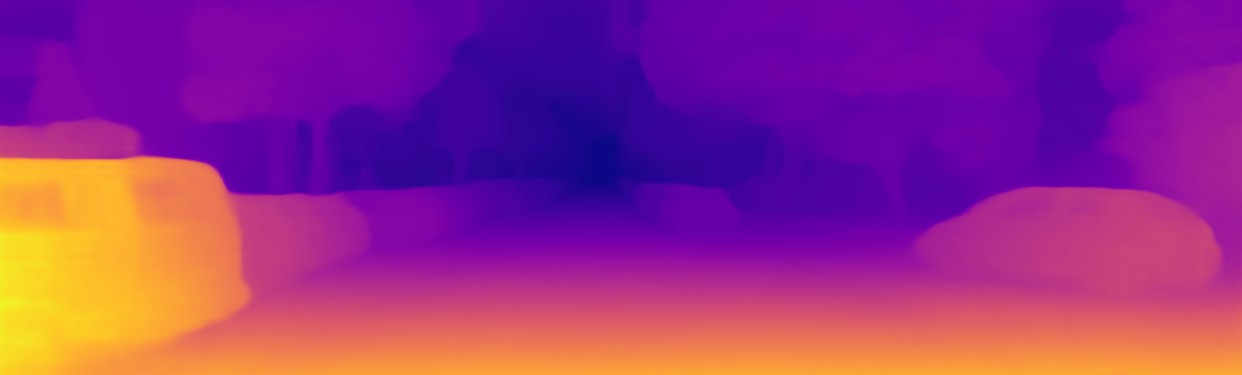}}
      \centerline{\includegraphics[width=\textwidth]{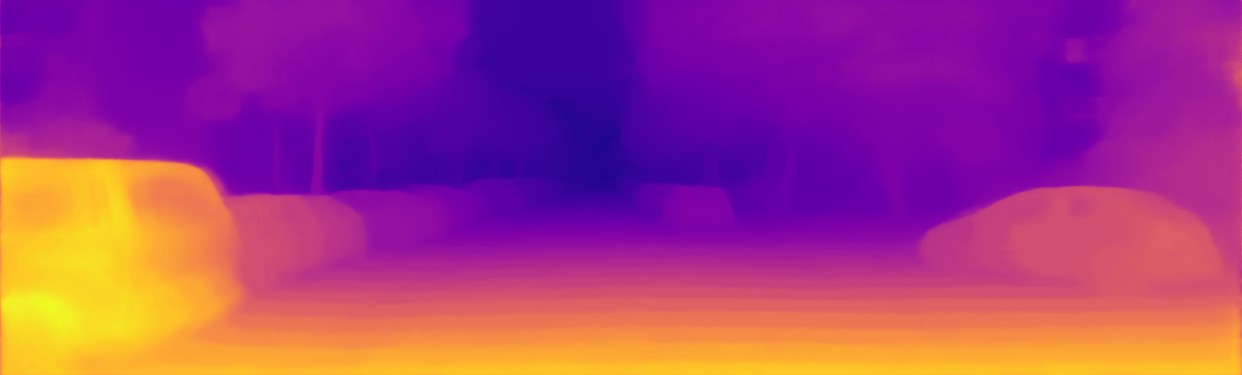}}
      \centerline{\includegraphics[width=\textwidth]{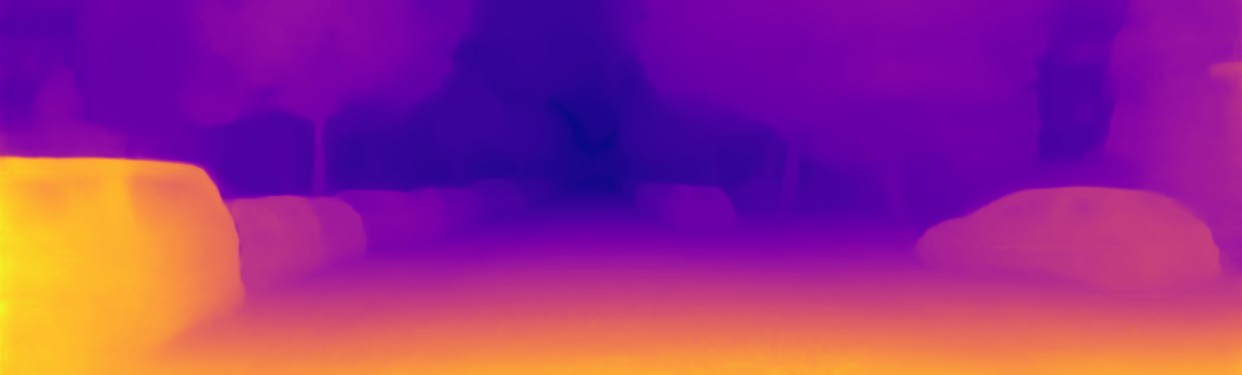}}
    \end{minipage}
    \begin{minipage}{0.2273\linewidth}
      \centerline{\includegraphics[width=\textwidth]{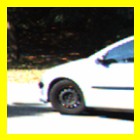}}
      \centerline{\includegraphics[width=\textwidth]{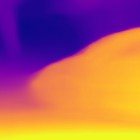}}
      \centerline{\includegraphics[width=\textwidth]{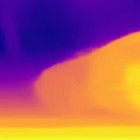}}
      \centerline{\includegraphics[width=\textwidth]{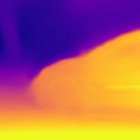}}
    \end{minipage}
    \centerline{(j)}
\end{minipage}
\begin{minipage}{0.28\linewidth}
  \begin{minipage}{0.7527\linewidth}
    \centerline{\includegraphics[width=\textwidth]{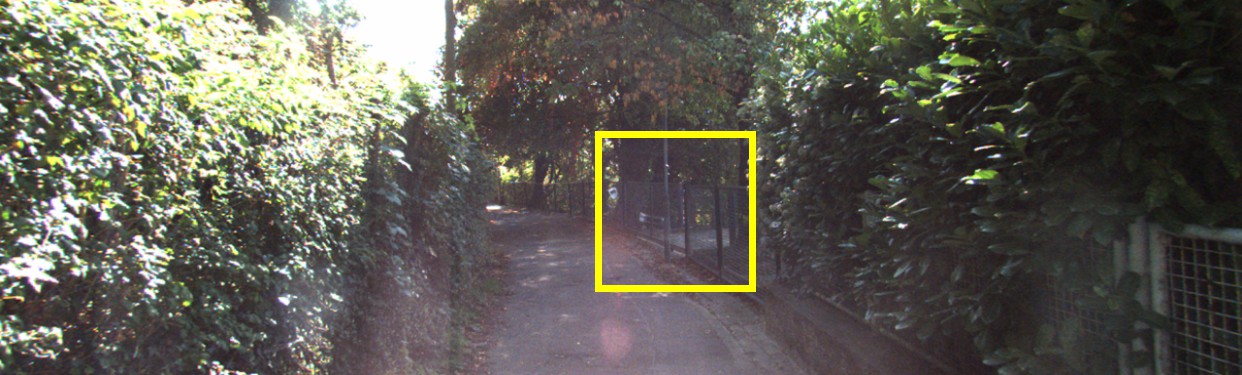}}
    \centerline{\includegraphics[width=\textwidth]{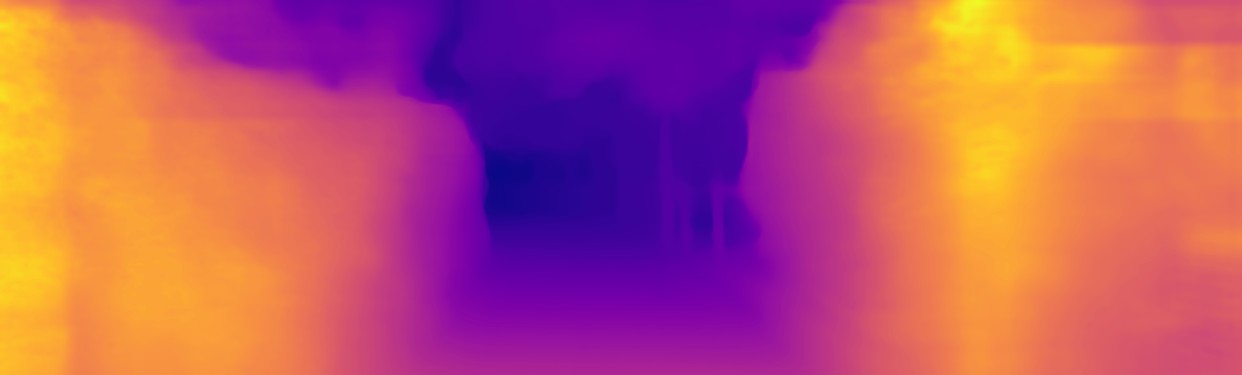}}
    \centerline{\includegraphics[width=\textwidth]{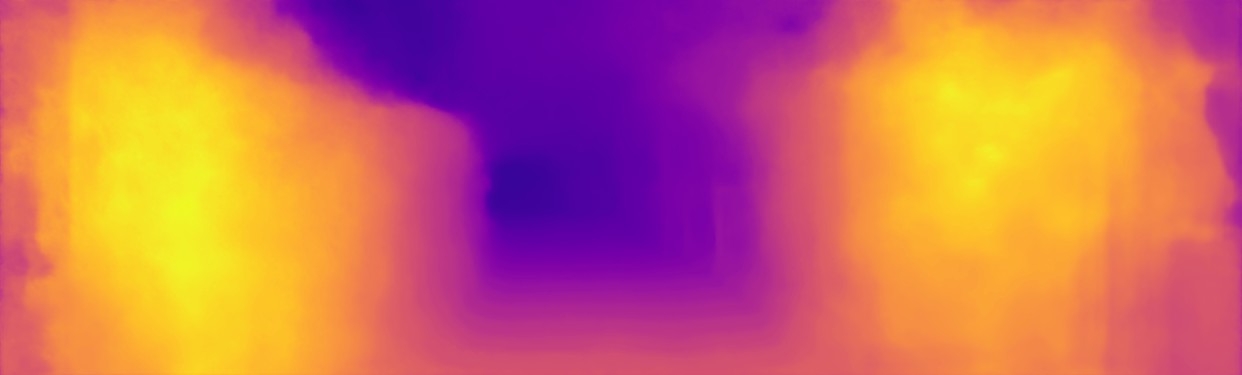}}
    \centerline{\includegraphics[width=\textwidth]{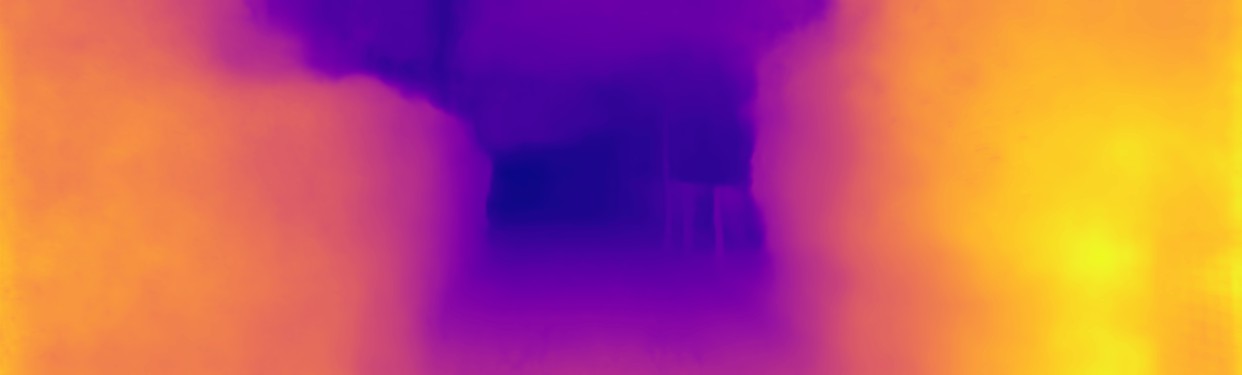}}
  \end{minipage}
  \begin{minipage}{0.2273\linewidth}
    \centerline{\includegraphics[width=\textwidth]{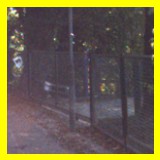}}
    \centerline{\includegraphics[width=\textwidth]{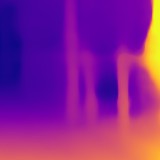}}
    \centerline{\includegraphics[width=\textwidth]{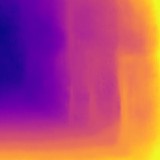}}
    \centerline{\includegraphics[width=\textwidth]{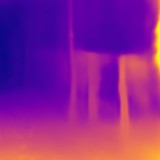}}
  \end{minipage}
  \centerline{(b)}
  \vspace{1pt}

  \begin{minipage}{0.7527\linewidth}
    \centerline{\includegraphics[width=\textwidth]{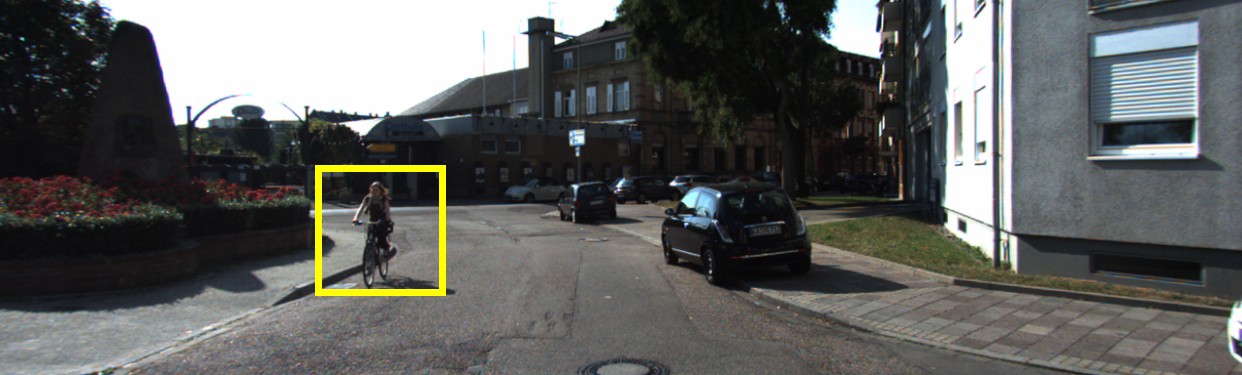}}
    \centerline{\includegraphics[width=\textwidth]{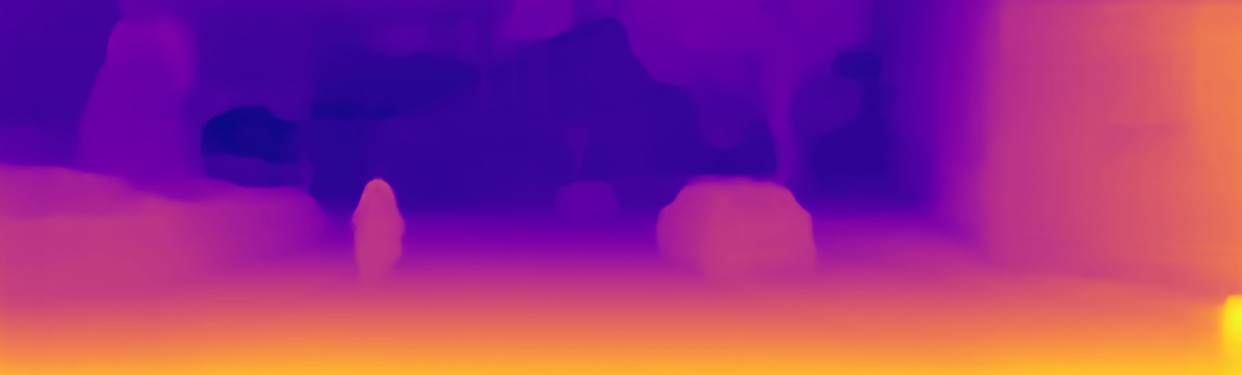}}
    \centerline{\includegraphics[width=\textwidth]{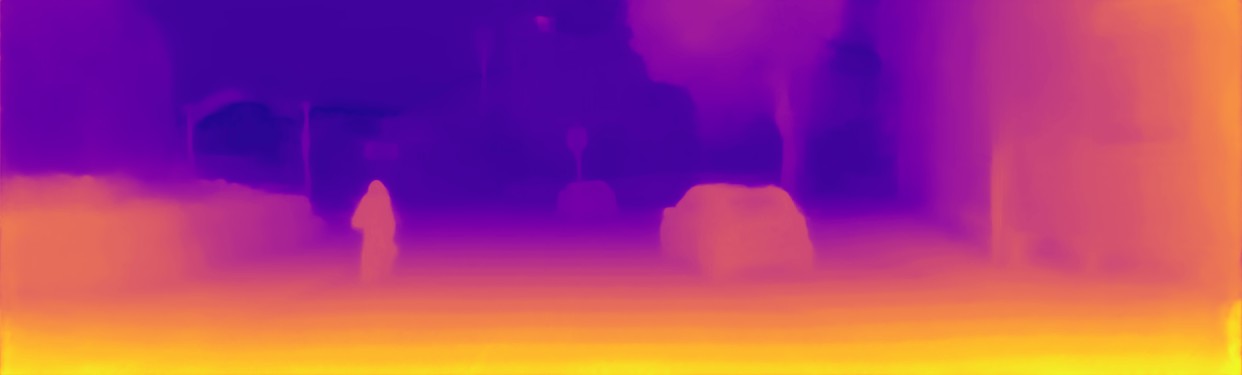}}
    \centerline{\includegraphics[width=\textwidth]{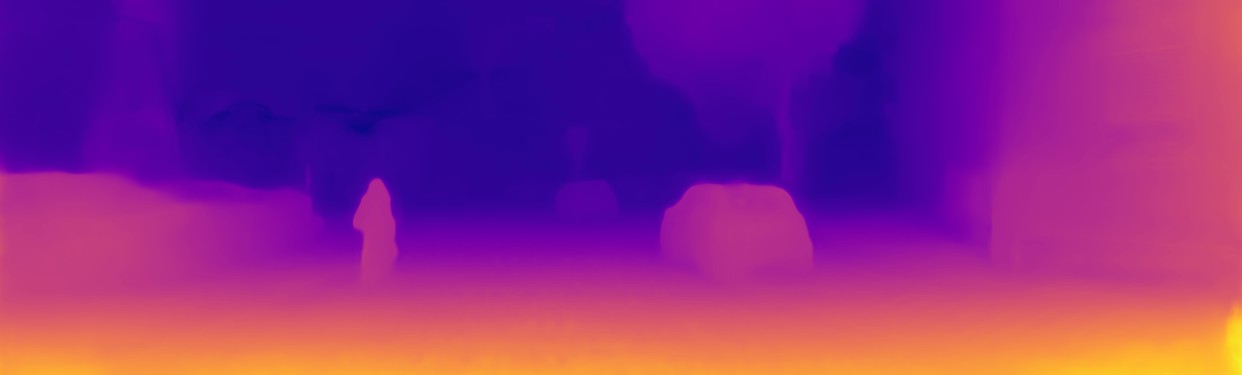}}
  \end{minipage}
  \begin{minipage}{0.2273\linewidth}
    \centerline{\includegraphics[width=\textwidth]{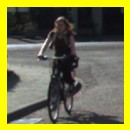}}
    \centerline{\includegraphics[width=\textwidth]{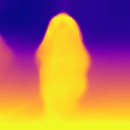}}
    \centerline{\includegraphics[width=\textwidth]{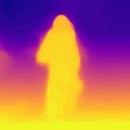}}
    \centerline{\includegraphics[width=\textwidth]{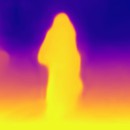}}
  \end{minipage}
  \centerline{(e)}
  \vspace{1pt}

  \begin{minipage}{0.7527\linewidth}
    \centerline{\includegraphics[width=\textwidth]{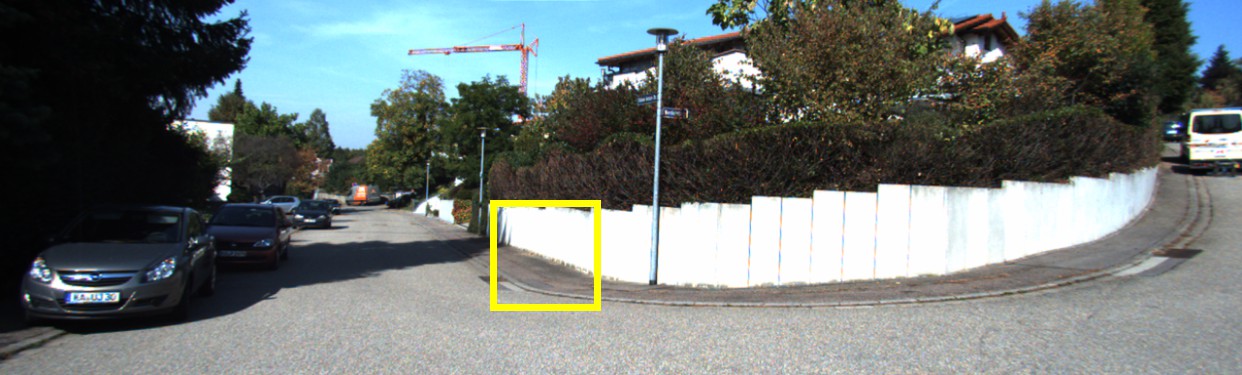}}
    \centerline{\includegraphics[width=\textwidth]{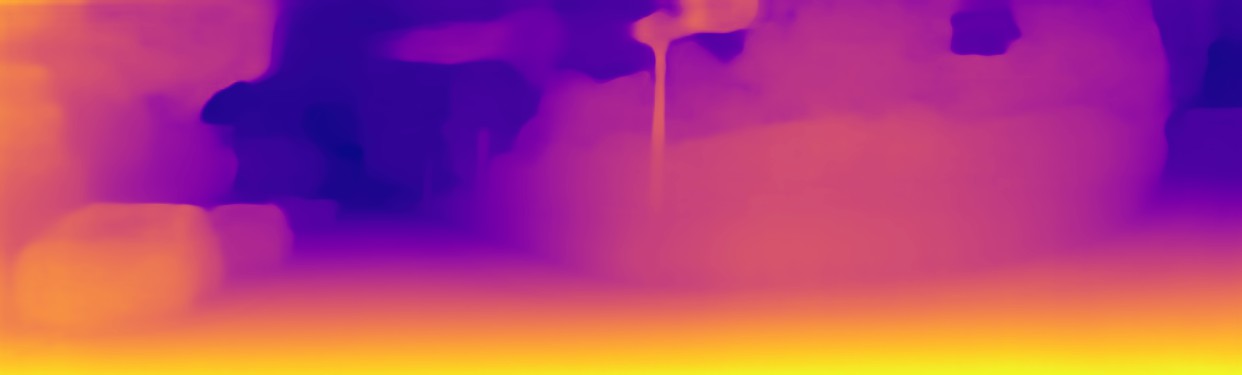}}
    \centerline{\includegraphics[width=\textwidth]{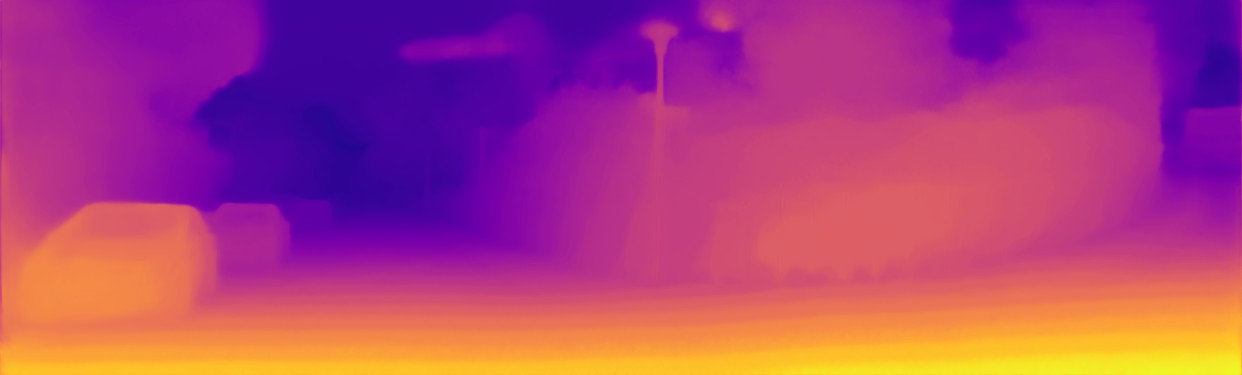}}
    \centerline{\includegraphics[width=\textwidth]{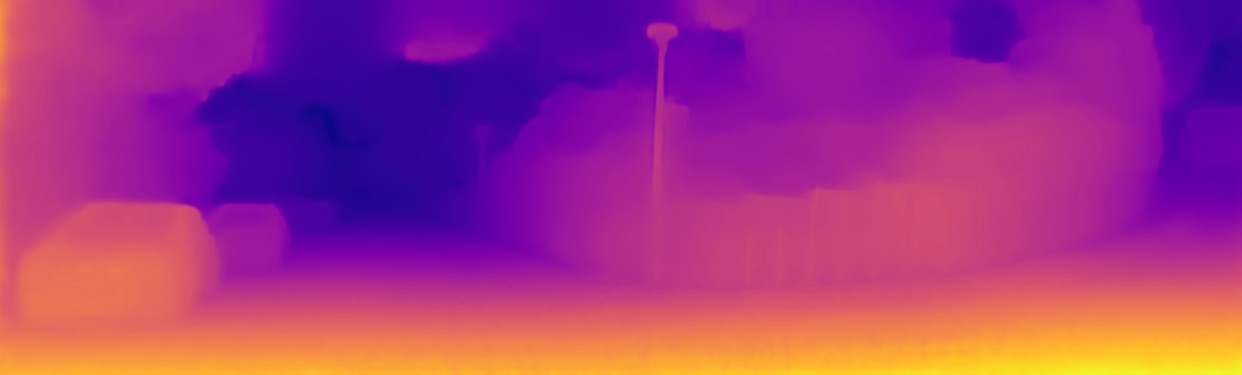}}
  \end{minipage}
  \begin{minipage}{0.2273\linewidth}
    \centerline{\includegraphics[width=\textwidth]{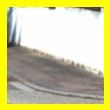}}
    \centerline{\includegraphics[width=\textwidth]{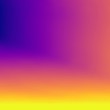}}
    \centerline{\includegraphics[width=\textwidth]{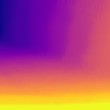}}
    \centerline{\includegraphics[width=\textwidth]{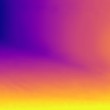}}
  \end{minipage}
  \centerline{(h)}
  \vspace{1pt}

  \begin{minipage}{0.7527\linewidth}
    \centerline{\includegraphics[width=\textwidth]{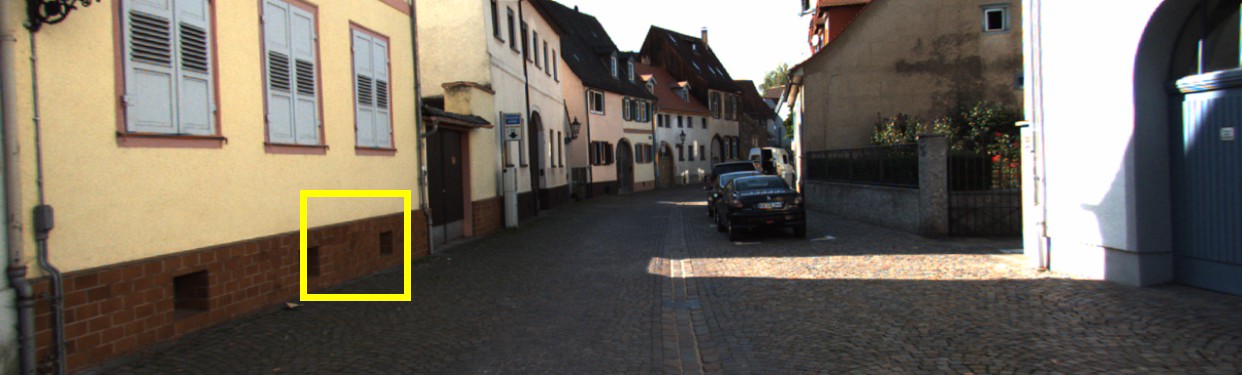}}
    \centerline{\includegraphics[width=\textwidth]{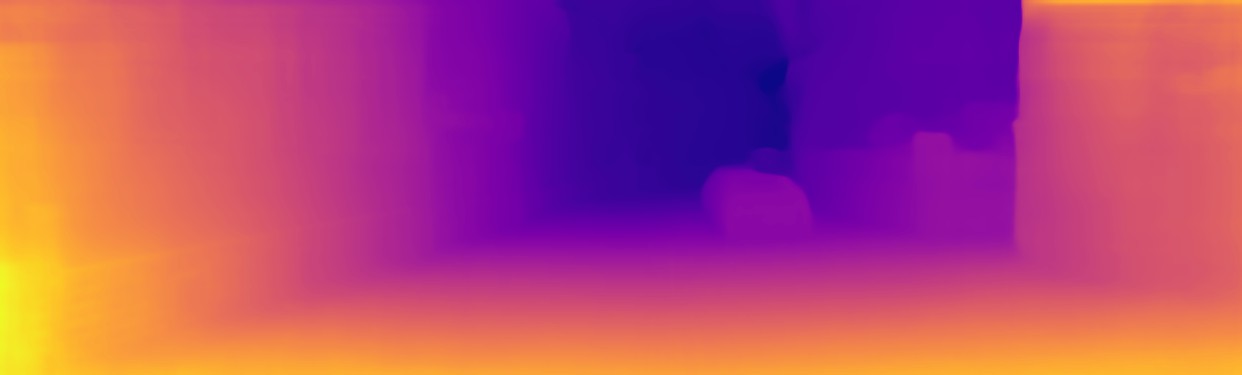}}
    \centerline{\includegraphics[width=\textwidth]{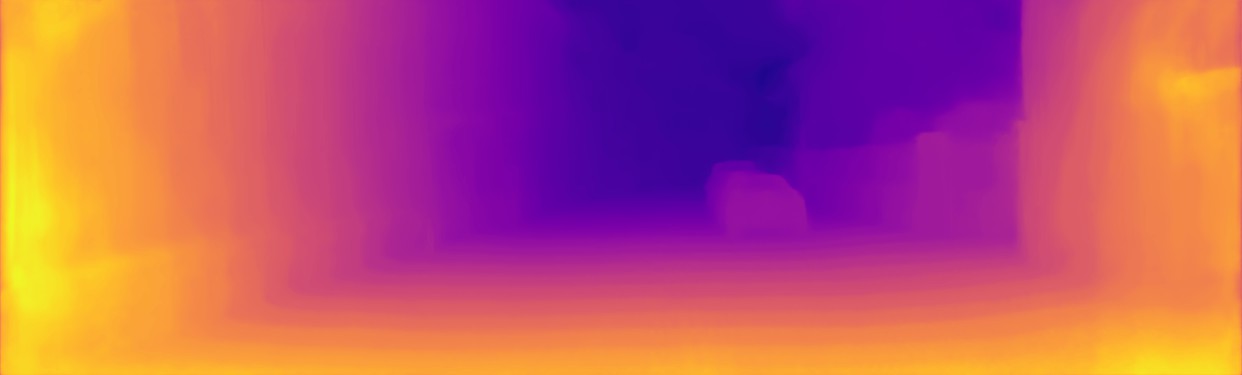}}
    \centerline{\includegraphics[width=\textwidth]{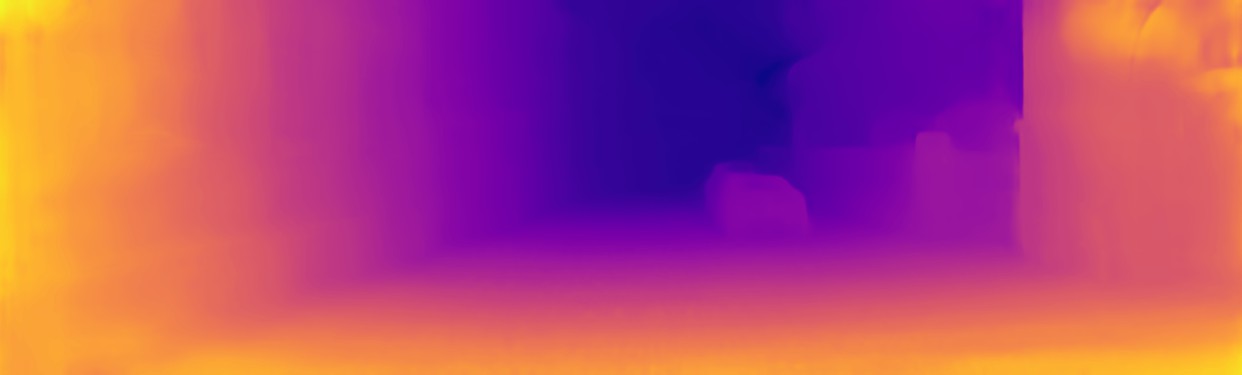}}
  \end{minipage}
  \begin{minipage}{0.2273\linewidth}
    \centerline{\includegraphics[width=\textwidth]{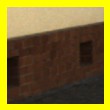}}
    \centerline{\includegraphics[width=\textwidth]{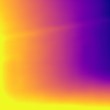}}
    \centerline{\includegraphics[width=\textwidth]{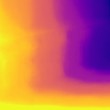}}
    \centerline{\includegraphics[width=\textwidth]{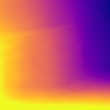}}
  \end{minipage}
  \centerline{(k)}
\end{minipage}
\begin{minipage}{0.28\linewidth}
\begin{minipage}{0.7527\linewidth}
  \centerline{\includegraphics[width=\textwidth]{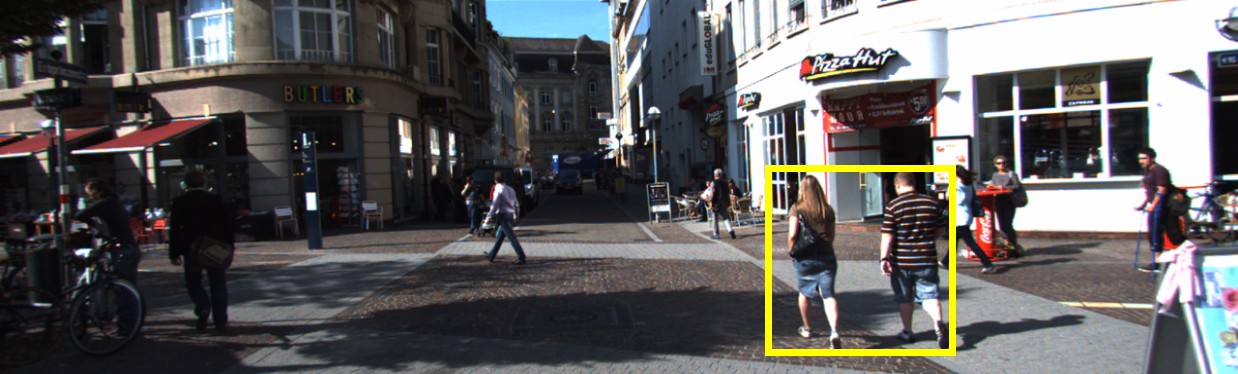}}
    \centerline{\includegraphics[width=\textwidth]{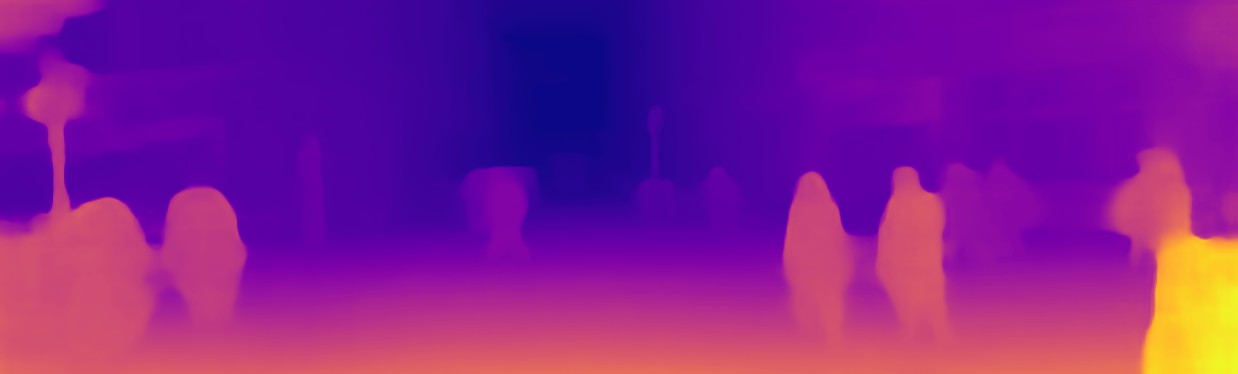}}
    \centerline{\includegraphics[width=\textwidth]{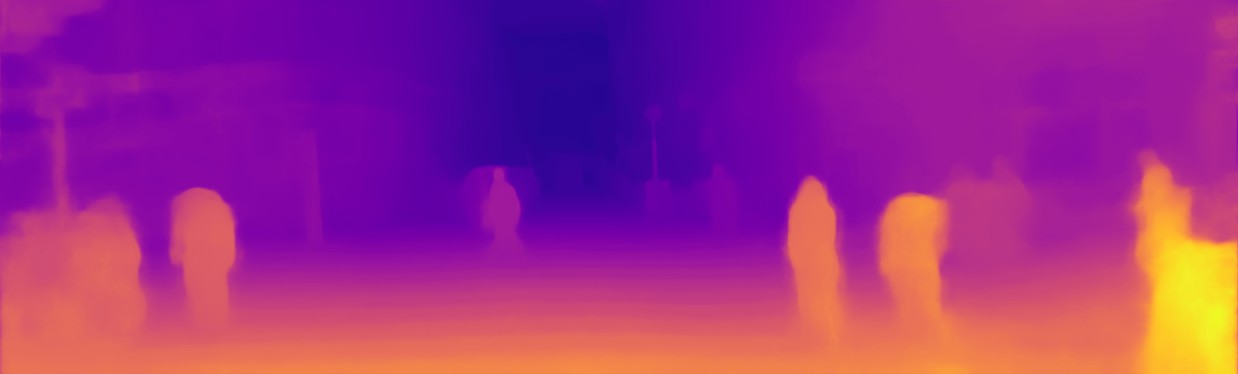}}
    \centerline{\includegraphics[width=\textwidth]{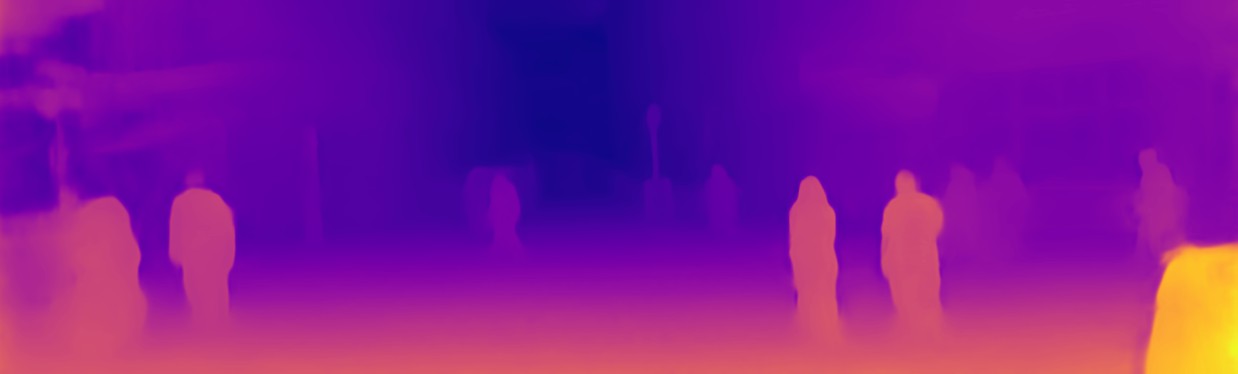}}
\end{minipage}
\begin{minipage}{0.2273\linewidth}
  \centerline{\includegraphics[width=\textwidth]{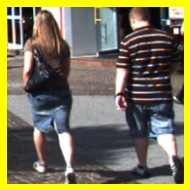}}
  \centerline{\includegraphics[width=\textwidth]{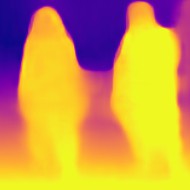}}
  \centerline{\includegraphics[width=\textwidth]{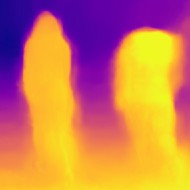}}
  \centerline{\includegraphics[width=\textwidth]{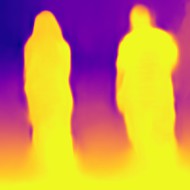}}
\end{minipage}
\centerline{(c)}
\vspace{1pt}

\begin{minipage}{0.7527\linewidth}
  \centerline{\includegraphics[width=\textwidth]{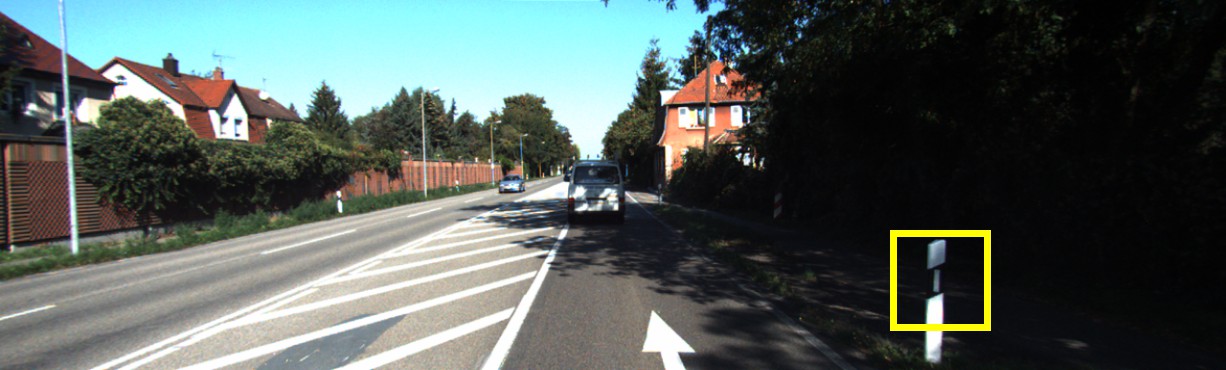}}
  \centerline{\includegraphics[width=\textwidth]{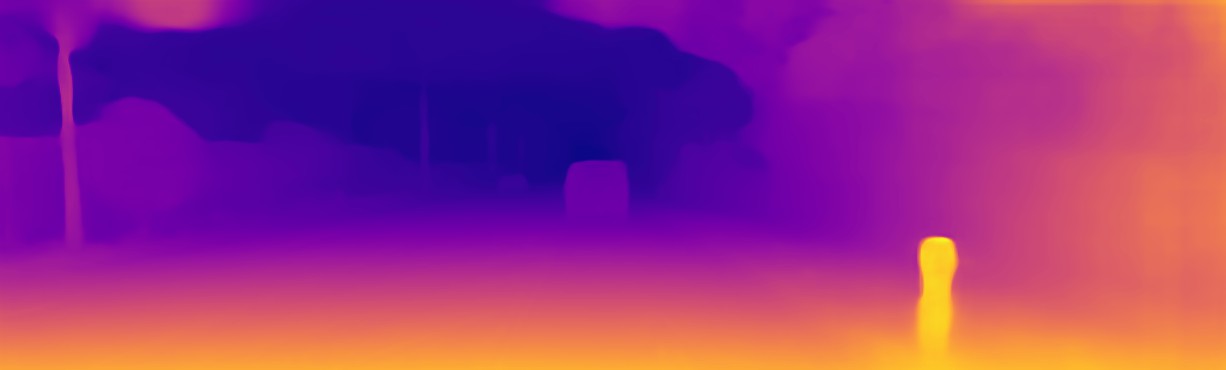}}
  \centerline{\includegraphics[width=\textwidth]{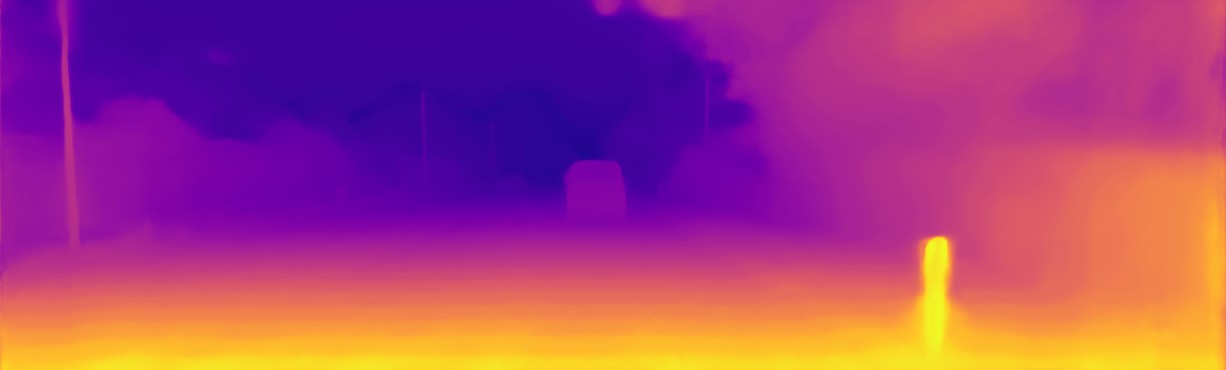}}
  \centerline{\includegraphics[width=\textwidth]{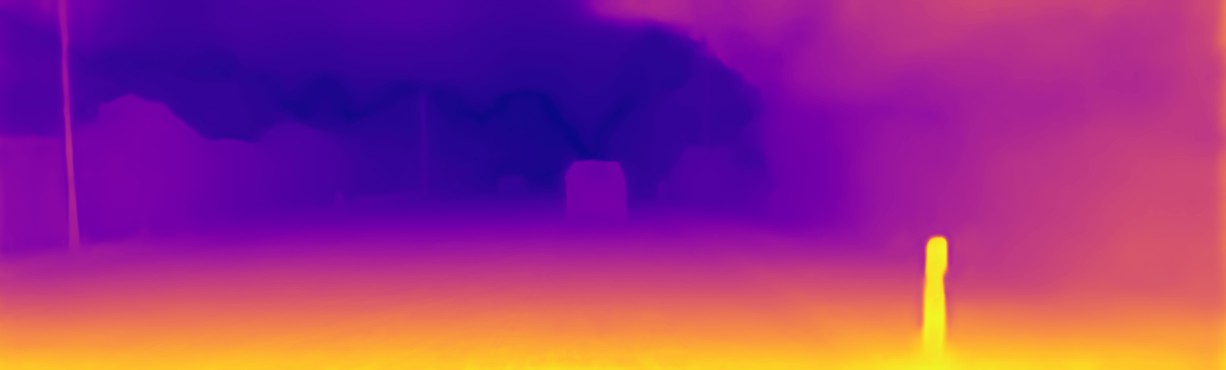}}
\end{minipage}
\begin{minipage}{0.2273\linewidth}
  \centerline{\includegraphics[width=\textwidth]{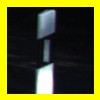}}
  \centerline{\includegraphics[width=\textwidth]{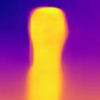}}
  \centerline{\includegraphics[width=\textwidth]{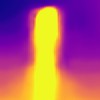}}
  \centerline{\includegraphics[width=\textwidth]{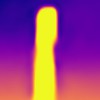}}
\end{minipage}
\centerline{(f)}
\vspace{1pt}

\begin{minipage}{0.7527\linewidth}
  \centerline{\includegraphics[width=\textwidth]{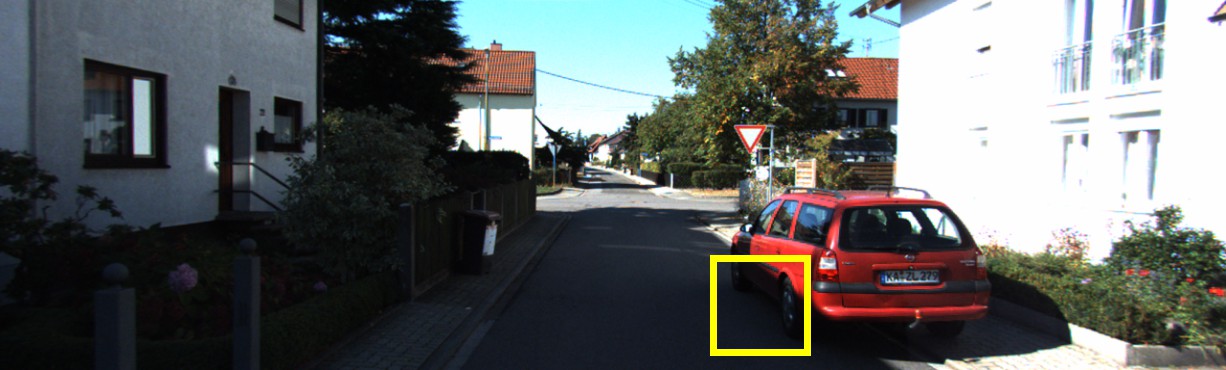}}
  \centerline{\includegraphics[width=\textwidth]{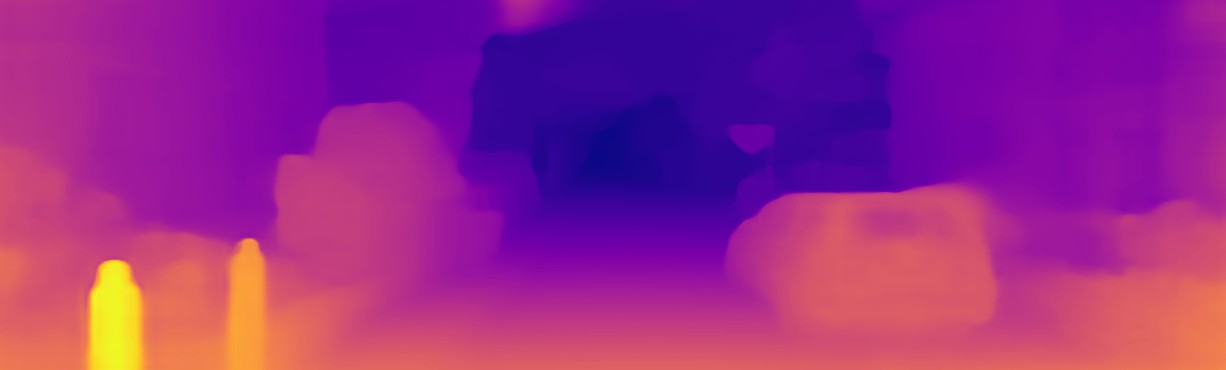}}
  \centerline{\includegraphics[width=\textwidth]{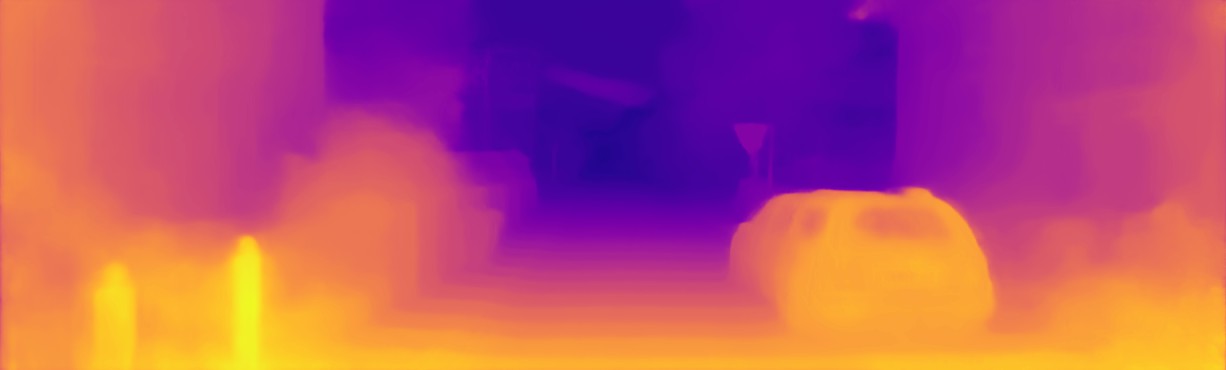}}
  \centerline{\includegraphics[width=\textwidth]{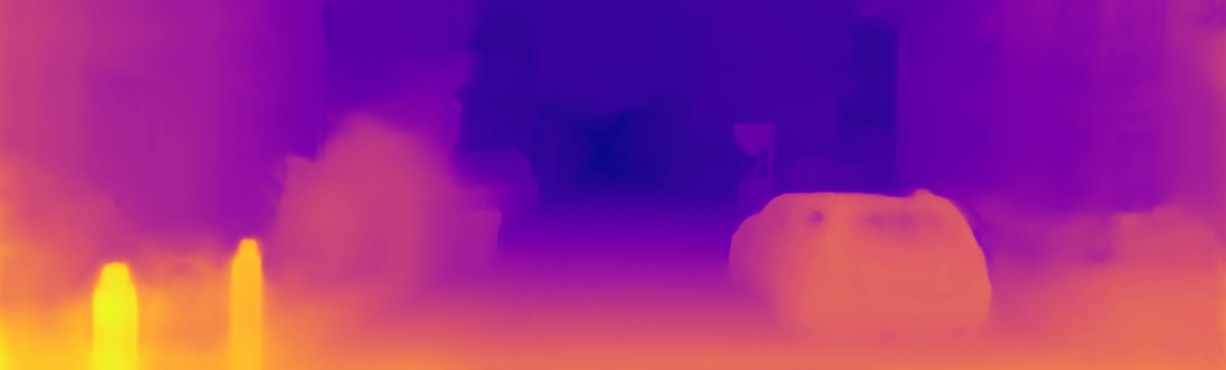}}
\end{minipage}
\begin{minipage}{0.2273\linewidth}
  \centerline{\includegraphics[width=\textwidth]{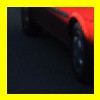}}
  \centerline{\includegraphics[width=\textwidth]{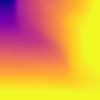}}
  \centerline{\includegraphics[width=\textwidth]{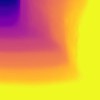}}
  \centerline{\includegraphics[width=\textwidth]{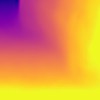}}
\end{minipage}
\centerline{(i)}
\vspace{1pt}

\begin{minipage}{0.7527\linewidth}
  \centerline{\includegraphics[width=\textwidth]{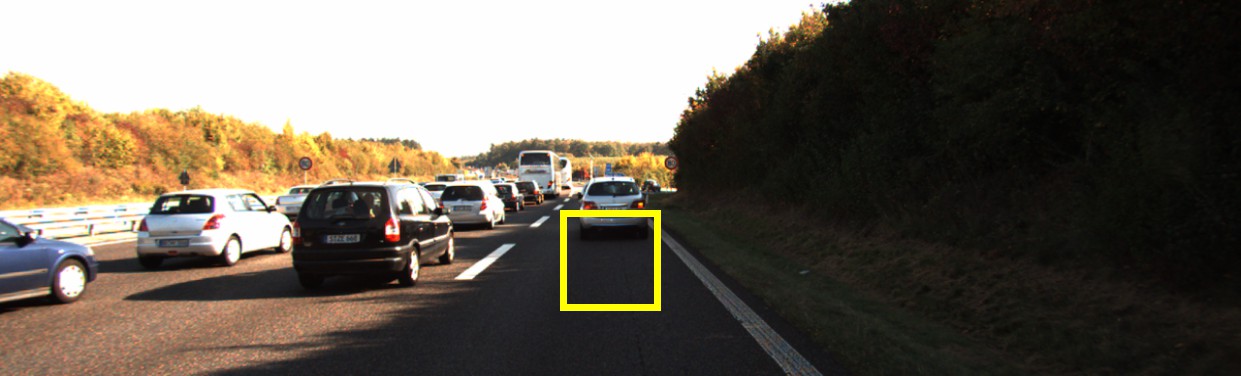}}
  \centerline{\includegraphics[width=\textwidth]{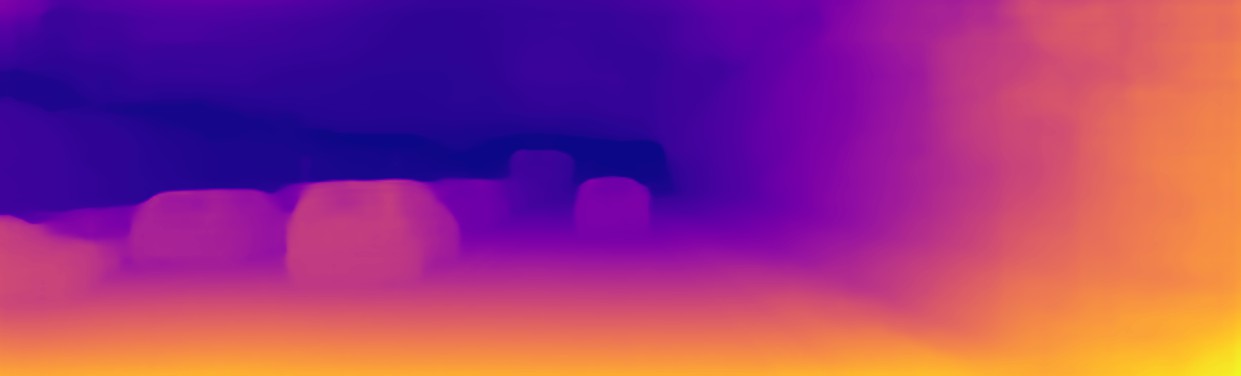}}
  \centerline{\includegraphics[width=\textwidth]{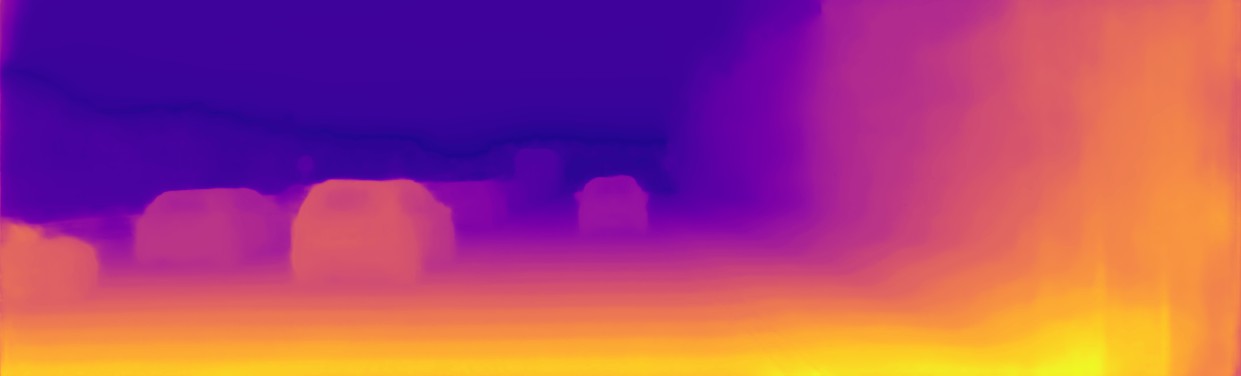}}
  \centerline{\includegraphics[width=\textwidth]{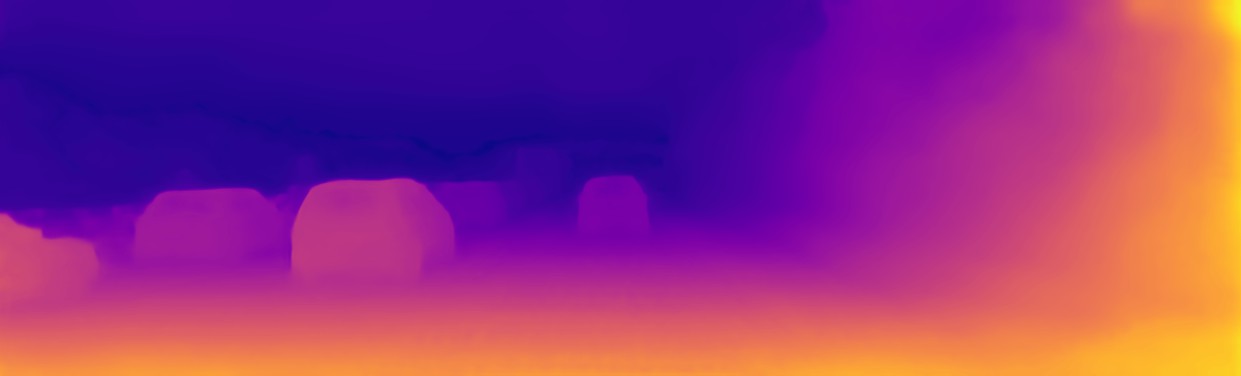}}
\end{minipage}
\begin{minipage}{0.2273\linewidth}
  \centerline{\includegraphics[width=\textwidth]{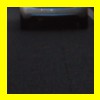}}
  \centerline{\includegraphics[width=\textwidth]{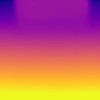}}
  \centerline{\includegraphics[width=\textwidth]{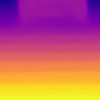}}
  \centerline{\includegraphics[width=\textwidth]{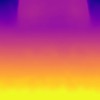}}
\end{minipage}
\centerline{(l)}
\end{minipage}
  \caption{
  Visualization results of DepthHints~\cite{Watson2019Self}, FAL-Net~\cite{Gonzalezbello2020Forget}, and our OCFD-Net on KITTI~\cite{Geiger2012We}.
  For showing the differences of the predicted depth maps more clearly, the images in the even columns are the enlarged versions of the yellow rectangle regions selected from the images in the odd columns, and the depth maps in the even columns are re-normalized.}
  \label{fig:add2}
\end{figure*}

Figure~\ref{fig:add2} shows the visualization results of OCFD-Net as well as two comparative methods, DepthHints~\cite{Watson2019Self} and FAL-Net~\cite{Gonzalezbello2020Forget}.
It can be seen that DepthHints predicts inaccurate depths on the regions close to object boundaries ((a)-(f) in Figure~\ref{fig:add2}), FAL-Net predicts unsmooth depths on the flat regions ((g)-(l) in Figure~\ref{fig:add2}), but our OCFD-Net could handle both the two cases effectively.

\section{Visualization results on the effects of the depth residual prediction
branch}
\begin{figure*}[t]
  \centering
  \begin{minipage}{0.13\linewidth}
      \small
      \leftline{Input images}
      \leftline{(Local regions)}
      \vspace{18pt}

      \leftline{$D^l_c$}
      \vspace{23pt}

      \leftline{$D^l_{res}$}
      \vspace{23pt}

      \leftline{$D^l_{f}$}
      \vspace{38pt}

      \leftline{Input images}
      \leftline{(Local regions)}
      \vspace{18pt}

      \leftline{$D^l_c$}
      \vspace{23pt}

      \leftline{$D^l_{res}$}
      \vspace{23pt}

      \leftline{$D^l_{f}$}
      \vspace{11pt}
  \end{minipage}
  \begin{minipage}{0.28\linewidth}
      \begin{minipage}{0.7527\linewidth}
        \centerline{\includegraphics[width=\textwidth]{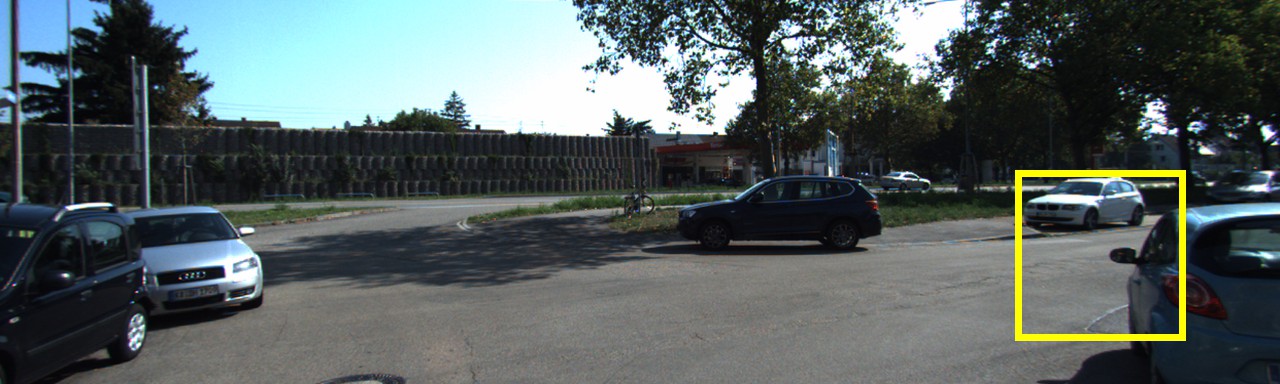}}
        \centerline{\includegraphics[width=\textwidth]{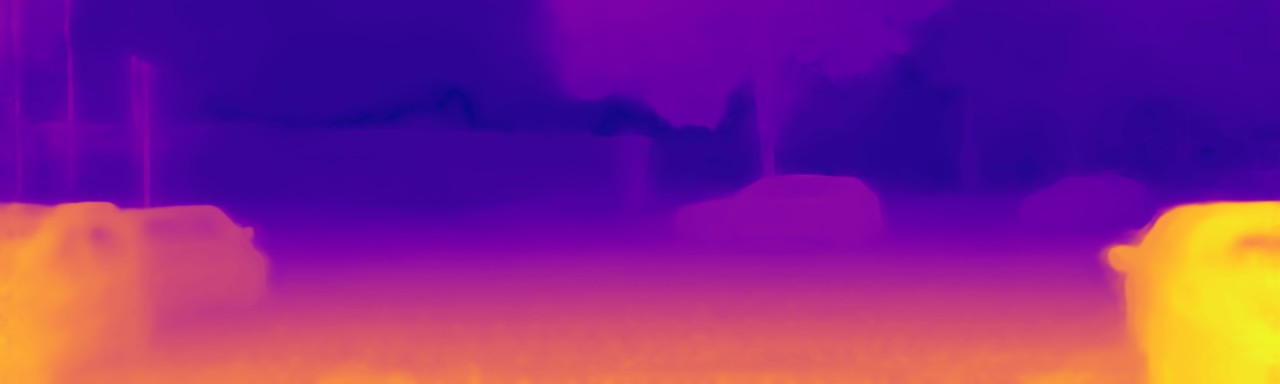}}
        \centerline{\includegraphics[width=\textwidth]{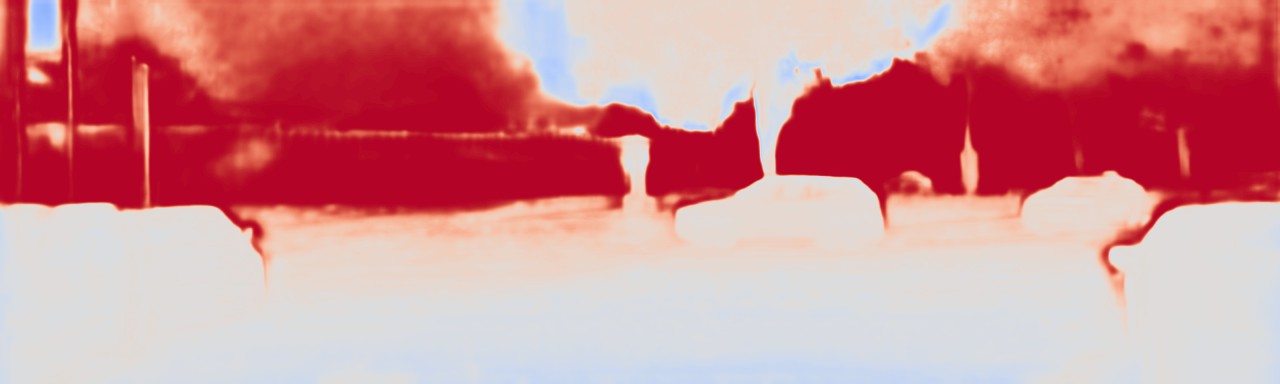}}
        \centerline{\includegraphics[width=\textwidth]{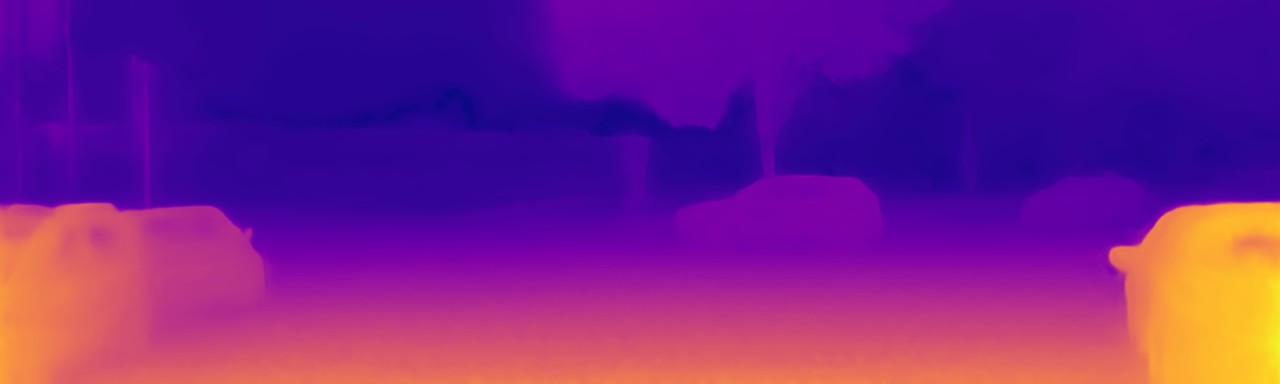}}
    \end{minipage}
    \begin{minipage}{0.2273\linewidth}
      \centerline{\includegraphics[width=\textwidth]{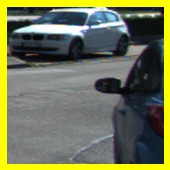}}
      \centerline{\includegraphics[width=\textwidth]{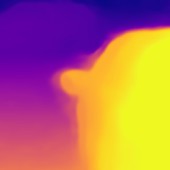}}
      \centerline{\includegraphics[width=\textwidth]{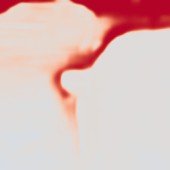}}
      \centerline{\includegraphics[width=\textwidth]{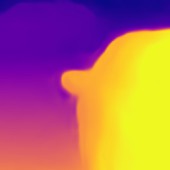}}
    \end{minipage}
    \centerline{(a)}
    \vspace{1pt}

    \begin{minipage}{0.7527\linewidth}
      \centerline{\includegraphics[width=\textwidth]{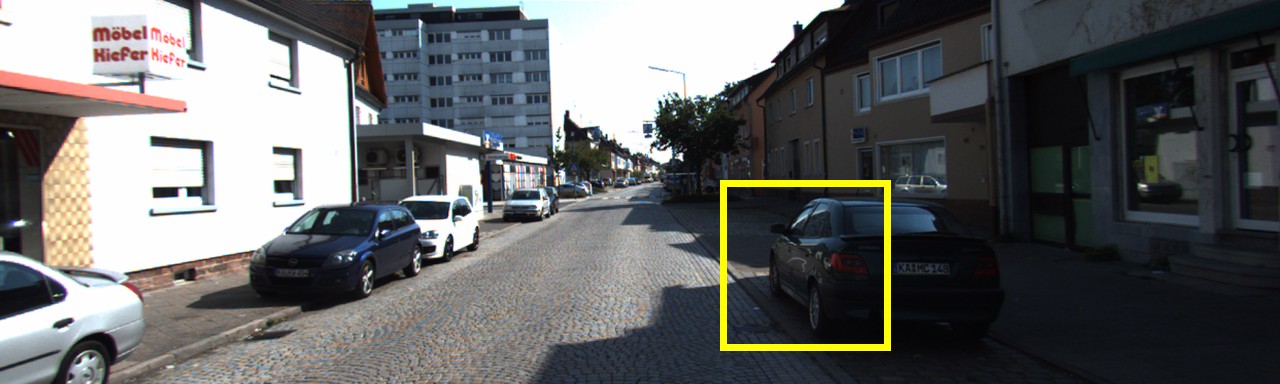}}
      \centerline{\includegraphics[width=\textwidth]{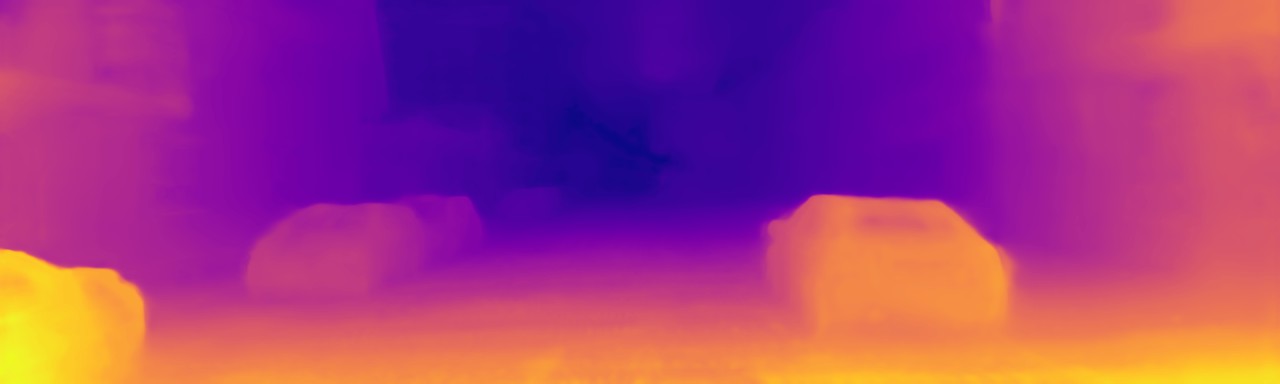}}
      \centerline{\includegraphics[width=\textwidth]{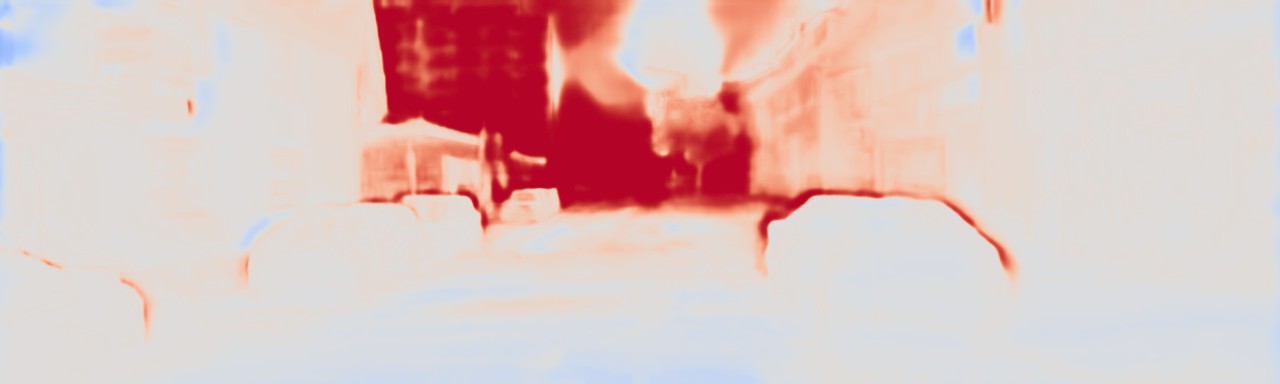}}
      \centerline{\includegraphics[width=\textwidth]{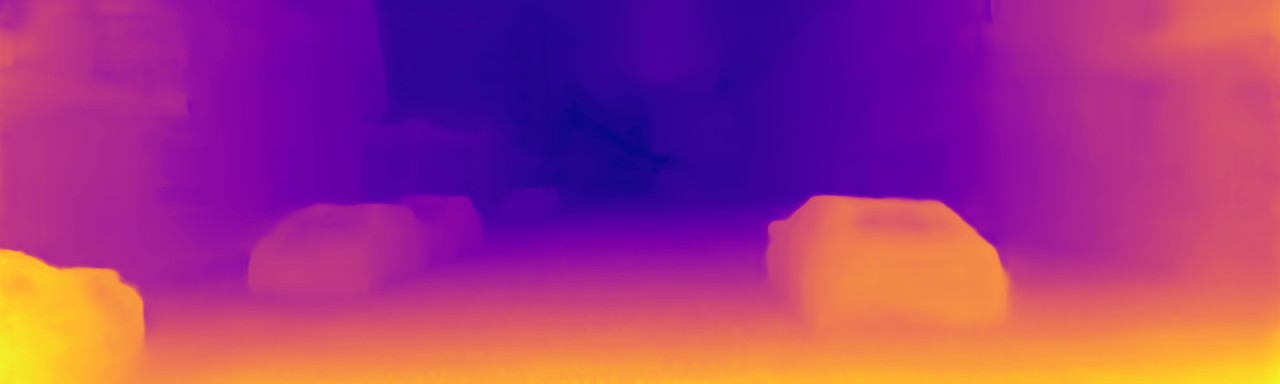}}
  \end{minipage}
  \begin{minipage}{0.2273\linewidth}
    \centerline{\includegraphics[width=\textwidth]{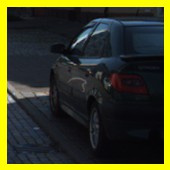}}
    \centerline{\includegraphics[width=\textwidth]{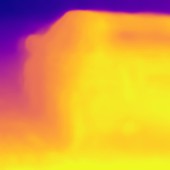}}
    \centerline{\includegraphics[width=\textwidth]{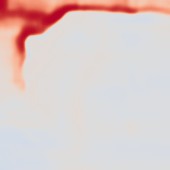}}
    \centerline{\includegraphics[width=\textwidth]{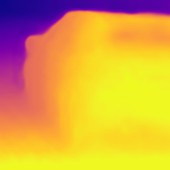}}
  \end{minipage}
    \centerline{(d)}
  \end{minipage}
  \begin{minipage}{0.28\linewidth}
  \begin{minipage}{0.7527\linewidth}
      \centerline{\includegraphics[width=\textwidth]{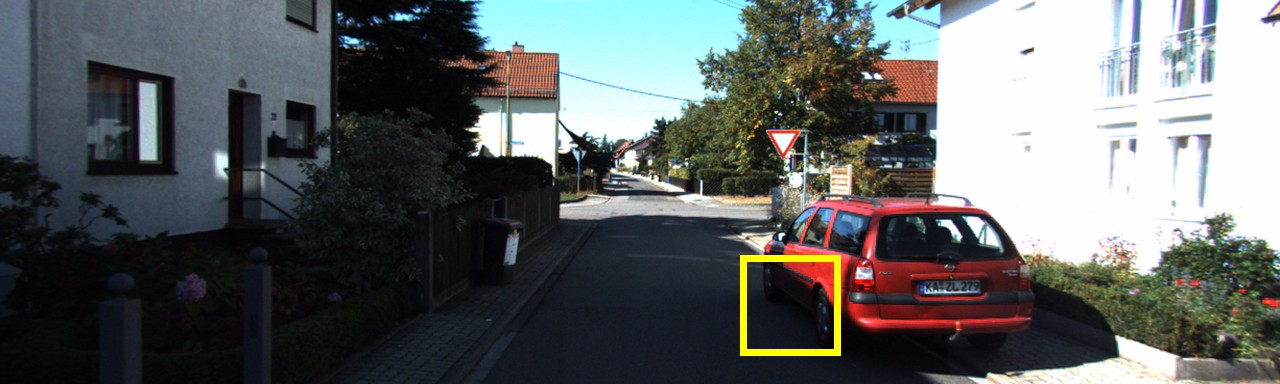}}
      \centerline{\includegraphics[width=\textwidth]{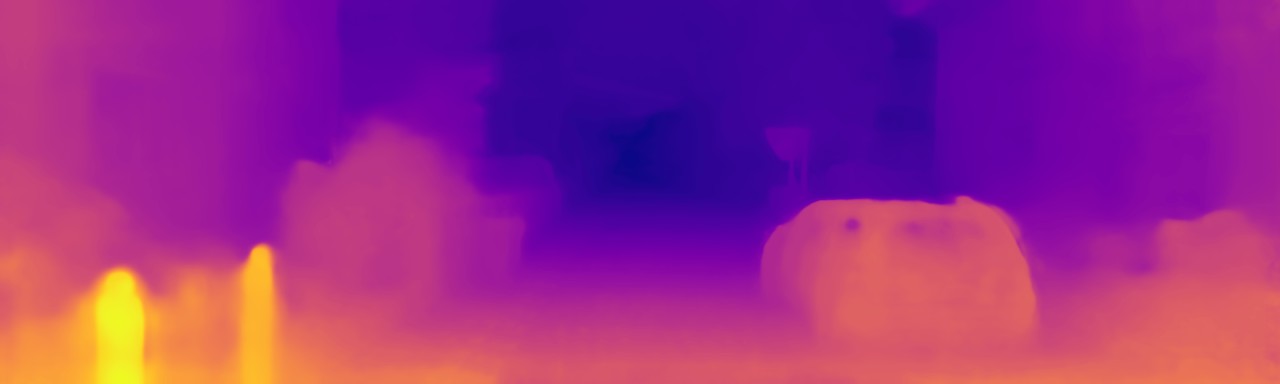}}
      \centerline{\includegraphics[width=\textwidth]{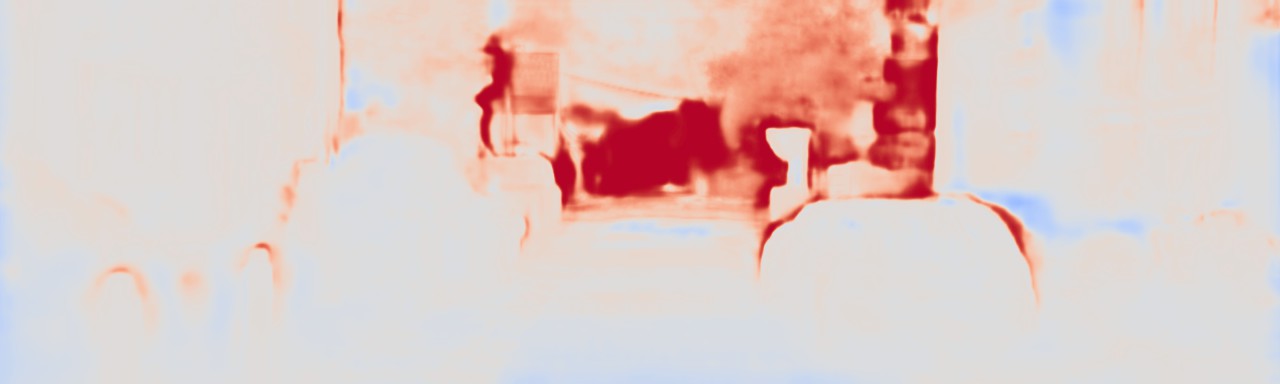}}
      \centerline{\includegraphics[width=\textwidth]{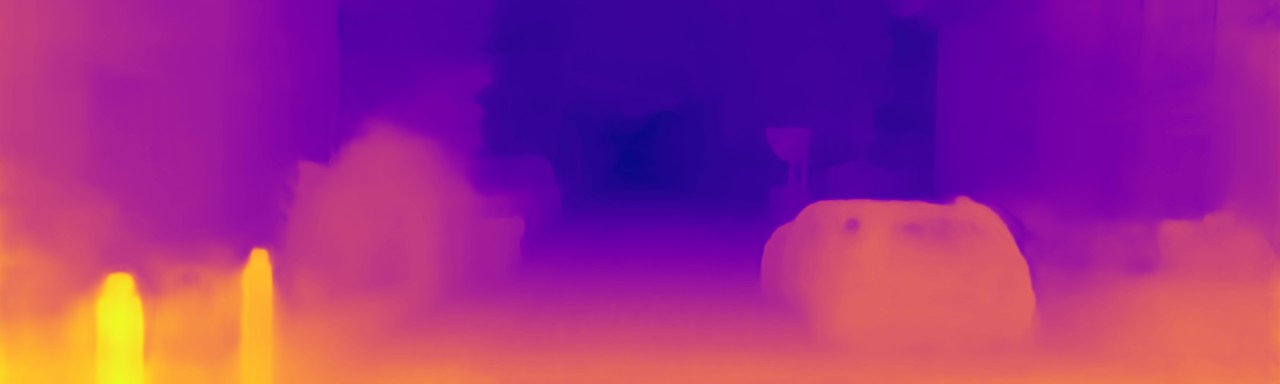}}
  \end{minipage}
  \begin{minipage}{0.2273\linewidth}
    \centerline{\includegraphics[width=\textwidth]{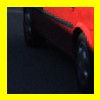}}
    \centerline{\includegraphics[width=\textwidth]{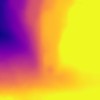}}
    \centerline{\includegraphics[width=\textwidth]{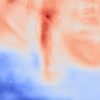}}
    \centerline{\includegraphics[width=\textwidth]{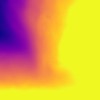}}
  \end{minipage}
  \centerline{(b)}
  \vspace{1pt}

  \begin{minipage}{0.7527\linewidth}
    \centerline{\includegraphics[width=\textwidth]{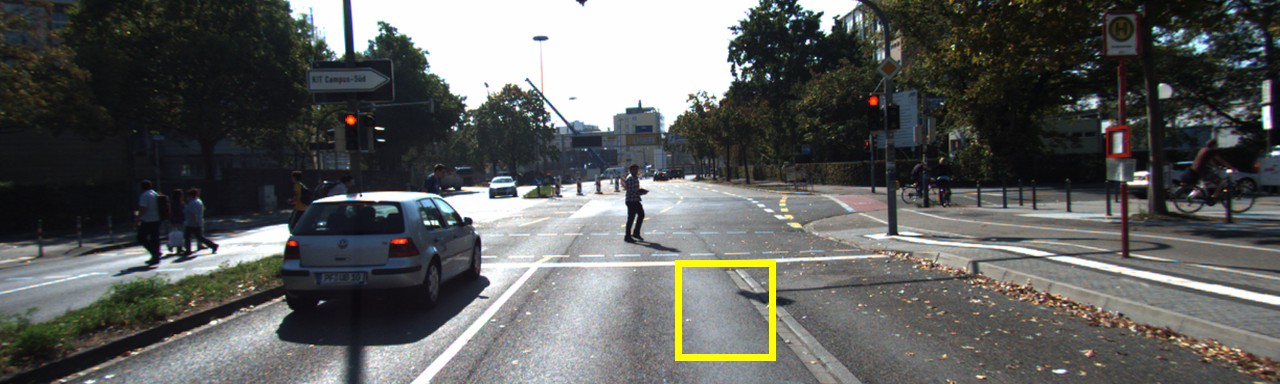}}
    \centerline{\includegraphics[width=\textwidth]{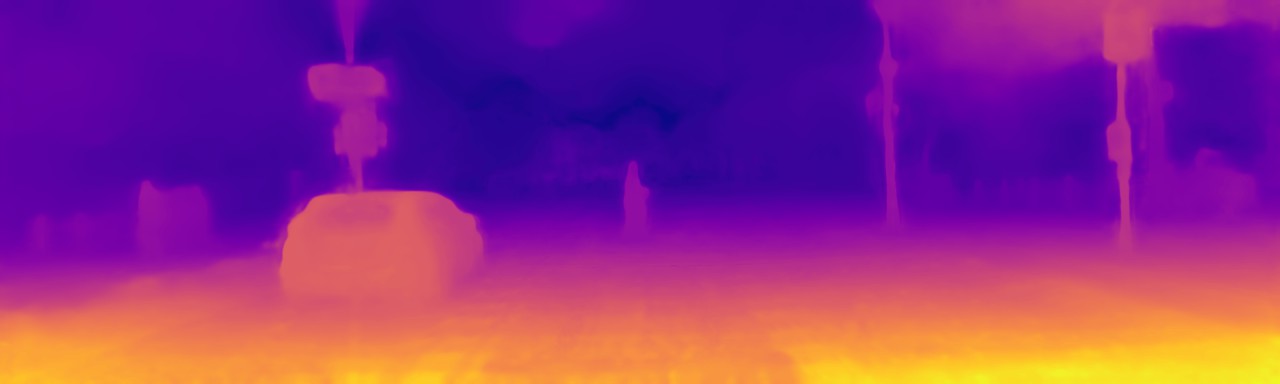}}
    \centerline{\includegraphics[width=\textwidth]{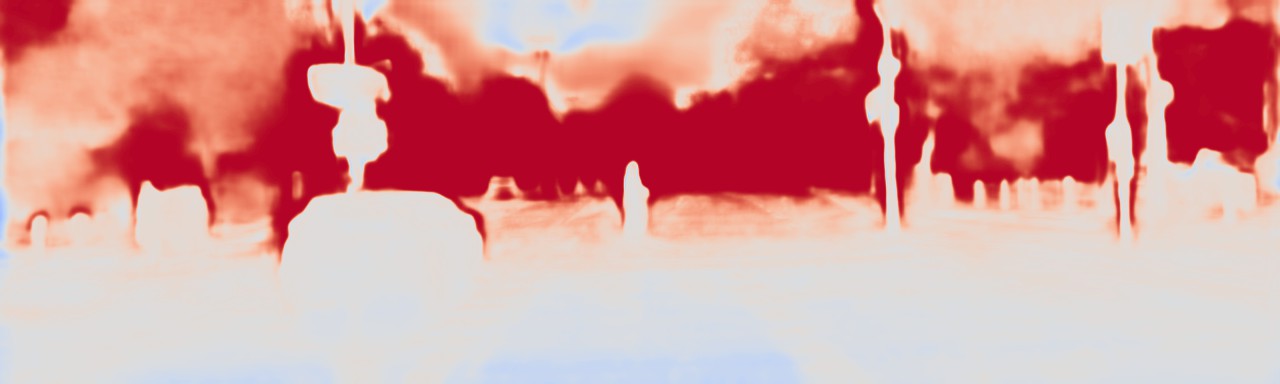}}
    \centerline{\includegraphics[width=\textwidth]{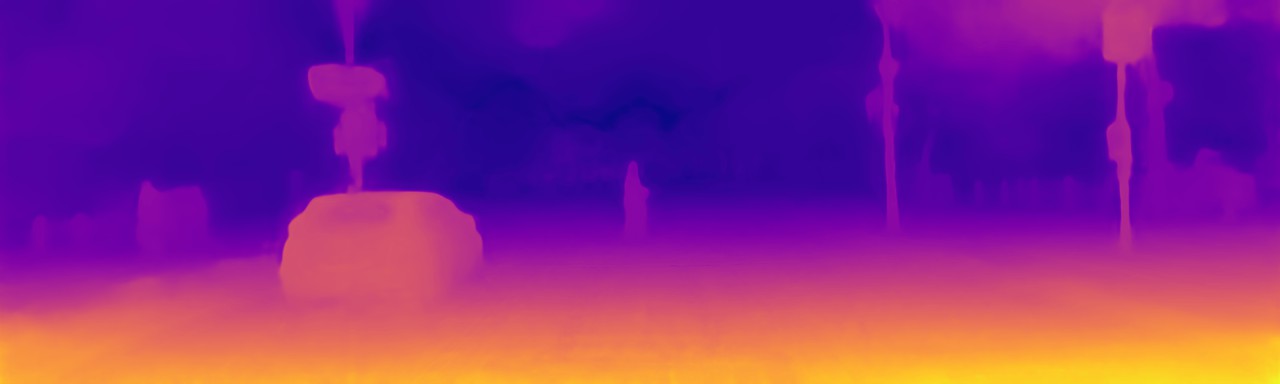}}
  \end{minipage}
  \begin{minipage}{0.2273\linewidth}
    \centerline{\includegraphics[width=\textwidth]{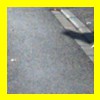}}
    \centerline{\includegraphics[width=\textwidth]{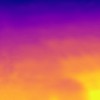}}
    \centerline{\includegraphics[width=\textwidth]{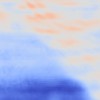}}
    \centerline{\includegraphics[width=\textwidth]{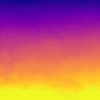}}
  \end{minipage}
  \centerline{(e)}
\end{minipage}
\begin{minipage}{0.28\linewidth}
  \begin{minipage}{0.7527\linewidth}
    \centerline{\includegraphics[width=\textwidth]{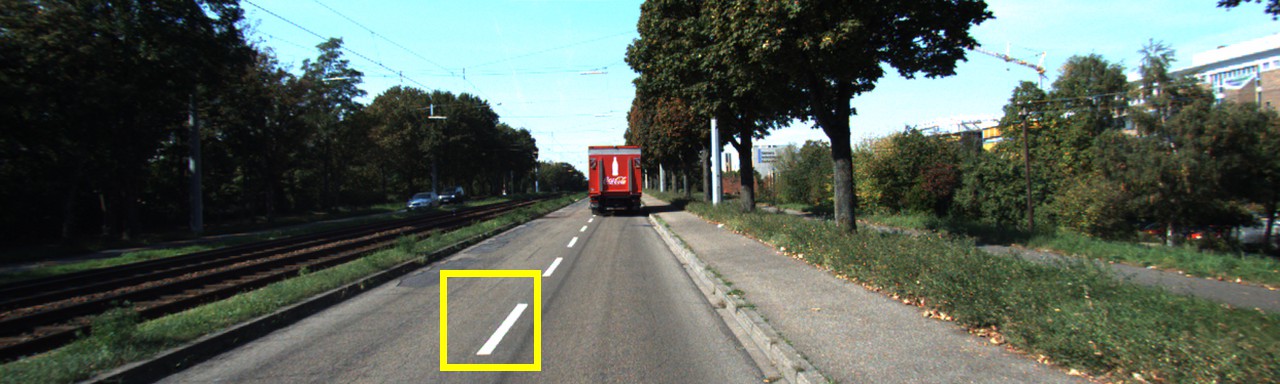}}
    \centerline{\includegraphics[width=\textwidth]{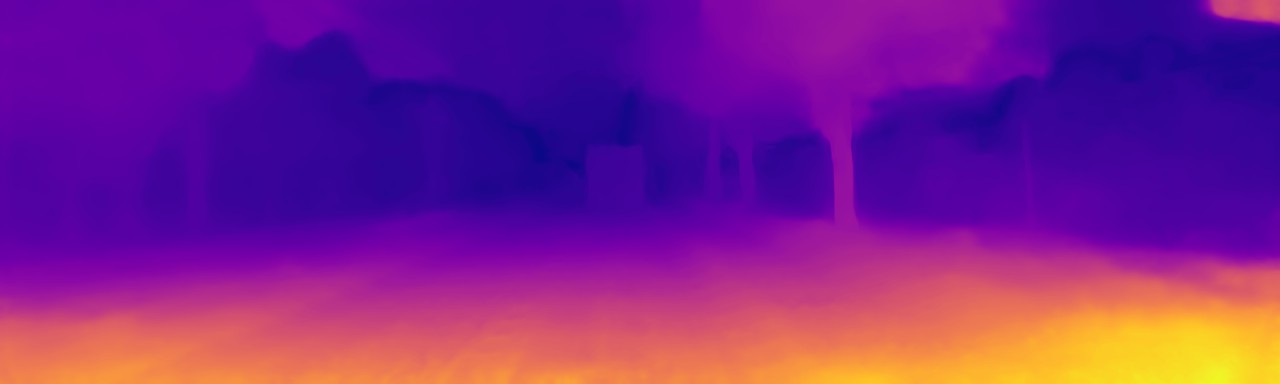}}
    \centerline{\includegraphics[width=\textwidth]{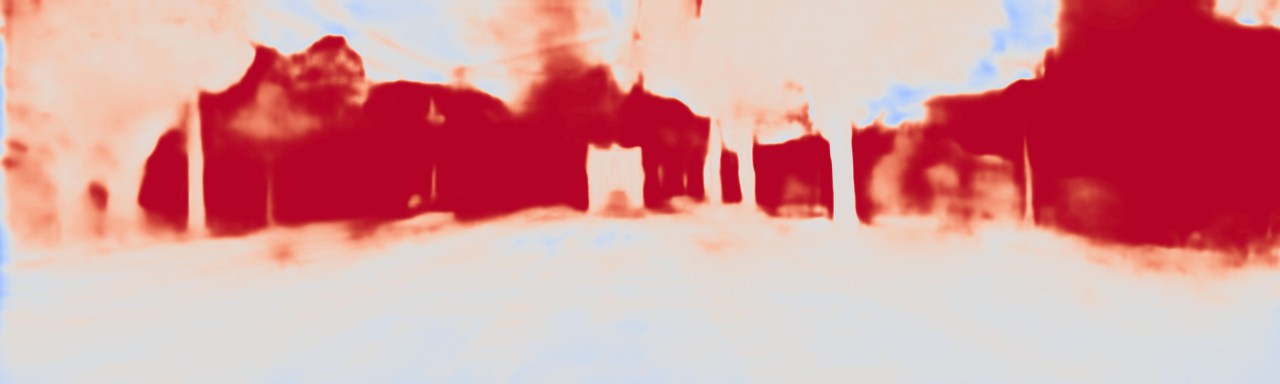}}
    \centerline{\includegraphics[width=\textwidth]{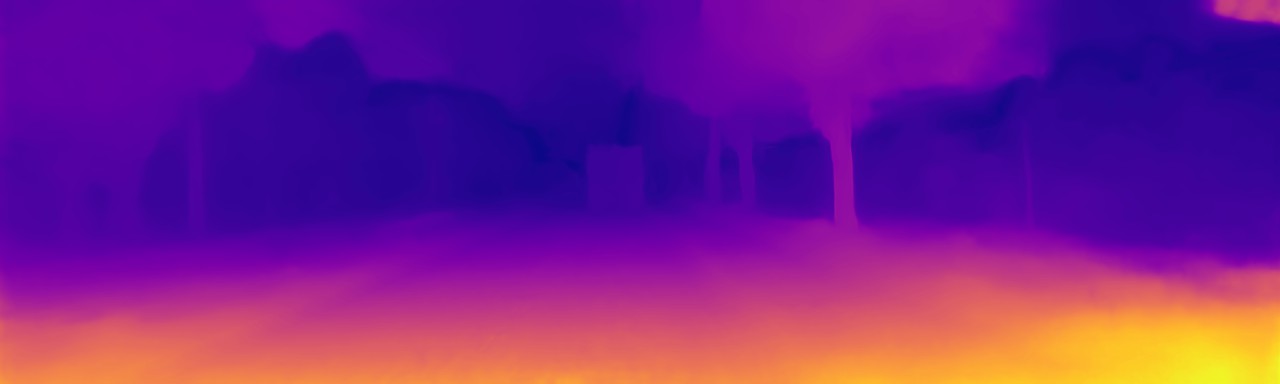}}
  \end{minipage}
  \begin{minipage}{0.2273\linewidth}
    \centerline{\includegraphics[width=\textwidth]{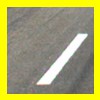}}
    \centerline{\includegraphics[width=\textwidth]{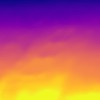}}
    \centerline{\includegraphics[width=\textwidth]{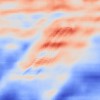}}
    \centerline{\includegraphics[width=\textwidth]{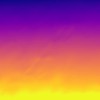}}
  \end{minipage}
\centerline{(c)}
\vspace{1pt}

\begin{minipage}{0.7527\linewidth}
  \centerline{\includegraphics[width=\textwidth]{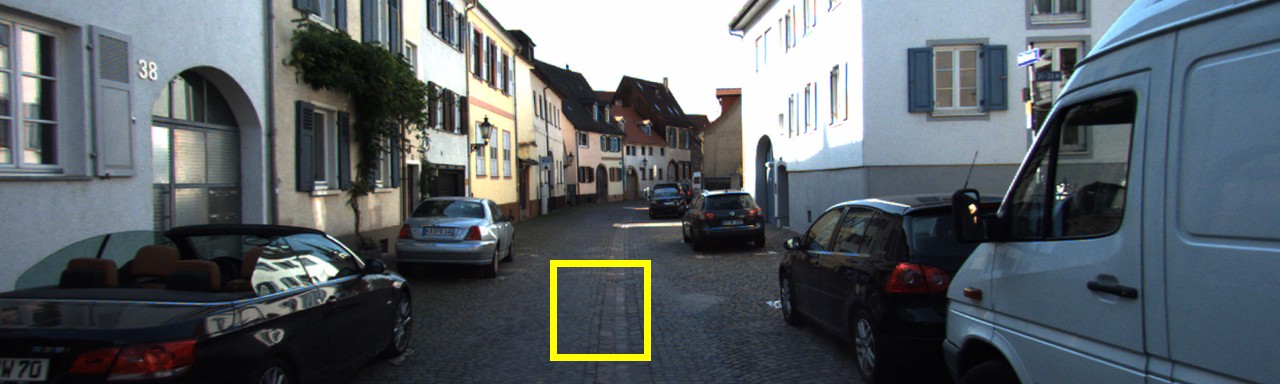}}
  \centerline{\includegraphics[width=\textwidth]{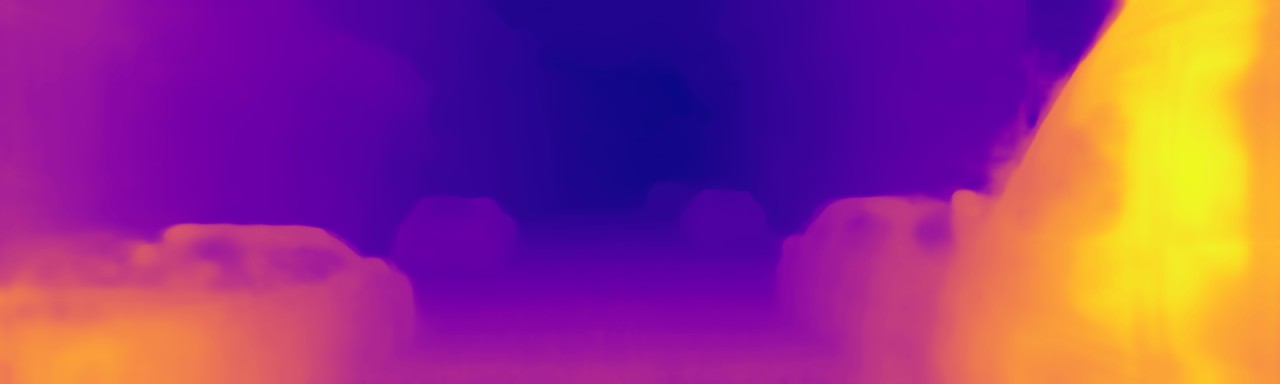}}
  \centerline{\includegraphics[width=\textwidth]{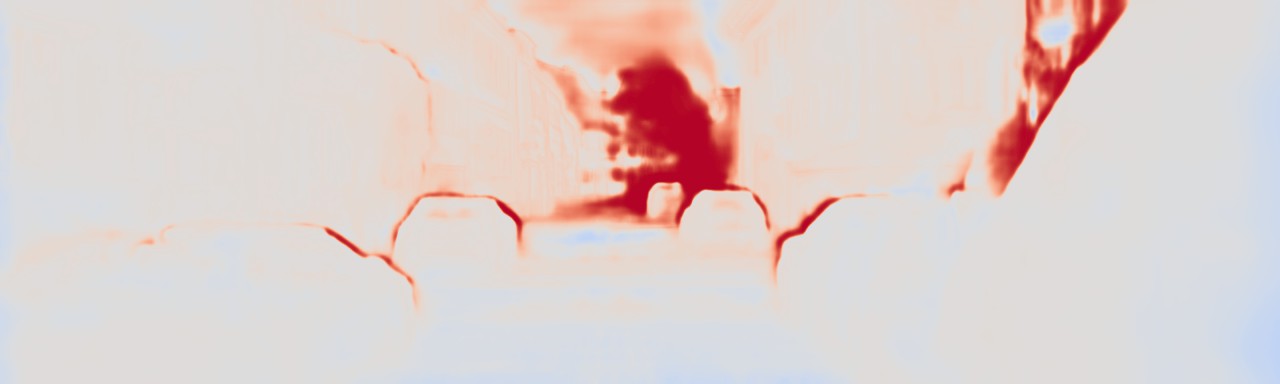}}
  \centerline{\includegraphics[width=\textwidth]{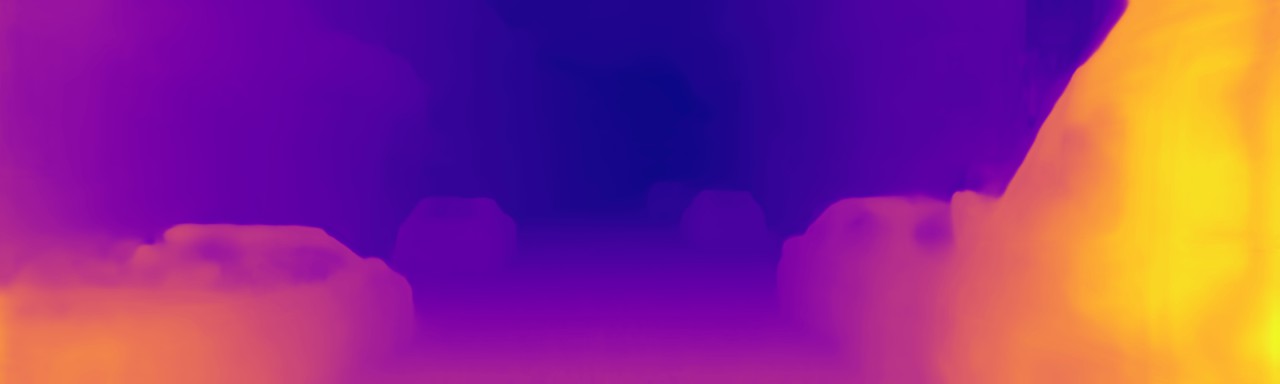}}
\end{minipage}
\begin{minipage}{0.2273\linewidth}
  \centerline{\includegraphics[width=\textwidth]{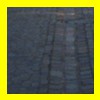}}
  \centerline{\includegraphics[width=\textwidth]{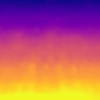}}
  \centerline{\includegraphics[width=\textwidth]{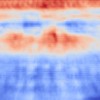}}
  \centerline{\includegraphics[width=\textwidth]{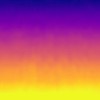}}
\end{minipage}
\centerline{(f)}
\end{minipage}
  
  \caption{Visualization results of the coarse-level depth map $D^l_c$, the scene depth residual map $D^l_{res}$ and the fine-level depth map $D^l_f$ on KITTI~\cite{Geiger2012We}.
  For showing the effects of the depth residual prediction branch more clearly, the images in the even columns are the enlarged versions of the yellow rectangle regions selected from the images in the odd columns, and the results in the even columns are re-normalized.
  For $D^l_{res}$, red indicates positive, and blue indicates negative.}
  \label{fig:add3}
\end{figure*}

Figure~\ref{fig:add3} illustrates the depth maps and the residual maps generated by OCFD-Net on KITTI~\cite{Geiger2012We}.
As seen from this figure, the depth residual map $D^l_{res}$ has large intensities on the relatively far regions, and provides smooth depth compensations for the coarse-level depth map $D^l_c$, resulting in the fine-level depth map $D^l_{f}$.

\end{document}